\documentclass[letterpaper]{article} 
\usepackage[draft]{aaai25}  
\usepackage{times}  
\usepackage{helvet}  
\usepackage{courier}  
\usepackage[hyphens]{url}  
\usepackage{graphicx} 
\usepackage{natbib}  
\usepackage{caption} 
\usepackage{algorithm}
\usepackage{algorithmic}
\usepackage{array}
\usepackage{booktabs}
\usepackage{pifont}
\usepackage{multirow}
\usepackage{subcaption}
\usepackage{subfig}
\usepackage{amsmath}
\usepackage{amssymb}

\usepackage{mathtools}
\usepackage{bm}
\usepackage{siunitx}
\usepackage{nicefrac}
\usepackage{cancel}
\usepackage{microtype}

\usepackage{newfloat}
\usepackage{listings}
\DeclareCaptionStyle{ruled}{labelfont=normalfont,labelsep=colon,strut=off} 
\floatstyle{ruled}
\newfloat{listing}{tb}{lst}{}
\floatname{listing}{Listing}
\title{FaceScore: Benchmarking and Enhancing Face Quality in Human Generation}
\author{
    Zhenyi Liao\textsuperscript{\rm 1,2 *},
    Qingsong Xie\textsuperscript{ \rm 2 $\dagger$},
    Chen Chen\textsuperscript{\rm 2},
    Haonan Lu\textsuperscript{\rm 2},
    Zhijie Deng\textsuperscript{ \rm 1 $\dagger$}
}
\affiliations{
    \textsuperscript{\rm 1}Qing Yuan Research Institute, SEIEE, Shanghai Jiao Tong University,
    \textsuperscript{\rm 2}OPPO AI Center\\

}

\usepackage{bibentry}

\begin{document}

\maketitle
\renewcommand{\thefootnote}{}
\footnotetext{$\dagger$ Corresponding authors.}
\footnotetext{$*$ Work done while at OPPO AI Center.}
\begin{abstract}
Diffusion models (DMs) have achieved significant success in generating imaginative images given textual descriptions. 
However, they are likely to fall short when it comes to real-life scenarios with intricate details.
The low-quality, unrealistic human faces in text-to-image generation are one of the most prominent issues, hindering the wide application of DMs in practice. 
Targeting addressing such an issue, we first assess the face quality of generations from popular pre-trained DMs with the aid of human annotators and then evaluate the alignment between existing metrics with human judgments. 
Observing that existing metrics can be unsatisfactory for quantifying face quality, we develop a novel metric named \textit{FaceScore (FS)} by fine-tuning the widely used ImageReward on a dataset of (win, loss) face pairs cheaply crafted by an inpainting pipeline of DMs. 
Extensive studies reveal \textit{FS} enjoys a superior alignment with humans. 
On the other hand, \textit{FS} opens up the door for enhancing DMs for better face generation.
With \textit{FS} offering image ratings,
we can easily perform preference learning algorithms to refine DMs like SDXL.
Comprehensive experiments verify the efficacy of our approach for improving face quality. 
The code is released at~\url{https://github.com/OPPO-Mente-Lab/FaceScore}.
\end{abstract}

%

\section{Introduction}
Diffusion models (DMs)~\cite{ho2020denoising,nichol2021improved,song2020score} have emerged as a prominent type of generative models, finding applications in various generative tasks such as audio generation~\cite{kongdiffwave,chen2020wavegrad}, video generation~\cite{blattmann2023stablevideodiffusionscaling,ho2022imagenvideohighdefinition,gupta2023photorealisticvideogenerationdiffusion}, and image inpainting~\cite{lugmayr2022repaint,avrahami2022blended,avrahami2023blended}. 
Among these tasks, text-to-image (T2I) DMs, such as Stable Diffusion (SD)~\cite{rombach2022high,podellsdxl}, Midjourney, and others~\cite{nichol2022glide,saharia2022photorealistic,ramesh2022hierarchical}, have garnered significant attention and achieved unprecedented success for their exceptional ability to generate content that surpasses human imagination. 

\begin{figure}[t]
\centering
\includegraphics[width=0.23\linewidth]{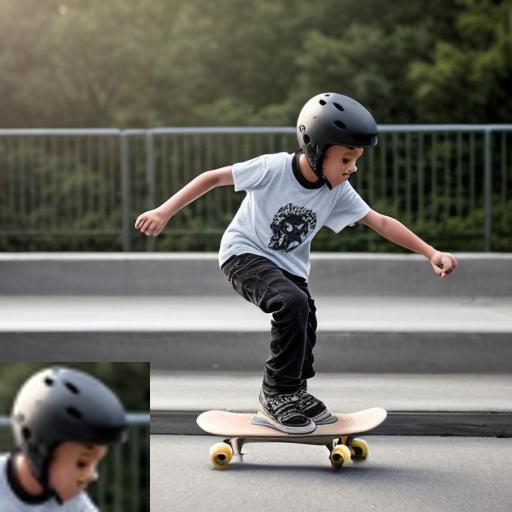}
\includegraphics[width=0.23\linewidth]{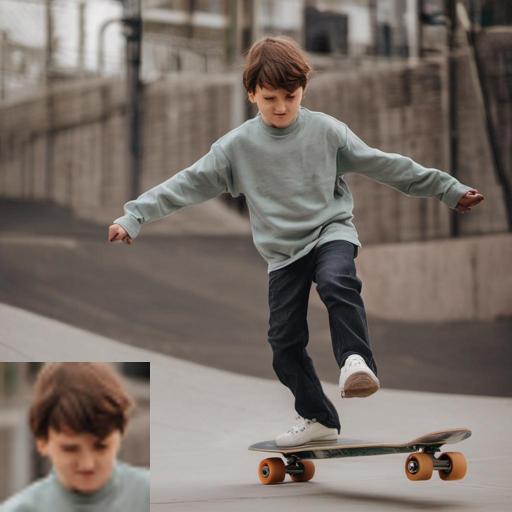}
\includegraphics[width=0.23\linewidth]{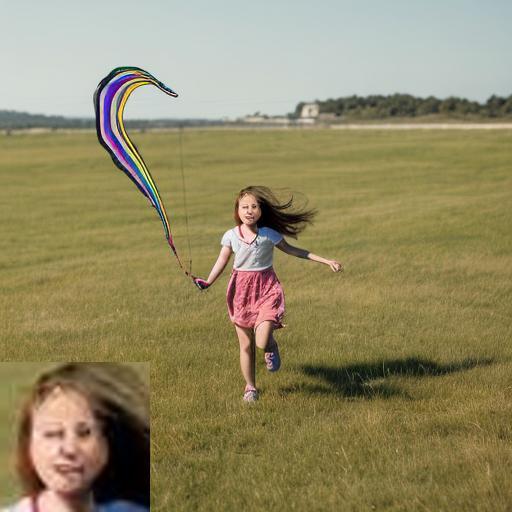} 
\includegraphics[width=0.23\linewidth]{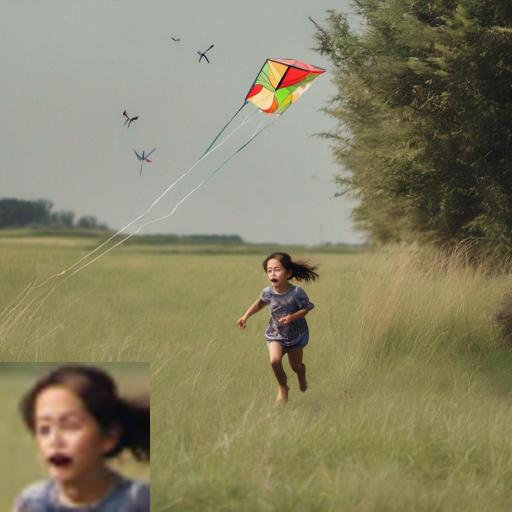} \\
\parbox{0.46\linewidth}{\centering \small A child is doing a trick \\on a skateboard}
\parbox{0.46\linewidth}{\centering \small A girl with a kite running \\in the grass}\\
\includegraphics[width=0.23\linewidth]{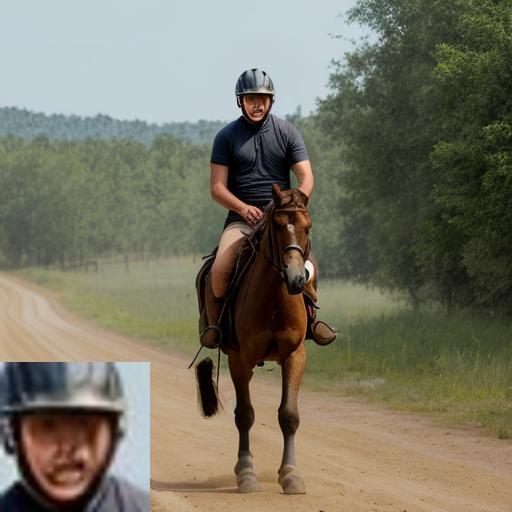} 
\includegraphics[width=0.23\linewidth]{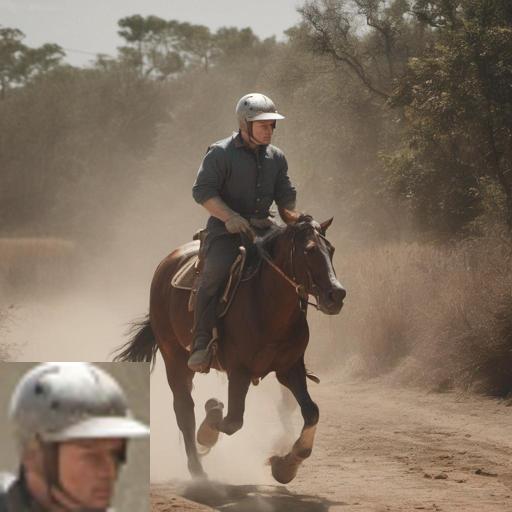} 
\includegraphics[width=0.23\linewidth]{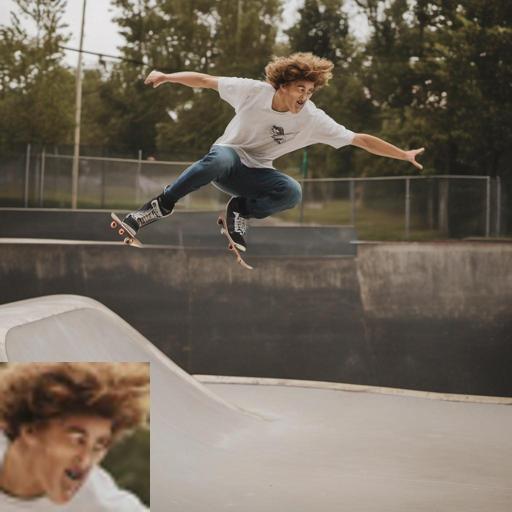} 
\includegraphics[width=0.23\linewidth]{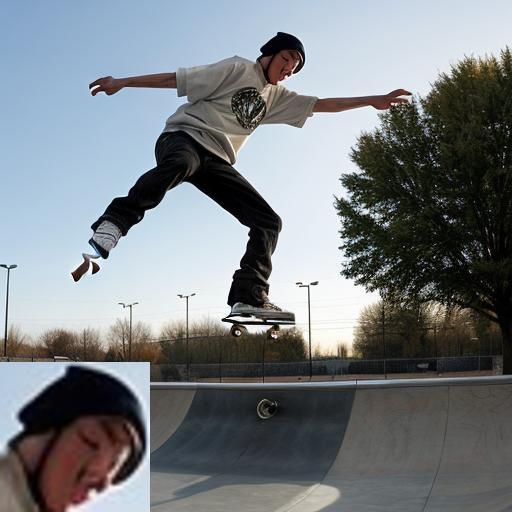}\\
\parbox{0.46\linewidth}{\centering \small A man in a helmet is riding a horse across a dirt road}
\parbox{0.46\linewidth}{\centering \small A male skater jumps in the air at a skate park}

\caption{Bad face generated by Realistic Vision V5.1 (the left one) and SDXL (the right one) with prompts below. 
Faces, especially small-scale faces, are highly likely to be vague and irrational.
We enlarge the face region and place it in the bottom left corner of the image.
Zoom in for face details.
}
\label{fig: bad face}
\end{figure}

Users can tolerate factual inaccuracies in imaginative generations, but the expectation changes when it comes to real-world settings, 
where distorted outcomes are routinely unacceptable. 
In particular, the generated bad faces (see Figure~\ref{fig: bad face}) are one of the most prominent issues for the human-oriented application of DMs. 
The possible causes of bad faces include that
(1) human faces encompass complex details, while their proportion within an image is often too small for DMs to attend to; 
(2) the amount of images containing human objects is limited for model training because of the involvement of human filters for safety considerations~\cite{esser2024scaling}.

To comprehensively investigate the bad face issue, we first empirically evaluate the face quality of generations from the prevalent Stable Diffusion V1.5 (SD1.5)~\cite{rombach2022high}, Realistic Vision V5.1 (RV5.1), and SDXL~\cite{podellsdxl}. 
We design a pipeline where human annotators rank the generated faces of the same prompt by different models and find that despite its smaller model size, RV5.1 achieves slightly superior results compared to SDXL. 
The evaluation results form a human preference dataset of face images, offering the possibility to quantify the alignment between human perception and existing popular metrics for synthetic images.
Thus, we assess ImageReward (IR)~\cite{xu2024imagereward}, Human Preference Score (HPS)~\cite{wu2023humanpreferencescorev2}, Aesthetic Score Predictor (ASP), and SER-FIQ~\cite{terhorst2020ser} for face quality assessment. 
We observe that these metrics can be unsatisfactory in assessing the rationality and aesthetic appeal of faces in synthetic images. 
That said, a new metric is urgently needed to bridge the gap.

By convention, the learning of an image metric entails access to a training dataset capturing the preference relationship, at the expense of human annotations. 
To avoid this, we innovatively propose to construct face-oriented preference data pairs based on the inpainting capacity of off-the-shelf pre-trained DMs---for a natural image containing faces, we detect, mask out, and inpaint the face regions to gain an image with degraded faces and hence a (win, loss) face pair.  
We fine-tune the typical ImageReward with such data pairs, yielding our \textit{FaceScore (FS)} metric. 
We conduct extensive studies to gain insights on the training behavior of \textit{FS} and also evidence that \textit{FS} enjoys a superior alignment with humans over existing metrics on face quality evaluation. 

We then leverage \textit{FS} to improve the face quality of existing DMs based on the preference learning paradigm. 
Specifically, \textit{FS} is used to rank the paired generations of the model of concern, and the model is tuned to adjust its likelihood based on the easy-to-use direct preference optimization (DPO)~\cite{wallace2024diffusion}. 
We clarify that other preference learning algorithms are compatible with \textit{FS}. 
Comprehensive experiments verify the efficacy of our approach for enhancing face quality, which also provides lateral evidence that \textit{FS} and human preferences are positively correlated.

In summary, our contributions can be listed as follows:
\begin{itemize}
    \item[$\bullet$] We perform the first investigation of the bad face issue of DMs and systematically assess a range of metrics for quantifying the face quality of synthetic images. 
    \item[$\bullet$] We propose \textit{FaceScore} (\textit{FS}) to reliably quantify the quality of generated faces, and prove that it surpasses existing metrics with a decent margin. 
    \item[$\bullet$] We leverage \textit{FS} to rate data pairs for preference learning and verify their efficacy on popular T2I diffusion models like SDXL through objective and subjective evaluations.
\end{itemize}

\section{Related Works}
\textbf{Text-to-image (T2I) diffusion models}~\cite{rombach2022high,nichol2022glide,saharia2022photorealistic,ramesh2022hierarchical} have undergone rapid developments and witnessed wide-\
spread applications. 
Given appropriate prompts as guidance, T2I DMs can generate visually appealing and semantically coherent images.
While T2I DMs excel at capturing the overall essence and content of the given prompts, they often struggle to generate intricate details and fine-grained features.

\subsubsection{Diffusion model fine-tuning and evaluation.}
Finetuning has empowered specific capabilities of DMs, such as extra image condition control~\cite{zhang2023adding}, adaptability to personal styles or figures~\cite{ruiz2023dreambooth,hulora}, instruction following~\cite{brooks2023instructpix2pix},
alleviation on gender and race bias~\cite{shenfinetuning}.
Aligning DMs with human preferences by fine-tuning is in emergence.
DMs can learn what humans find appealing by utilizing publicly available text-image datasets with annotations, such as Pick-a-pic~\cite{kirstain2023pick} and the Human Preference Dataset~\cite{wu2023humanpreferencescorev2}.
Reward function gradients~\cite{xu2024imagereward,clarkdirectly} and reinforcement learning methods~\cite{blacktraining,fan2024reinforcement} can also be applied.
Paralleling the alignment of large language models with human preference, direct preference optimization~\cite{rafailov2024direct} is also adopted as a counterpart for diffusion models~\cite{wallace2024diffusion}.
Still, they often fail to generate satisfactory faces.

For evaluation, HPS and IR bridge the gap for human preference object metrics.
They individually fine-tune CLIP~\cite{radford2021learning} and BLIP~\cite{li2022blip} as the backbone models to score the images for the degree of human preference.
However, these metrics focus on aesthetic appeal globally instead of local areas like faces, ignoring details generation.
Furthermore, the lack of a comprehensive human preference dataset for faces also hampers the progress in enhancing face quality in synthetic images.
Here we contribute a human preference dataset and an objective metric specifically for faces to fill up the gap.
\begin{figure*}[t]
    \centering
    \includegraphics[width=0.3\linewidth]{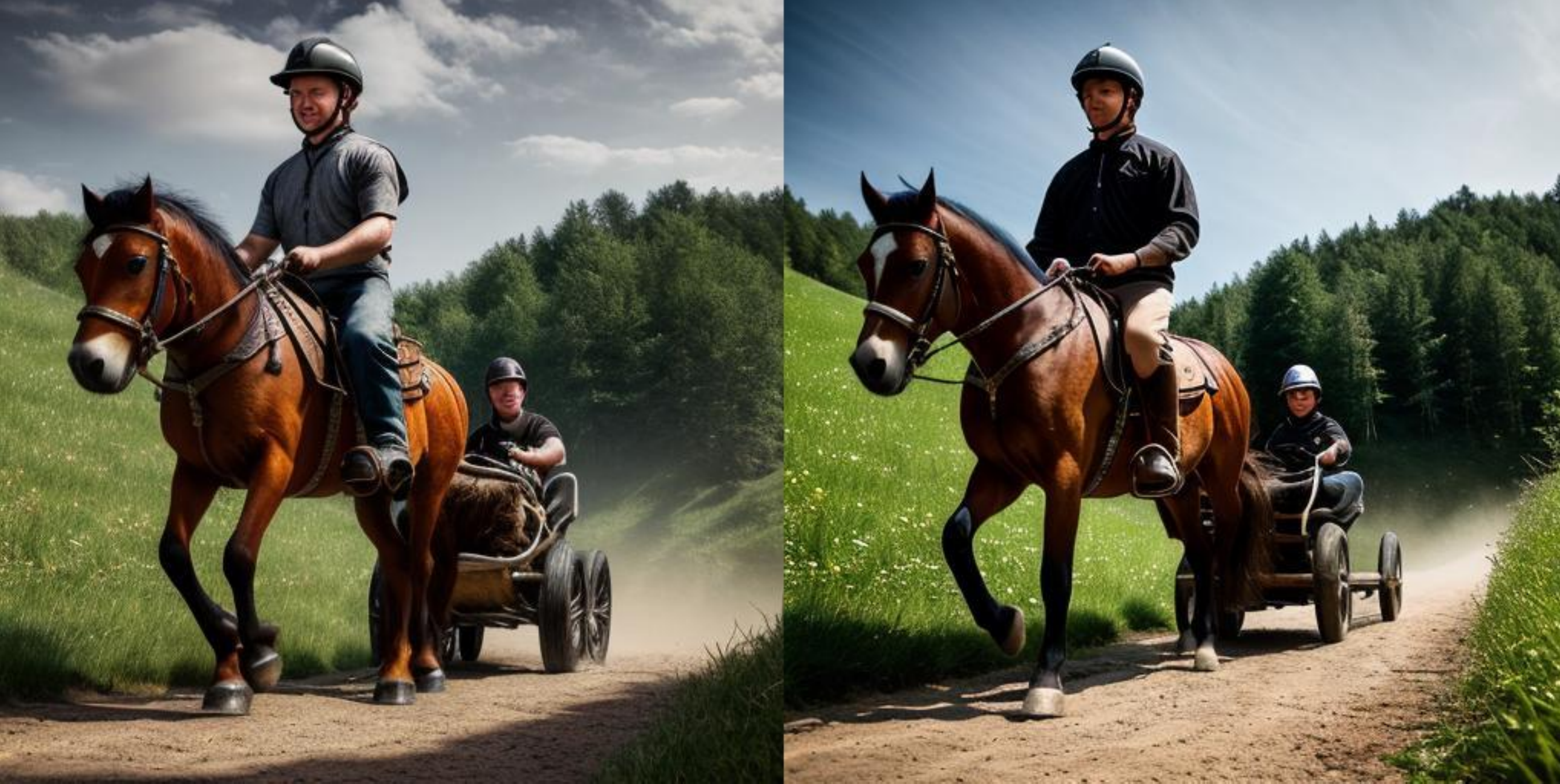}  \includegraphics[width=0.3\linewidth]{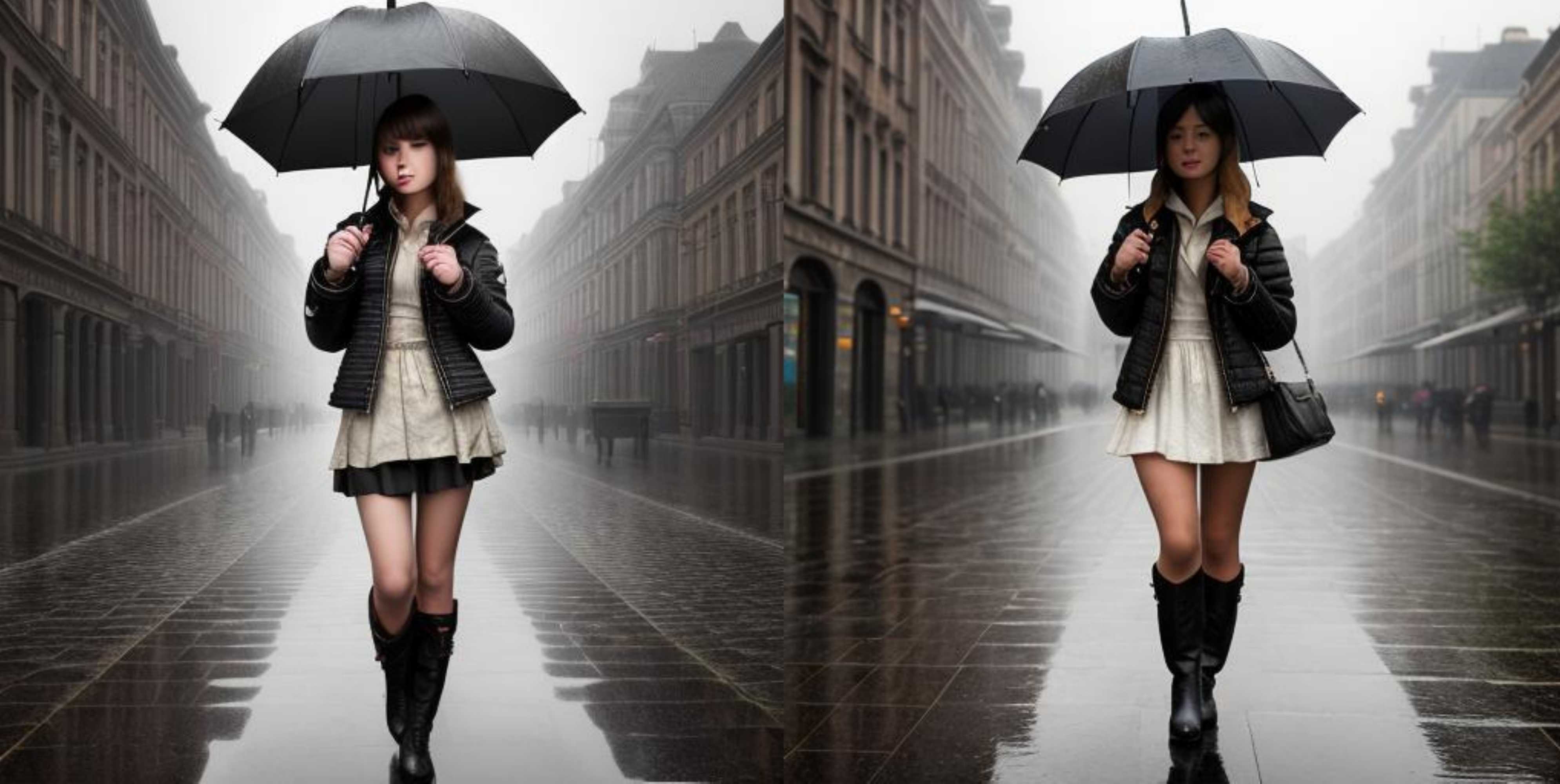} \includegraphics[width=0.3\linewidth]{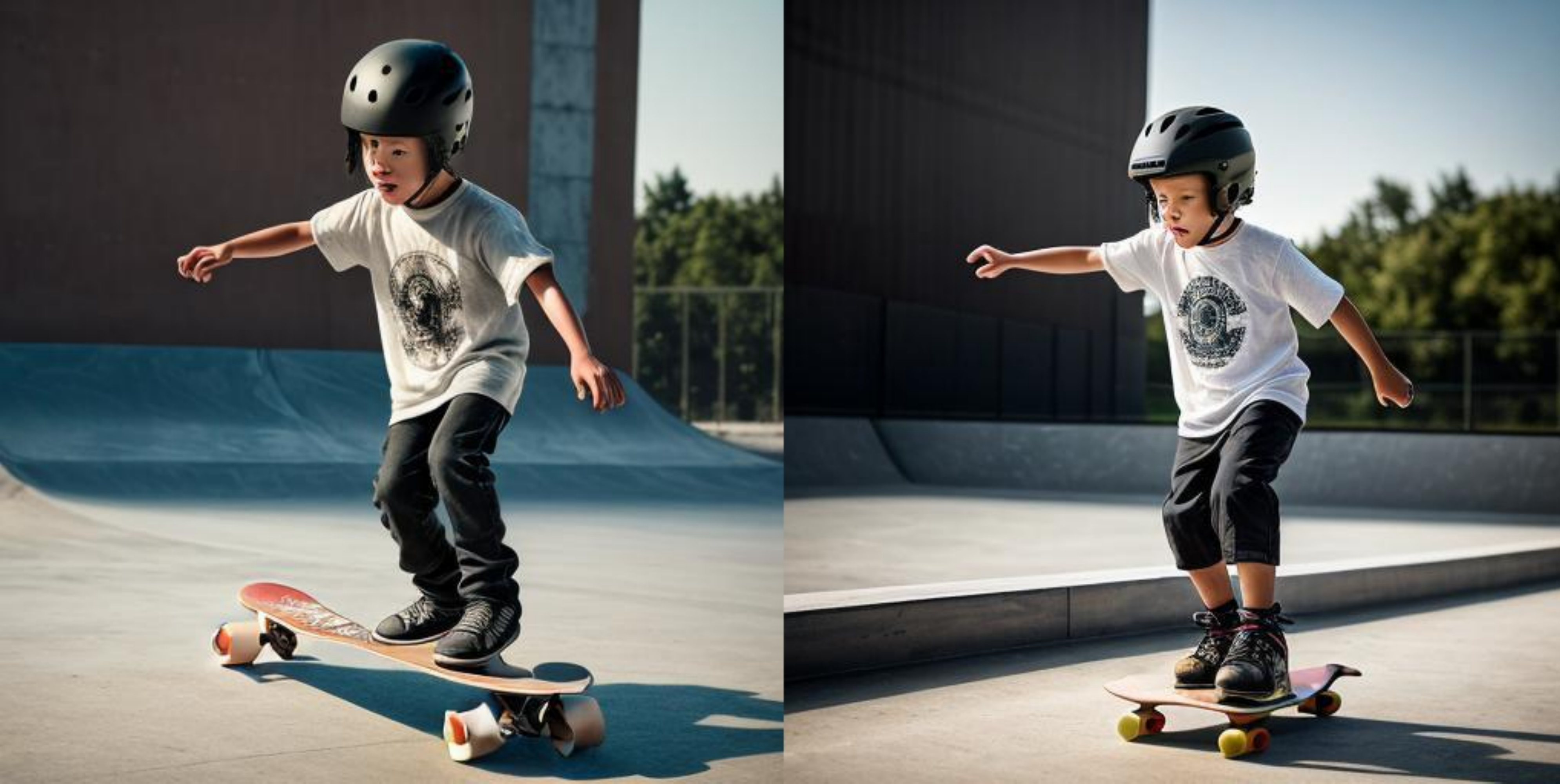} \\
    \parbox{0.3\linewidth}{\centering \small A man guiding a pony with \\a boy riding on it}
    \parbox{0.3\linewidth}{\centering \small A girl in a jacket and boots \\with a black umbrella}
    \parbox{0.3\linewidth}{\centering \small A child is doing a trick\\ on a skateboard}
    \caption{Comparison between generations sampled without (left) and with (right) negative prompts from Realistic Vision V5.1.
  Experiments are under the same conditions except for negative prompts, set as ``bad face, deformed, poorly drawn face, mutated, ugly, bad anatomy''.
  Enhancement can be observed in the face region with negative prompts.
  However, the generation still suffers from low quality.
  Zoom in for more face details.}
    \label{fig: neg}
\end{figure*}

\subsubsection{Detail generation.}
Previous studies have acknowledged the problem of detail generation like incorrect hands in DMs~\cite{podellsdxl}. 
HandRefiner~\cite{lu2023handrefiner} leverages a lightweight post-processing solution and utilizes ControlNet~\cite{zhang2023adding} modules to re-inject correct hand information for inpainting.
A concurrent work related to ours is HumanRefiner~\cite{fang2024humanrefinerbenchmarkingabnormalhuman}, whose sampling pipeline incorporates the inpainting process for better limbs, leading to slower inference speed.
We improve the face quality for the model by fine-tuning, and only need one sampling process, instead of in an inpainting way.

\section{Preliminary}
Let $x \in \mathcal{X}$ denote a natural image, i.e., $x \sim p_{data}$.
{Diffusion models} (DMs) gradually add Gaussian noise to $x$ in the forward process and are trained to perform denoising to achieve image generation~\cite{ho2020denoising,song2019generative}. 
Typically, the forward process takes the following transition kernel
\begin{equation}
    q(x_t|x_{0}) = \mathcal{N}(x_t;\alpha_tx_{0},\sigma_t^2 I),t = 1,\dots,T,
\end{equation}
where $x_0:= x$, and $\alpha_t$ and $\sigma_t$ are the pre-defined schedule parameters.
The forward process eventually renders $x_T \sim \mathcal{N}(0, I)$, i.e., the final state $x_T$ amounts to a white noise. 
The generation process of DMs reverses the above procedure with a $\theta$-parameterized Gaussian kernel:
\begin{equation}
    p_\theta(x_{t-1}|x_t) = \mathcal{N}(x_{t-1};\mu_\theta(x_t,t),\sigma^2_{t|t-1}\frac{\sigma^2_{t-1}}{\sigma_t^2} I),
\end{equation}
where $\sigma^2_{t|t-1}=\sigma_t^2-\frac{\alpha^2_{t}}{\alpha^2_{t-1}}\sigma^2_{t-1}$.
The mean prediction model $\mu_\theta(x_t,t)$ can be parameterized as a noise prediction one $\epsilon_\theta(x_t,t)$~\cite{ho2020denoising}, which is usually implemented as a U-Net~\cite{ronneberger2015u}. 

For efficient training and sampling, DMs can be shifted in the latent space~\cite{rombach2022high} with the help of an auto-encoder~\cite{esser2021taming}. 
Specifically, the image $x$ is first projected by the encoder to a low-dimensional latent representation $z = \mathcal{E}(x)$, and $z$ can be projected back to the image space by a decoder $\mathcal{D}$.

\section{Human Preference on Generated Face Images}
In this section, we first expose the bad face issue of existing DMs and test how good existing image-wise metrics are for quantifying the face quality of synthetic images.  
We then develop \textit{FaceScore (FS)} as a more qualified metric to assess the rationality and aesthetic appeal of generated face images. 

\subsubsection{The Bad Face Issue.}
The difficulties of DMs for generating intricate details, especially realistic human faces and hands, are no longer novel~\cite{podellsdxl}. 
As shown in Figure~\ref{fig: bad face}, images generated by RV5.1 and SDXL usually contain distorted faces. 
As previously discussed, the issue may originate from the scarcity of reliable face data in model training. 
To alleviate this,
it is a common practice to introduce negative prompts based on the classifier-free guidance (CFG) technique~\cite{ho2021classifier} to increase the chances of generating high-quality faces. 
Figure~\ref{fig: neg} displays results regarding this, where we see negative prompts indeed contribute to enhancing the face quality but the generated faces are still unsatisfactory. 
DM-based inpainting technique~\cite{avrahami2023blended,rombach2022high} can be performed to refine the face regions after generation, but the impainted faces can still be low-quality due to the fundamental pathology of the face generation capability of existing DMs, and sometimes we observe even worse outcomes. 

\subsubsection{Evaluation of Existing DMs.}
\begin{figure}[t]
    \begin{tabular}{ccc}
        Score 1 & Score 2 & Score 3 \\
        \includegraphics[width=2.25cm]{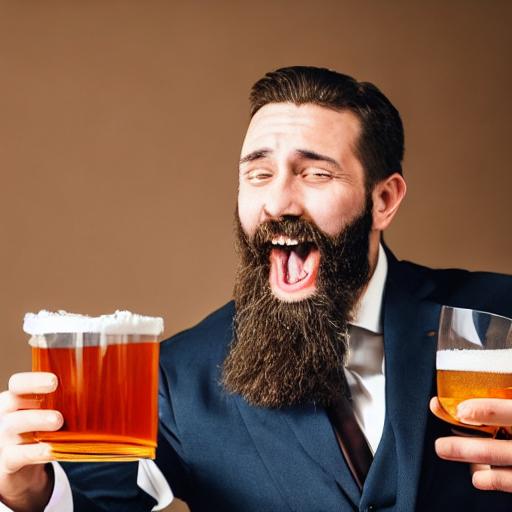}
        & \includegraphics[width=2.25cm]{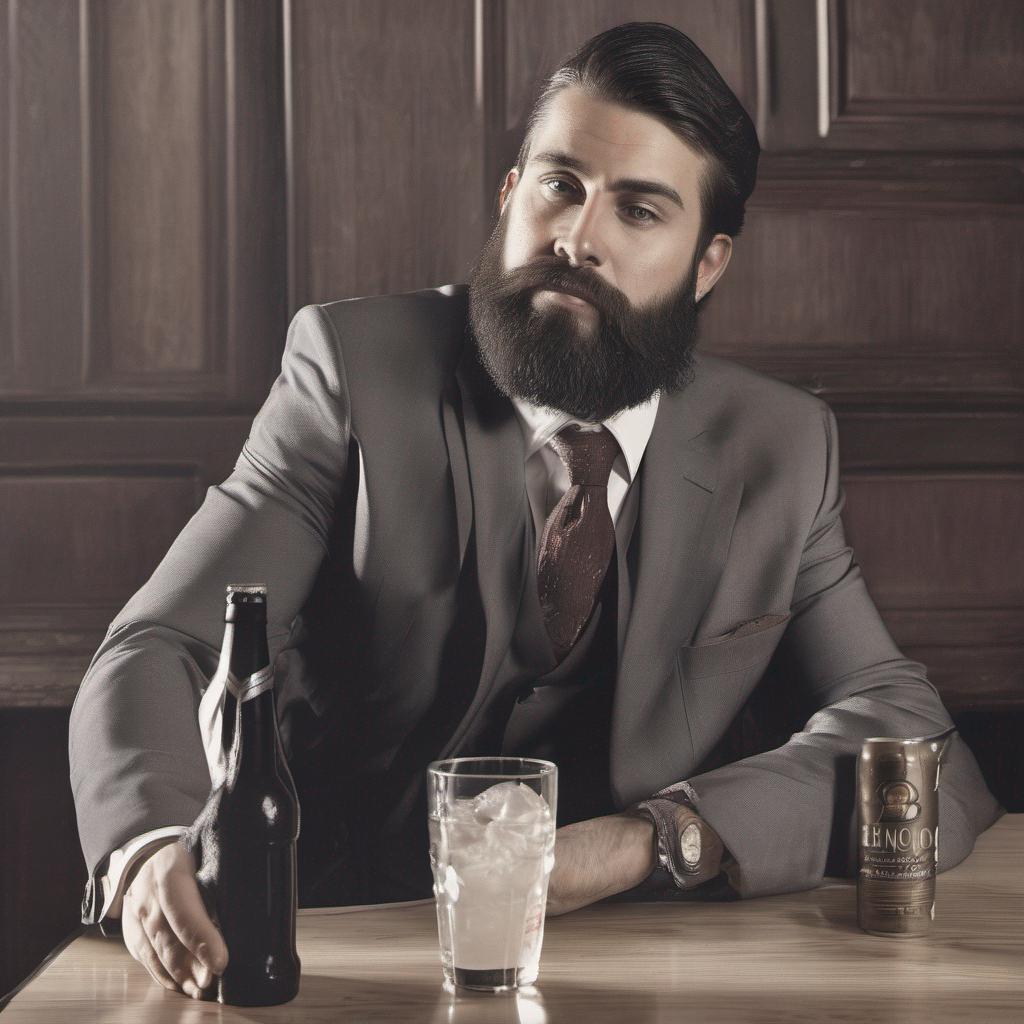}
        & \includegraphics[width=2.25cm]{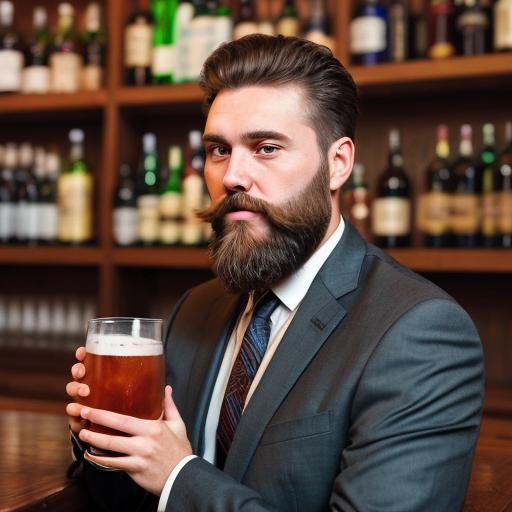} \\
        \multicolumn{3}{c}{Bearded man in a suit about to enjoy an adult beverage}\\
    \end{tabular}
    \caption{An example of the human-annotated triplet. The image with higher face quality is assigned a higher score. In each triplet, there are 3 binary comparisons.}
    \label{fig:facerank}
\end{figure}

Next, we conduct a detailed manual evaluation of the face generation quality across three popular DMs: SD1.5, RV5.1, and SDXL. 
Specifically, we leverage the following pipeline for evaluation:
\begin{itemize}
    \item select 1k prompts related to human subjects in the MS-COCO 2017 5K validation dataset~\cite{lin2014microsoft}, which includes descriptions of human-centric in\&outdoor scenes and single\&multi-person scenarios;
    \item for each prompt, generate a triplet of images (see Figure~\ref{fig:facerank} for an example) with the three DMs (the triplet is discarded if there are no valid faces in any image);
    \item introduce five human annotators to individually rank the triplet of each prompt based on face quality;
    the best image in the triplet receives a score of 3 and the worst receives a score of 1; 
    \item integrate the annotation results based on majority voting to avoid individual biases.
\end{itemize}

We calculate the 
frequency of equal to and more than three annotators among the five giving the same ratings for each image, obtaining 90.05\%, which reflects the high agreement of the five annotators and the annotation is convincing.
Figure~\ref{fig:facerank} presents an example of the annotated triplet (more in the Appendix) and Table~\ref{tab:annotation} displays the statistics of human preference over the three DMs. 
As shown, although the face quality of RV5.1 is not good enough (see Figure~\ref{fig: neg}), it still slightly surpasses the larger SDXL, which strengthens the concerns about the bad face issue of existing DMs. 
On the other hand, SD1.5 falls behind the other two DMs clearly.
\begin{table}[t]
    \centering
    \begin{tabular}{lrrrr}
    \toprule
    Models & Score 1 & Score 2  & Score 3  & Average score\\
    \midrule
     SD1.5 & 76.20\% & 18.65\% & 5.15\%  & 1.2895  \\
     RV5.1 & 12.22\% & 37.20\% & 50.58\% & \textbf{2.3836} \\ 
     SDXL  & 11.58\% & 44.16\% & 44.26\% & 2.3268  \\
    \bottomrule
    \end{tabular}
    \caption{Face quality comparisons between SD1.5, RV5.1, and SDXL.
    We present the proportion of each kind of score as well as the average score of each model. }
    \label{tab:annotation}
\end{table}

\subsubsection{Evaluation of Existing Metrics.}
A good metric can enable automatic, scalable evaluation of the face quality of the generations, avoiding expensive and time-consuming labeling processes by humans and paving the way for the development of new models. 
Next, we take an investigation on this---evaluating how well existing image-wise metrics are aligned with human preference on generated faces, based on the annotated triplets above. 

Concretely, we concern with IR, HPS, and ASP, which are prevalent for evaluating human preference or aesthetic quality in text-to-image generation.
We also take SER-FIQ~\cite{terhorst2020ser}, which accounts for face quality for recognition.
Intuitively, HPS and IR concentrate on the global image instead of the local area, so they are not suitable for evaluating the quality of generated faces. 
Thereby, we also develop variants of them, i.e., LocalHPS and LocalIR, where we detect the local face regions with a detector~\cite{deng2020retinaface} and send them into the original scoring pipeline with a default prompt ``A face'' for specific face evaluation.

We are majorly interested in the relative relationships of the metric evaluations on various images instead of the absolute numerical values. 
Luckily, the aforementioned pipeline for evaluation forms a small dataset containing roughly 1k annotated triplets, where each triplet forms three pairwise comparisons.
Thus, we evaluate the alignment between the metric and the human preference on such paired data.
This is, in fact, a metric score-based binary classification, so we list the corresponding accuracy in Table~\ref{tab:Acc}. 
We can observe that the performance of IR and ASP is unsatisfactory, perhaps due to their more attention on global image features, and LocalIR performs slightly better. 
SER-FIQ is poor as well because it is applied to evaluate the suitability of the face images for recognition and hence can be biased for the assessment of human preference on generated faces.
HPS and LocalHPS are the best among the metrics. 
Nonetheless, there is still considerable room for further improvement. 
\begin{table}[t]
    \centering
    \setlength{\tabcolsep}{3mm}
    \begin{tabular}{lrrr}
    \toprule[1.5pt]
    Methods & Backbone & Prompt & Acc.(\%) \\
    \midrule
     SER-FIQ & \textminus &  \textbf{\ding{55}} &  45.28 \\
    Random guess & \textminus &  \textbf{\ding{55}}  &  50.00 \\
    IR &  BLIP &  \textbf{\ding{51}} & 60.32 \\
    ASP & CLIP &  \textbf{\ding{55}} &  66.78 \\
    LocalIR &  BLIP &  \textbf{\ding{51}} & 69.68 \\
    HPS &  CLIP &  \textbf{\ding{51}} &  75.04 \\
     LocalHPS &  CLIP &  \textbf{\ding{51}} &  75.32 \\
    \midrule
    \textit{FS} (ours) &  BLIP &  \textbf{\ding{55}} &  \textbf{81.15} \\
    \bottomrule[1.5pt]
\end{tabular}
    \caption{Ranking alignment of existing popular metrics with human preference on generated face images. 
    We also include the proposed \textit{FaceScore (FS)} into comparison. }
    \label{tab:Acc}
\end{table}

\section{FaceScore: a Metric for Synthetic Face Images}
Given the above findings, we aim to develop a new metric to better quantify the human preference of synthetic face images. 
We dub such a metric as \textit{FaceScore (FS)} and expect it to correlate with both the rationality and aesthetic appeal of face generations. 
To achieve this, we construct a preference dataset on face images in an automatic and scalable way, based on which we perform model fine-tuning to obtain \textit{FS}. 
We also investigate proper strategies for the learning of \textit{FS}.

\subsubsection{Dataset construction.}
Given that popular open-source human preference datasets~\cite{kirstain2023pick,xu2024imagereward} are not specifically for faces, we are required to collect preference data on faces by ourselves. 
However, such data should be medium- to large-sized so that \textit{FS} is tuned with minimal biases, causing high labeling costs. 
To address this, we propose to leverage the inpainting capacity of off-the-shelf pre-trained DMs for constructing an automatic collection pipeline for paired data.
Specifically, we
\begin{itemize}
    \item detect the face regions of the natural images containing human faces in the LAION dataset~\cite{schuhmann2022laion} with existing detectors~\cite{deng2020retinaface}, obtaining face masks $M$;
    \item mask out and inpaint the face regions with a  DM-based inpainting pipeline~\cite{rombach2022high}.
\end{itemize}

\begin{figure}[t]
    \centering
    \begin{tabular}{cc}
        \textit{Original} & \textit{Inpainted} \\
        \includegraphics[width=3cm]{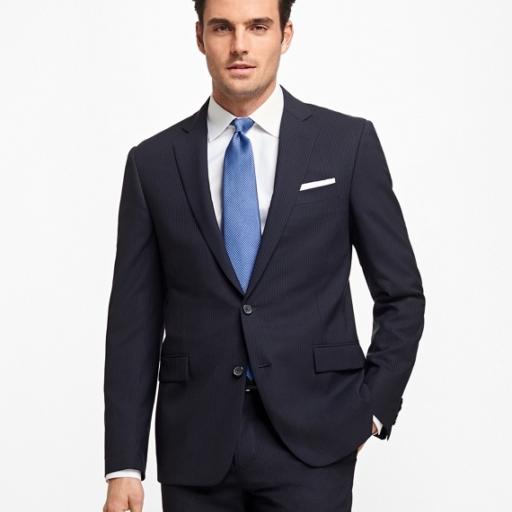}
        & \includegraphics[width=3cm]{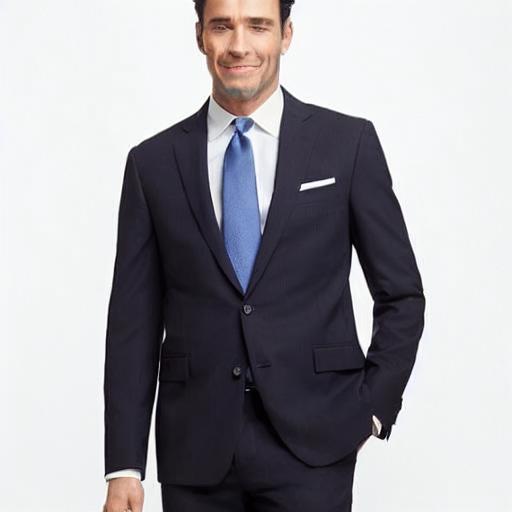} \\
        \multicolumn{2}{c}{ \textit{A man in a suit and tie standing up}} \\
    \end{tabular}
    \caption{An example of a face pair. We use the inpainting pipeline and control the noise strength for a degraded version, thereby forming a (win, loss) face pair.}
    \label{fig:facepair}
    \end{figure}

\begin{figure}
    \centering
\includegraphics[width=\linewidth]{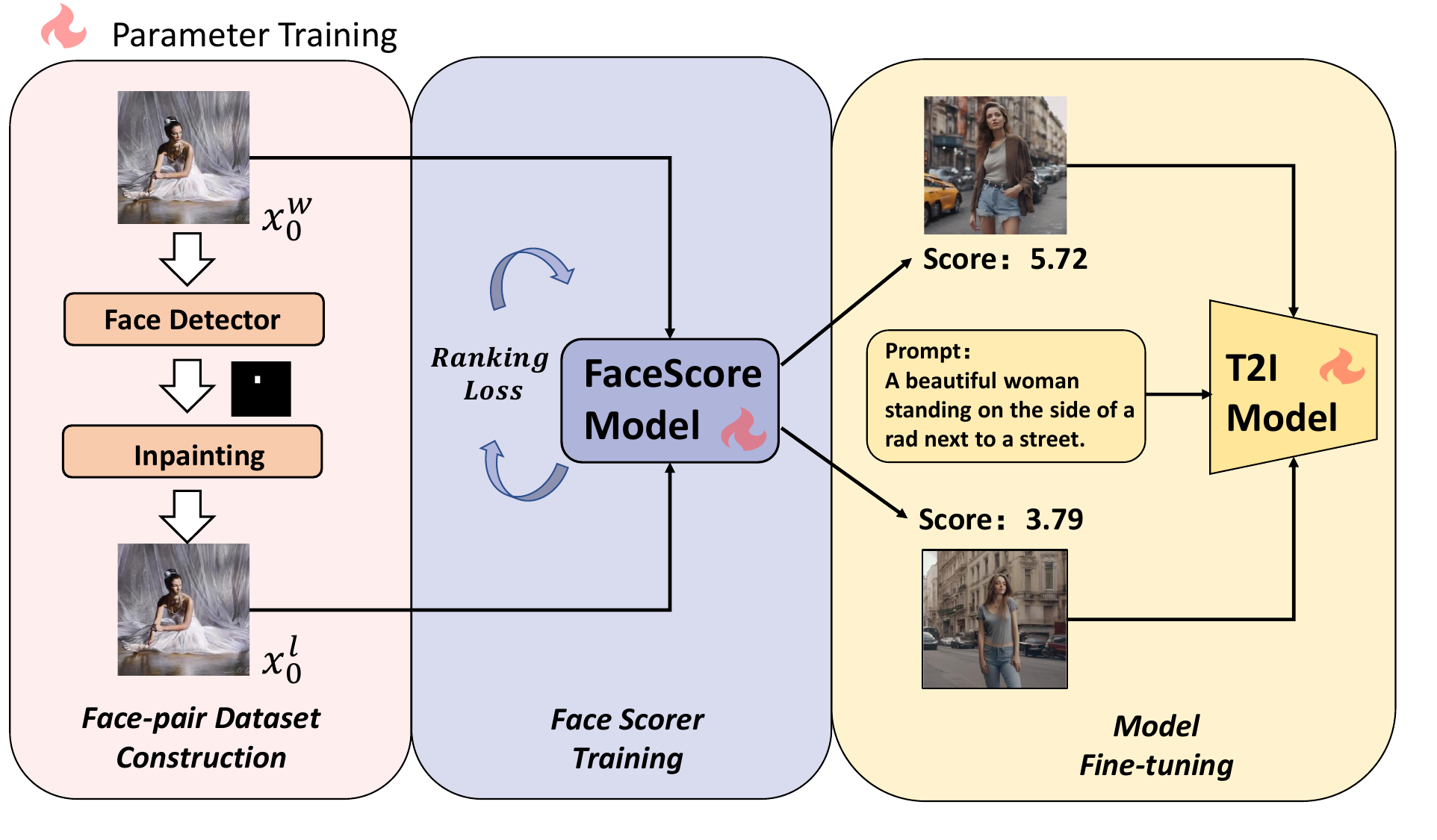}
    \caption{Overview of our pipeline.
    We leverage the inpainting pipeline on face images to get a negative sample, thus forming a (win, loss) face pair.
    We can use such a pair in fine-tuning an aesthetic scorer specifically for face quality.
    With such a metric, we can filter the data to fine-tune T2I diffusion models for better face quality.}
    \label{fig:pipeline}
\end{figure}

We plot the procedure in the left column of Figure~\ref{fig:pipeline}. 
The underlying hypothesis behind this is that the face quality of an inpainted image $x^{l}$ is worse than that of the original one $x^{w}$. 
This can be easily fulfilled by controlling the noise strength involved in the inpainting pipeline, and we have empirically verified this (an example is provided in Figure~\ref{fig:facepair}).

The above pipeline eventually produces a dataset $\mathcal{D}$ of 46k $(x^{w}, x^{l})$ pairs based on 23k natural images.
One natural image may correspond to a few inpainted images, since an image may contain more than one human face.

\subsubsection{Ranking Loss.}
We then would like to learn a scorer 
$s_\phi: \mathcal{X} \to \mathbb{R}$ 
to fit the preference dataset $\mathcal{D}$. \footnote{We can also input the text prompt corresponding to the image $x$ to the scorer, but omit it here for simplicity. } 
Drawn inspiration from the modeling of human preference over the aesthetic appeal of generated images~\cite{xu2024imagereward}, we utilize a naive ranking loss to tune $s_\phi$. 
Specifically, given a random mini-batch $\mathcal{B}$ from $\mathcal{D}$, we minimize the following loss:
\begin{equation}
    L_{rank}(\phi) = -\frac{1}{|\mathcal{B}|}\sum_{(x^{w},x^{l})\in \mathcal{B}}[\log(\sigma(s_\phi(x^{w})-s_\phi(x^{l})))],
\end{equation}
where $\sigma(\cdot)$ denotes the sigmoid function. 
Other possible learning principles are left as future work. 

\subsubsection{Fine-tuning IR.}
Considering the prevalence of IR and the improved capacity of BLIP architecture~\cite{li2022blip} over conventional CLIP~\cite{radford2021learning} for modeling human preference~\cite{xu2024imagereward}, we adopt IR to initialize our scorer $s_\phi$ and then perform fine-tuning to avoid the cold start problem. 
Noting that we only care about the face quality rather than the properties of the whole image, we detect faces in the image, as done in LocalIR, and tune the model on only the face regions.
The prompt is set to ``A face'' by default. 
We freeze the first 70\% layers of the backbone and train with a learning rate of $10^{-5}$. 
We find \textit{FS} holds a superior ability to rank human-annotated face images (see Table~\ref{tab:Acc}) and conduct the following ablation studies. 

\subsubsection{Global vs Local.}
As discussed, \textit{FS} only attends to the face regions of the images with the help of a face detector, empowered by the observation that LocalIR is better than vanilla IR in Table~\ref{tab:Acc}. 
We perform a set of empirical studies on this in Table~\ref{tab: setting}, where Global refers to considering the whole image following previous methods, while Local refers to cropping faces and setting default prompt as mentioned above.
As shown, the local strategy exceeds the global one by a considerable margin. 
The reason is that the face mostly only occupies a tiny part of the image, so intuitively the global evaluation is inaccurate. 

\begin{table}[t]
    \begin{minipage}{0.23\textwidth}
    \centering
    \setlength{\tabcolsep}{1mm}
    \begin{tabular}{lr}
    \toprule[1.5pt]
    Strategy & Alignment\\
    \midrule
    Global & 60.86\% \\
    Local/Fixed 0.1 & 79.67\%\\
    Local/Fixed 0.4   & 79.39\% \\
    Local/Adaptive & 81.15\% \\
    \bottomrule[1.5pt]
    \end{tabular}
    \caption{Human preference alignment under various settings on training the scorer.}
    \label{tab: setting}
\end{minipage}
\hfill
    \begin{minipage}{0.23\textwidth}
    \centering
    \setlength{\tabcolsep}{1mm}
    \begin{tabular}{lr}
    \toprule[1.5pt]
    Face area ratio & Noise factor\\
    \midrule
    (0\%,0.4\%] &Discarded \\
    (0.4\%,1\%] & 0.1 \\
    (1\%,10\%]&  0.2 \\
    (10\%,100\%]&  0.4 \\
    \bottomrule[1.5pt]
    \end{tabular}
    \caption{The adaptive strategy mapping the face area ratio to a specific noise factor.}
    \label{tab:ratio2factor}
    \end{minipage}
\end{table}

\subsubsection{Noise Factor.}
In the inpainting pipeline, the noise strength controls how much noise is added to the original image, directly influencing how similar the image is to the source one.
The larger the noise strength, the more the inpainted image differs from the source one. 
In fact, we roughly need bad faces that distribute similarly with the generations from the DM for tuning,
so we advocate controlling the noise factor to avoid the out-of-distribution bad faces. 
Given the observation that smaller faces are more easily destroyed during the inpainting process, 
we propose to adjust the noise factor based on the ratio of the area of the face regions compared to the whole image and identify a mapping strategy in Table~\ref{tab:ratio2factor}. 
We provide a comparison between this adaptive strategy and using fixed noise factors for dataset construction in
Table~\ref{tab: setting}, where Fixed 0.1 (0.4) refers to fixing the noise strength to 0.1 (0.4). 
We can observe that the adaptive strategy gets the best results than fixing the noise strength.

\begin{table}[t]
    \centering
    \setlength{\tabcolsep}{5mm}
    \begin{tabular}{lr}
    \toprule[1.5pt]
    Model & \textit{FS} \\
    \midrule
    Stable Diffusion V1.5& 0.75\\
    Stable Diffusion V2.1& 2.34\\
    SDXL & 2.36 \\
    Realistic Vision V5.1 & 3.14\\
    DreamLike V2.0 & 3.18\\
    Kandinsky-3~\cite{arkhipkin2024kandinsky30technicalreport} & 3.52\\
    Stable Cascade~\cite{perniaswurstchen} & 3.53\\
    ProtoVision V6.6 & 3.63\\
    Playground V2.5~\cite{li2024playgroundv25insightsenhancing} & 3.98\\
    PixArt-$\alpha$ ~\cite{chenpixart}& 3.98\\
    SD3~\cite{esser2024scaling} & 4.00 \\
    Hunyuan~\cite{li2024hunyuanditpowerfulmultiresolutiondiffusion} & 4.10\\
    Kolors~\cite{kolors} & 4.49 \\
    \bottomrule[1.5pt]
    \end{tabular}
    \caption{Unnormalized \textit{FS} for different text-to-image diffusion models. We use the same prompts selected from MS-COCO mentioned above in the evaluation. 
    }
    \label{tab:different fs}
\end{table}

\subsubsection{Quantitative Comparisons.}
\begin{table}[t]
    \centering
    \begin{tabular}{lrrrr}
    \toprule[1.5pt]
    Model & PC & SD1.5 & RV5.1 & SDXL   \\
    \midrule
    SER-FIQ & -0.1161 & 0.4704 & 0.4415 & 0.4153 \\
    LocalHPS & 0.5815 & 0.4349 & 0.5448 & 0.4575 \\
    FS & 0.6855 & 0.5301 & 0.7284 & 0.7012 \\
    \bottomrule[1.5pt]
    \end{tabular}
    \caption{The Pearson correlation (PC) between human ranking and the metrics. 
     We also report the average (normalized) score for each set of images generated from each DM.
    }
    \label{tab:global}
\end{table}
\begin{figure*}[t]
  \centering
  \begin{tabular}{ccccccc}
    \toprule  
    \includegraphics[width=2cm]{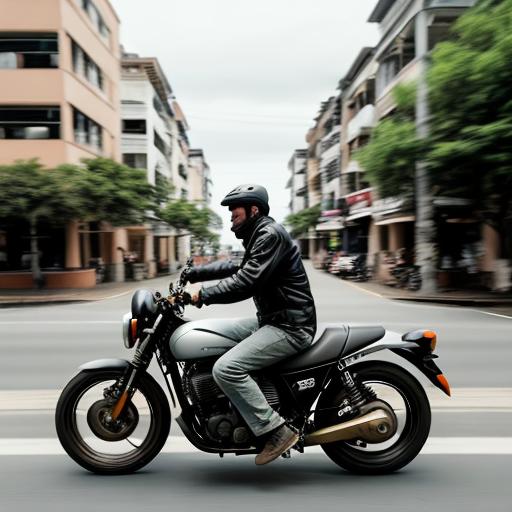} & \includegraphics[width=2cm]{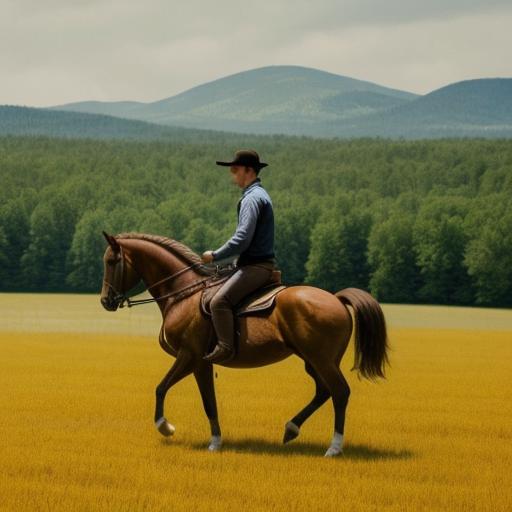} & \includegraphics[width=2cm]{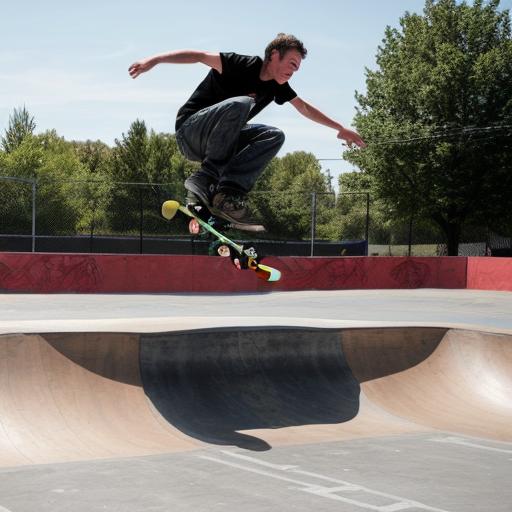} & \includegraphics[width=2cm]{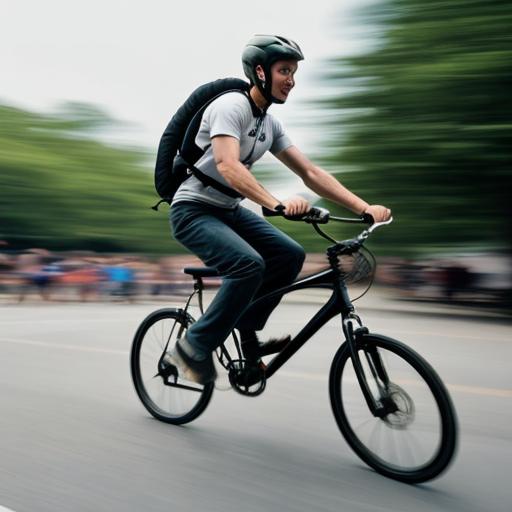} & \includegraphics[width=2cm]{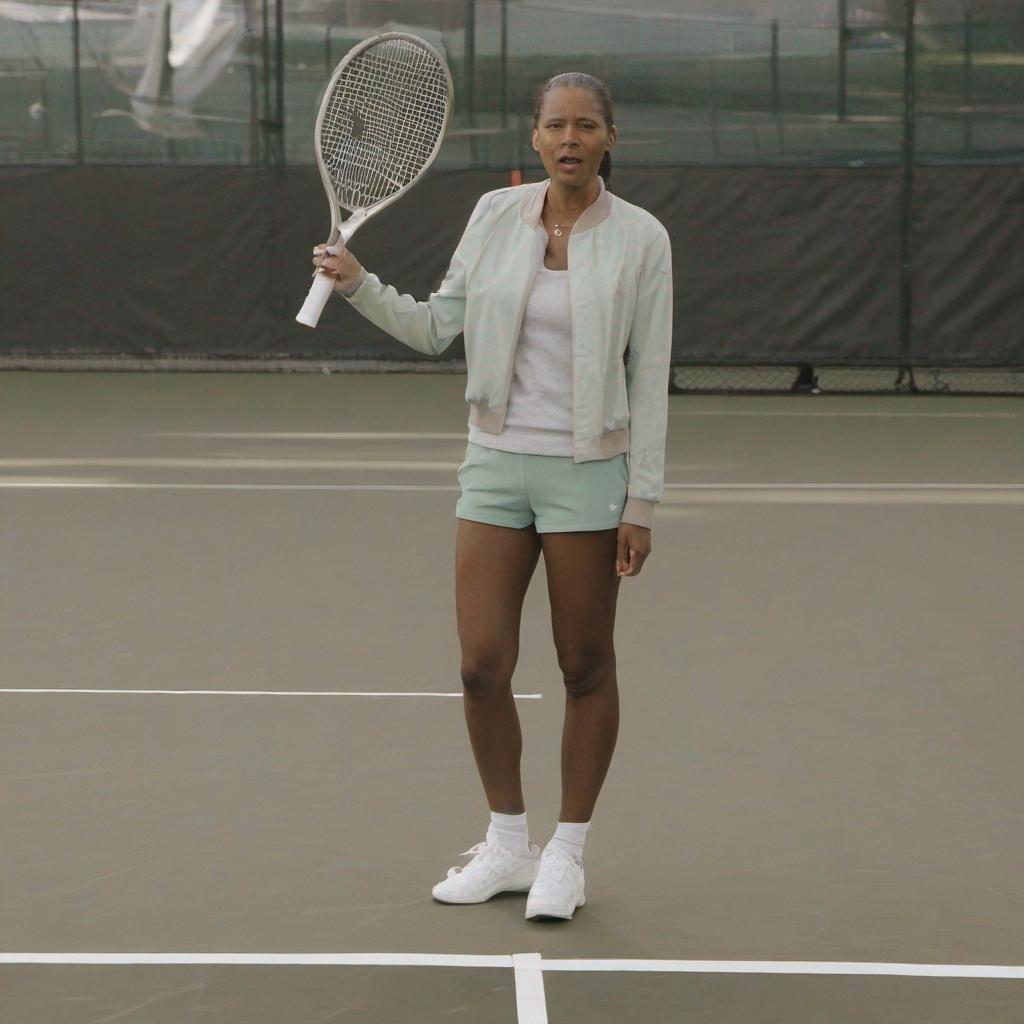} & \includegraphics[width=2cm]{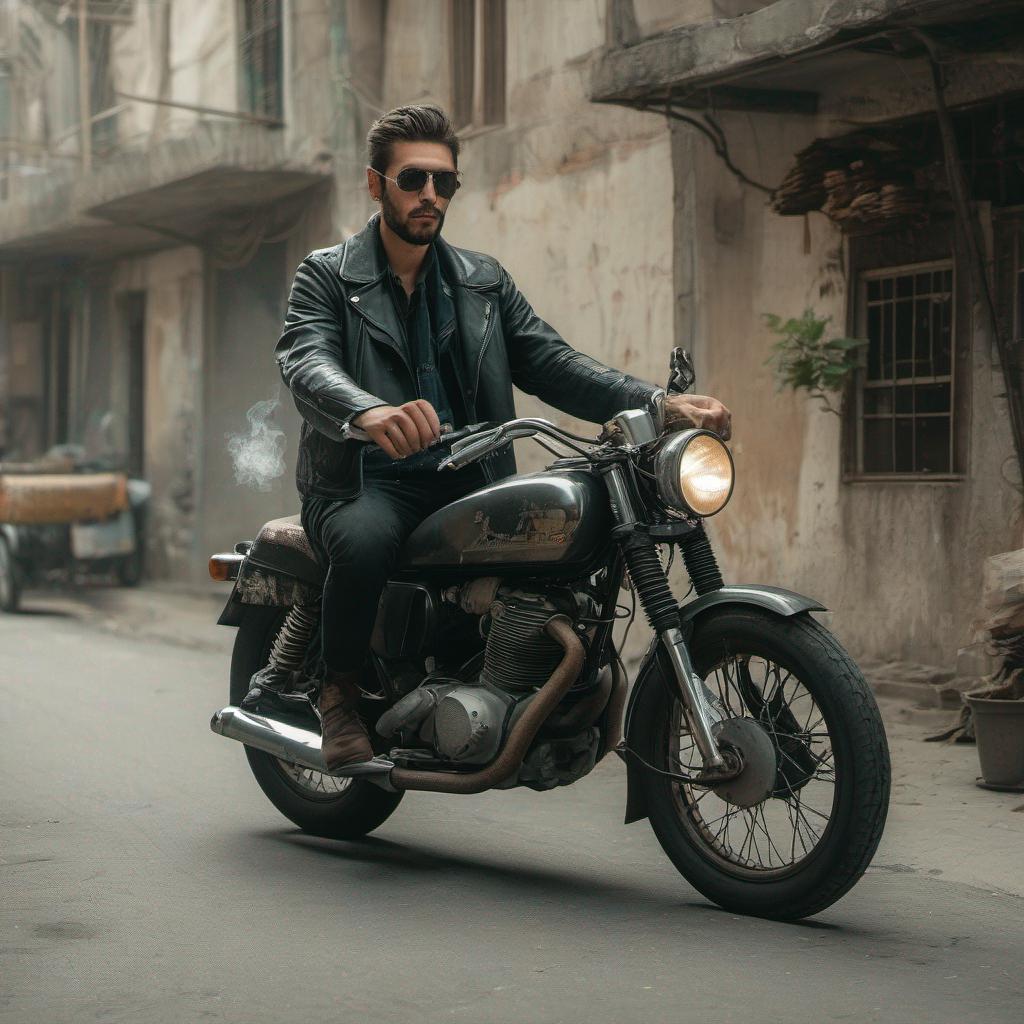} & \includegraphics[width=2cm]{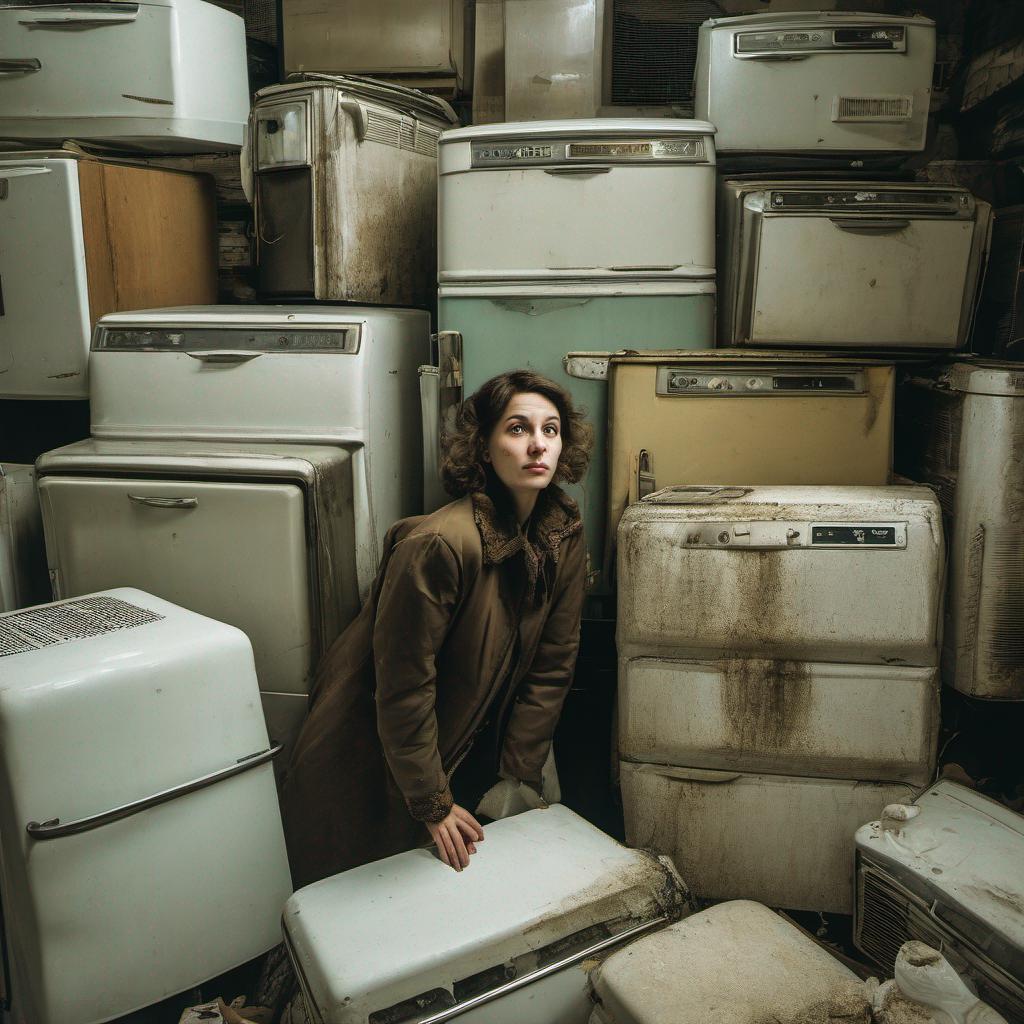} \\
    \midrule  
    -6.09 & -4.71 & -2.54 & 0.31 & 2.33 & 4.83 & 6.18 \\
    \bottomrule
  \end{tabular}
  \caption{Examples of synthetic face images and the corresponding \textit{FS}. 
  We see a positive correlation between the score and the rationality and aesthetic appeal of faces.}
  \label{fig: scores}
\end{figure*}
Apart from the binary ranking performance in Table~\ref{tab:Acc}, we provide more quantitative results of \textit{FS} here. 
In particular, we only compare to the out-performing LocalHPS and the face-oriented SER-FIQ in this part. 
We normalize their scores to $[0, 1]$ individually and report the Pearson correlation (PC) between the score-based rankings and human rankings on the aforementioned human-annotated images as well as the average score of each DM in Table~\ref{tab:global}. 
We note that FS enjoys a decently higher PC compared to other metrics.
Besides, \textit{FS} can showcase subtle differences between RV5.1 and SDXL, aligned with human scorers (see Table~\ref{tab:global}). 
To show the generalization capacity of \textit{FS}, we also evaluate it on more recent DMs, with the results listed in Table~\ref{tab:different fs} for different T2I diffusion models.
The results echo our subjective feeling that the performant Hunyuan~\cite{li2024hunyuanditpowerfulmultiresolutiondiffusion} and Kolors~\cite{kolors} can usually produce images containing human faces with higher user preference. 
In Figure~\ref{fig: scores}, we illustrate some randomly selected face images generated by SDXL and the corresponding \textit{FS}s, which implies that rationality and aesthetic appeal of faces are positively correlated with \textit{FS}. 

\section{Enhancing Face Quality based on \textit{FS}}
We can naturally leverage \textit{FS} for preference learning to equip pre-trained DMs with better face generation quality.
Here, we conduct an initial study with DPO~\cite{wallace2024diffusion} and clarify that other algorithmic choices are compatible.
To perform DPO, we collect 400 prompts related to humans from the MS-COCO validation dataset, and for each prompt, we generate 50 images and utilize \textit{FS} for scoring each image. 
This way, we obtain a set of on-policy sample pairs characterizing preference on face quality, which we call FS-DPO.
Letting $\mathcal{D}_{p}$ denote the preference dataset, the DPO loss for fine-tuning DMs~\cite{wallace2024diffusion} takes the form
\begin{equation}
\begin{aligned}
    &L_{DPO} = -\mathrm{E}_{(x_0^w,x_0^l)\sim \mathcal{D}_{p},t\sim\mathcal{U}(0,T),x_t^w\sim q(x_t^w|x_0^w),x_t^l\sim q(x_t^l|x_0^l)} \\
    &\log \sigma(-\beta T(||\epsilon^w-\epsilon_{\theta}(x_t^w,t,c)||_2^2-||\epsilon^w-\epsilon_{ref}(x_t^w,t,c)||_2^2 \\
    &-(||\epsilon^l-\epsilon_{\theta}(x_t^l,t,c)||_2^2-||\epsilon^l-\epsilon_{ref}(x_t^l,t,c)||_2^2))),
\end{aligned}
\end{equation}
where $x_t^*=\alpha_t x_0^*+\sigma_t\epsilon^*,\epsilon^*\sim\mathcal{N}(0, I),*\in \{w,l\}$, and the hyperparameter $\beta$ controls the strength of regularization.

\begin{table}[t]
    \centering
    \setlength{\tabcolsep}{1mm}
    \begin{tabular}{lrccccc}
    \toprule
     Models && Base &FaceLoRA& Neg & DPO & FS-DPO \\
     \midrule
    PS && 0.222&0.220 &0.222&0.227&0.225\\
    IR        &&0.767&0.612&0.669&1.031&0.913\\
    HPS       &&0.291&0.291&0.299&0.311&0.304\\
    FS        &&2.210&2.354&2.476&2.595&\textbf{4.084}\\
    \bottomrule
    \end{tabular}
    \caption{Quantitative comparisons between our fine-tuned model and baselines.}
    \label{tab:baselines}
\end{table}
\begin{figure}[t]
    \centering
    \includegraphics[width=1\linewidth]{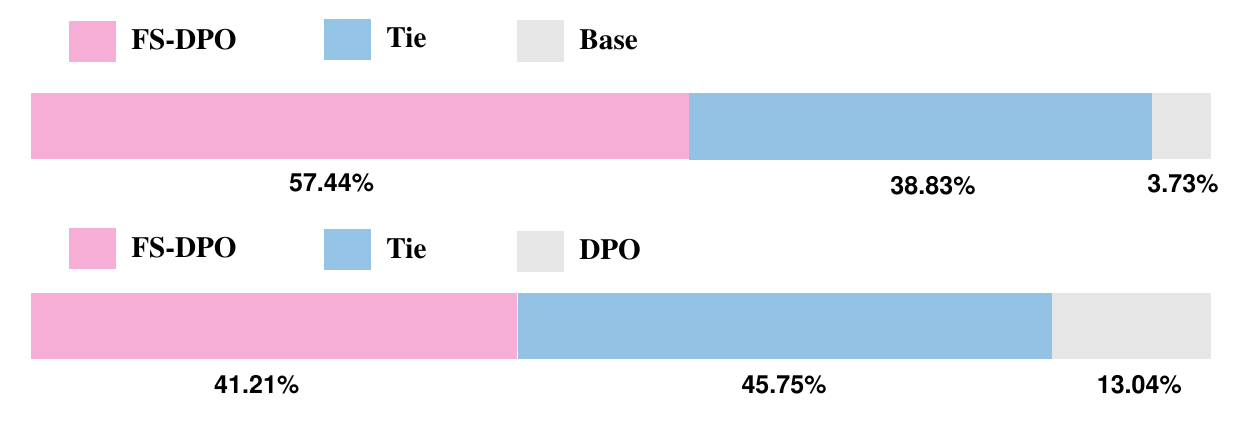}
    \caption{Human evaluation on face quality between our fine-tuned model and the SDXL-Base/SDXL-DPO.}
    \label{fig:human eval}
\end{figure}

\subsubsection{Implementation Details.}
The preference dataset contains roughly 20k images.
We perform LoRA training~\cite{hulora} with the learning rate $6\cdot 10^{-6}$, batch size 8 and gradient accumulation 2. 
We assess our method's effectiveness through quantitative and qualitative analysis.

\subsubsection{Baselines and Evaluation.}
 For baselines, we choose all the backbone as SDXL due to its wider acceptance, including the original SDXL (SDXL-Base), SDXL with a LoRA for real faces\footnote{https://civitai.com/models/232746/real-humans} (SDXL-FaceLoRA), SDXL with negative prompts (SDXL-Neg) and  Diffusion-DPO~\cite{wallace2024diffusion} (SDXL-DPO).
No extra inpainting processes for post-hoc improvement are performed. 
We set the negative prompt as previously mentioned.
To evaluate face quality, we sample 1k prompts from HumanArt~\cite{ju2023human} and report the average \textit{FS}. 
For the evaluation of general generation capability, we also leverage the HPSv2 evaluation dataset containing 3.2k prompts~\cite{wu2023humanpreferencescorev2} and report PickScore~\cite{kirstain2023pick}, IR, and HPS for comparisons.

\subsubsection{Main Results.}
\begin{figure*}[t]
    \centering
    \parbox{0.18\linewidth}{\centering SDXL-Base}
    \parbox{0.18\linewidth}{\centering SDXL-FaceLoRA}
    \parbox{0.18\linewidth}{\centering SDXL-N}
    \parbox{0.18\linewidth}{\centering SDXL-DPO}
    \parbox{0.18\linewidth}{\centering SDXL-FS-DPO}\\
    \includegraphics[width=0.18\linewidth]{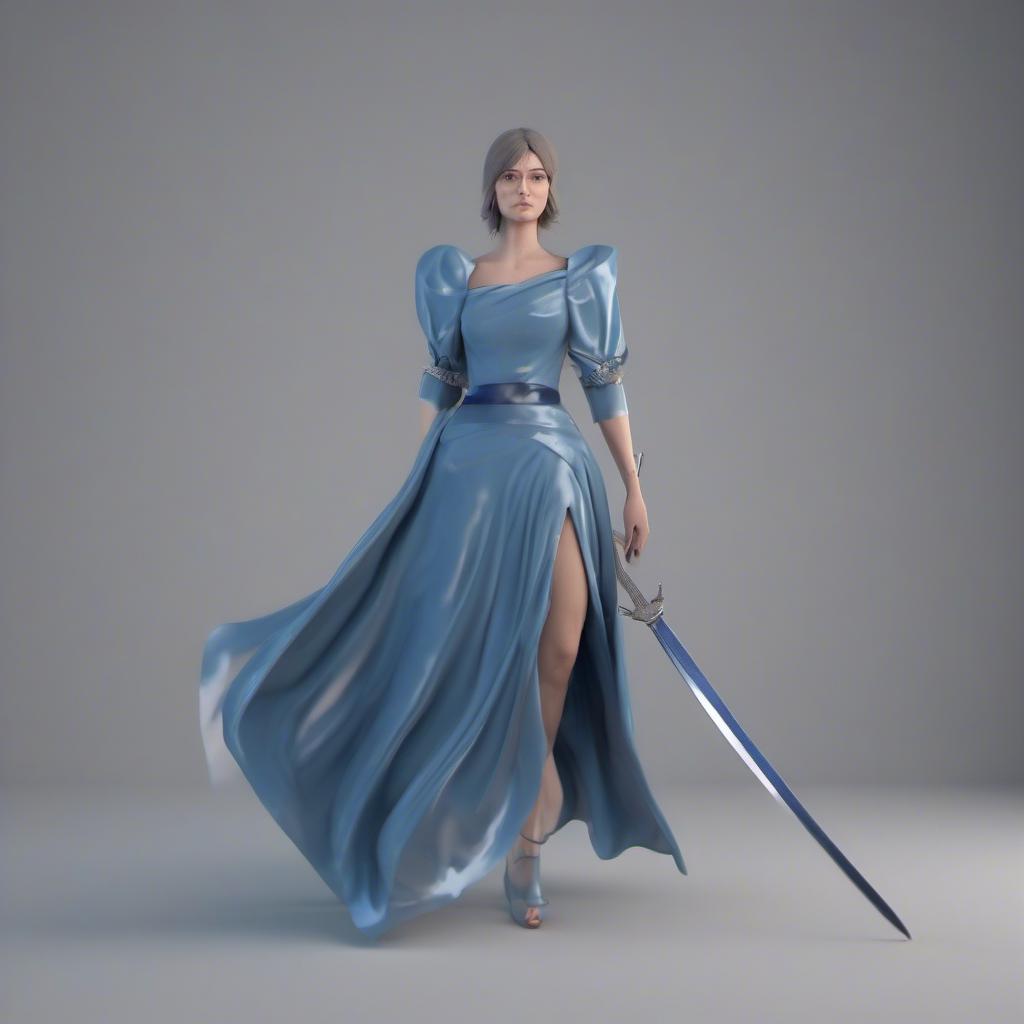}
    \includegraphics[width=0.18\linewidth]{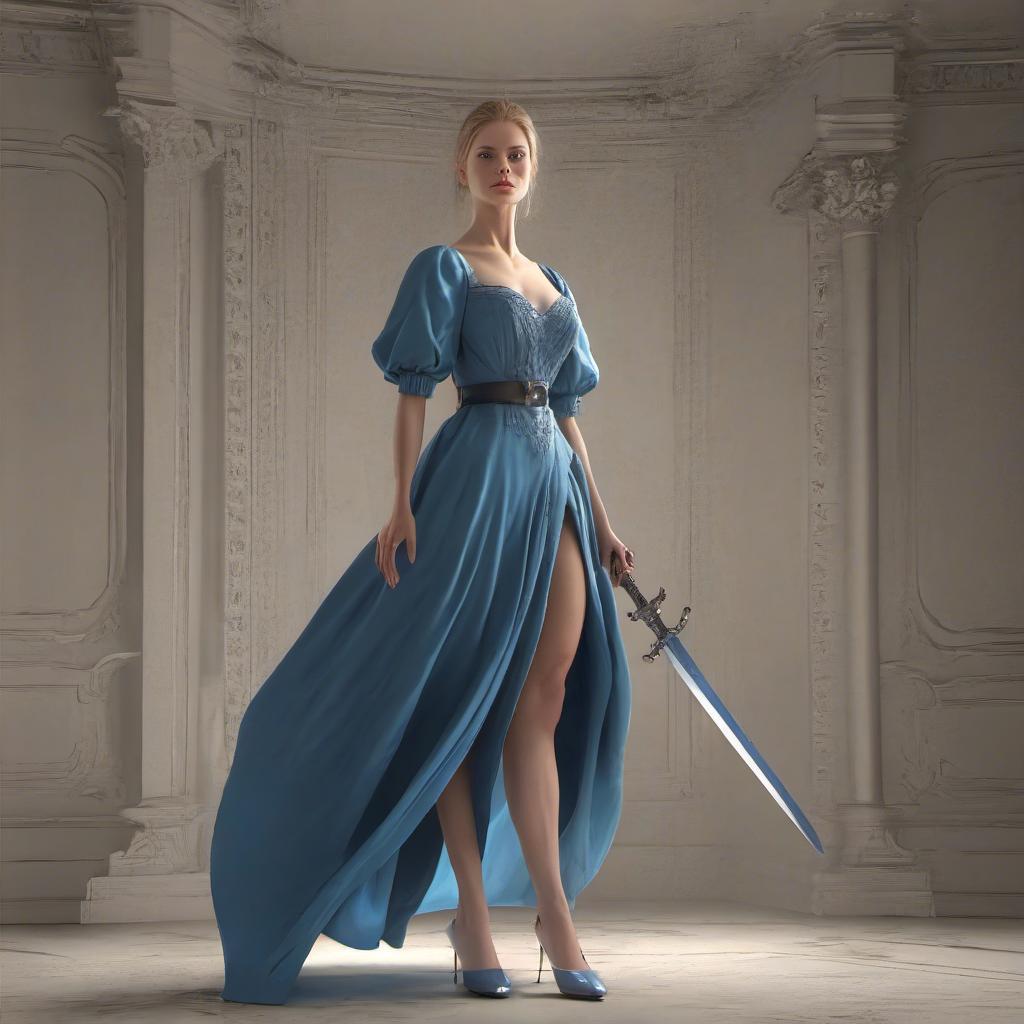}
    \includegraphics[width=0.18\linewidth]{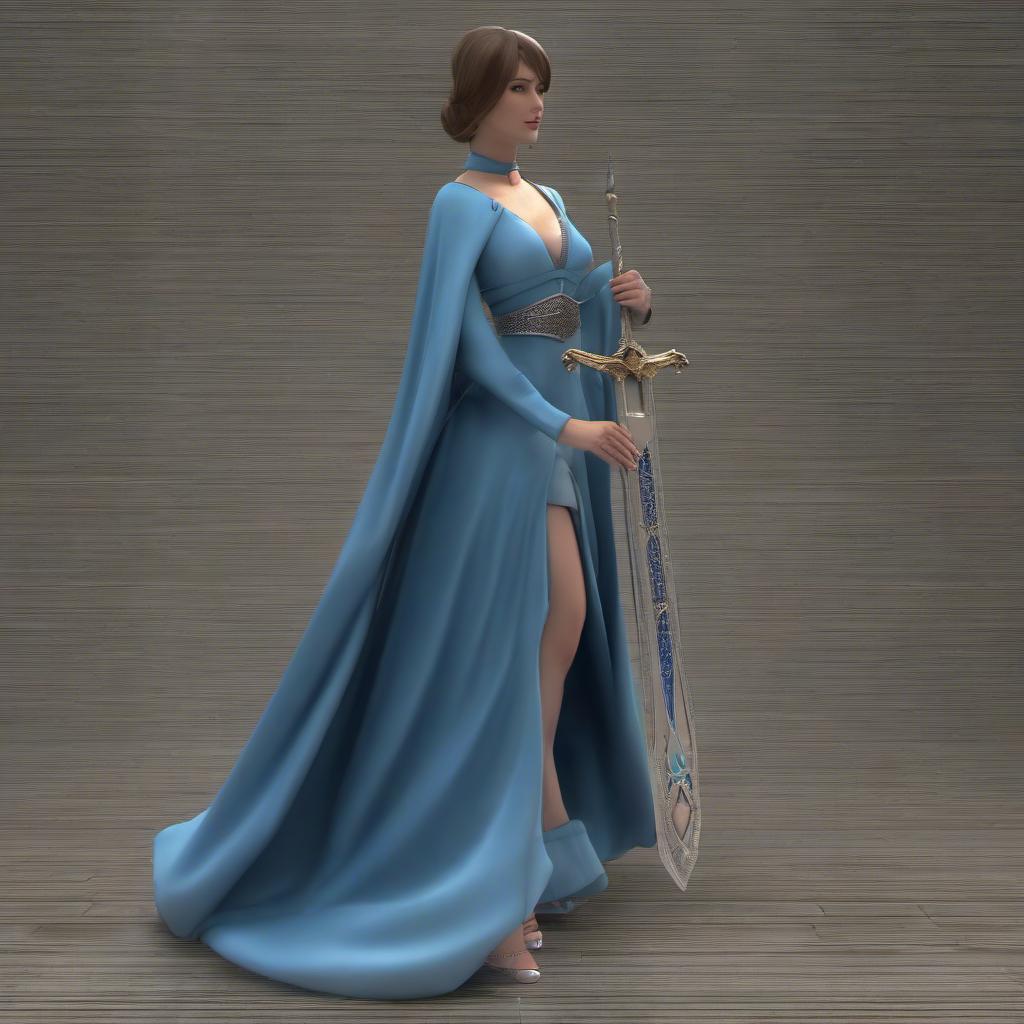}
    \includegraphics[width=0.18\linewidth]{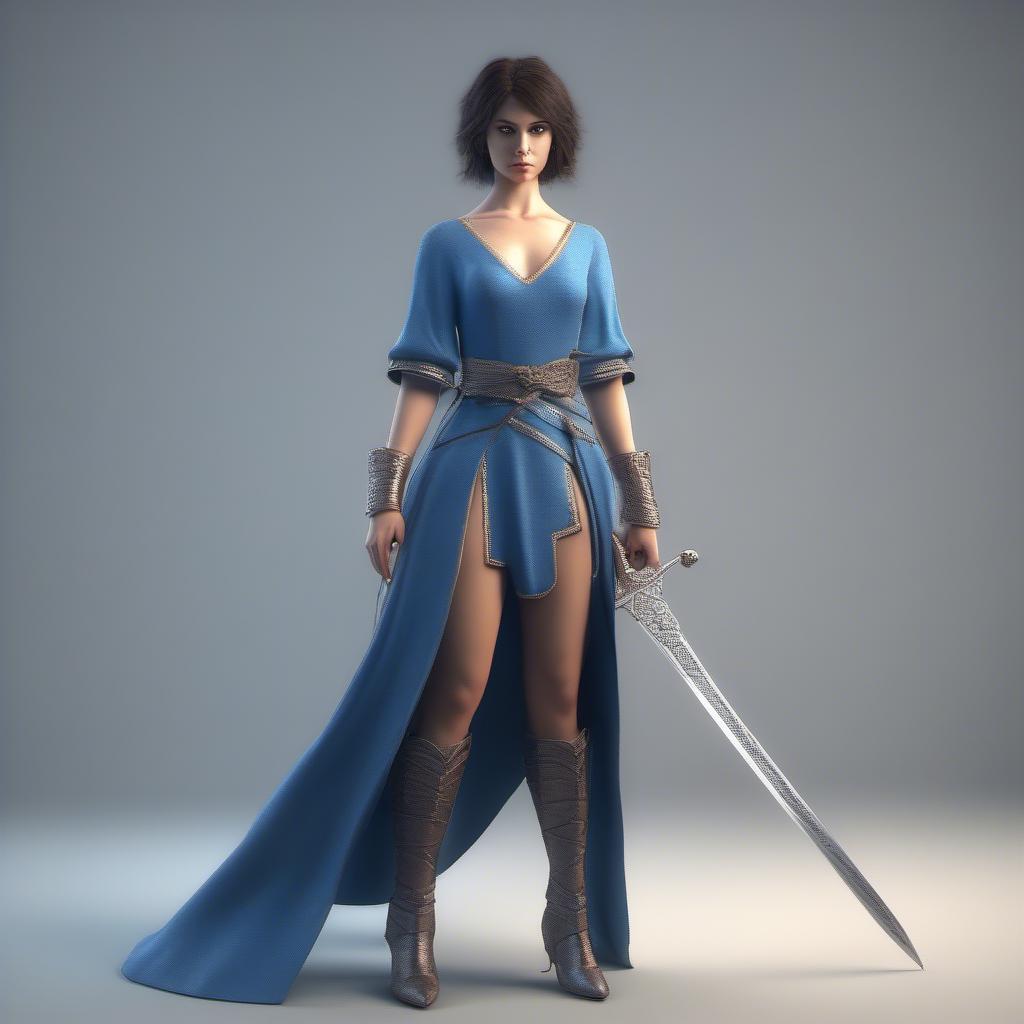}
    \includegraphics[width=0.18\linewidth]{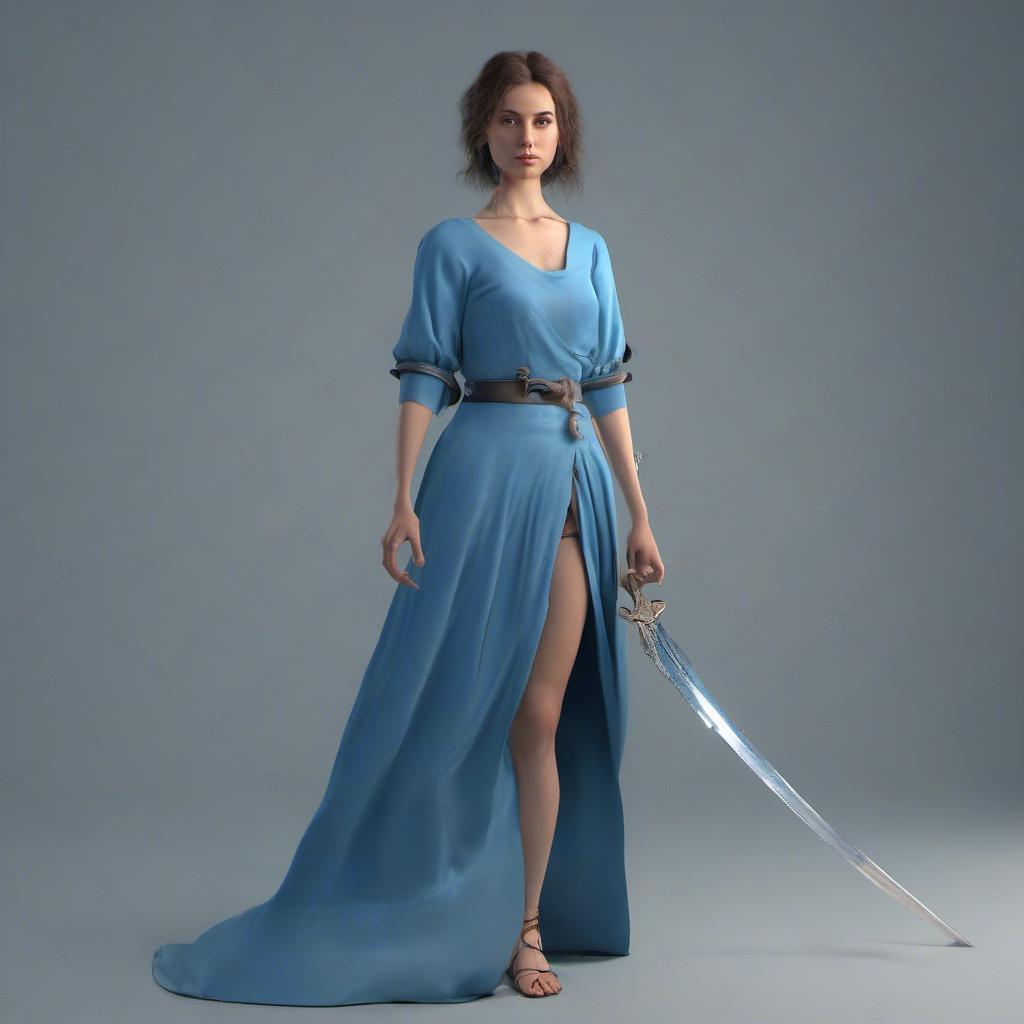}
    \parbox{1\linewidth}{\centering A 3d rendered image of a woman in a blue dress holding a sword.}
    \includegraphics[width=0.18\linewidth]{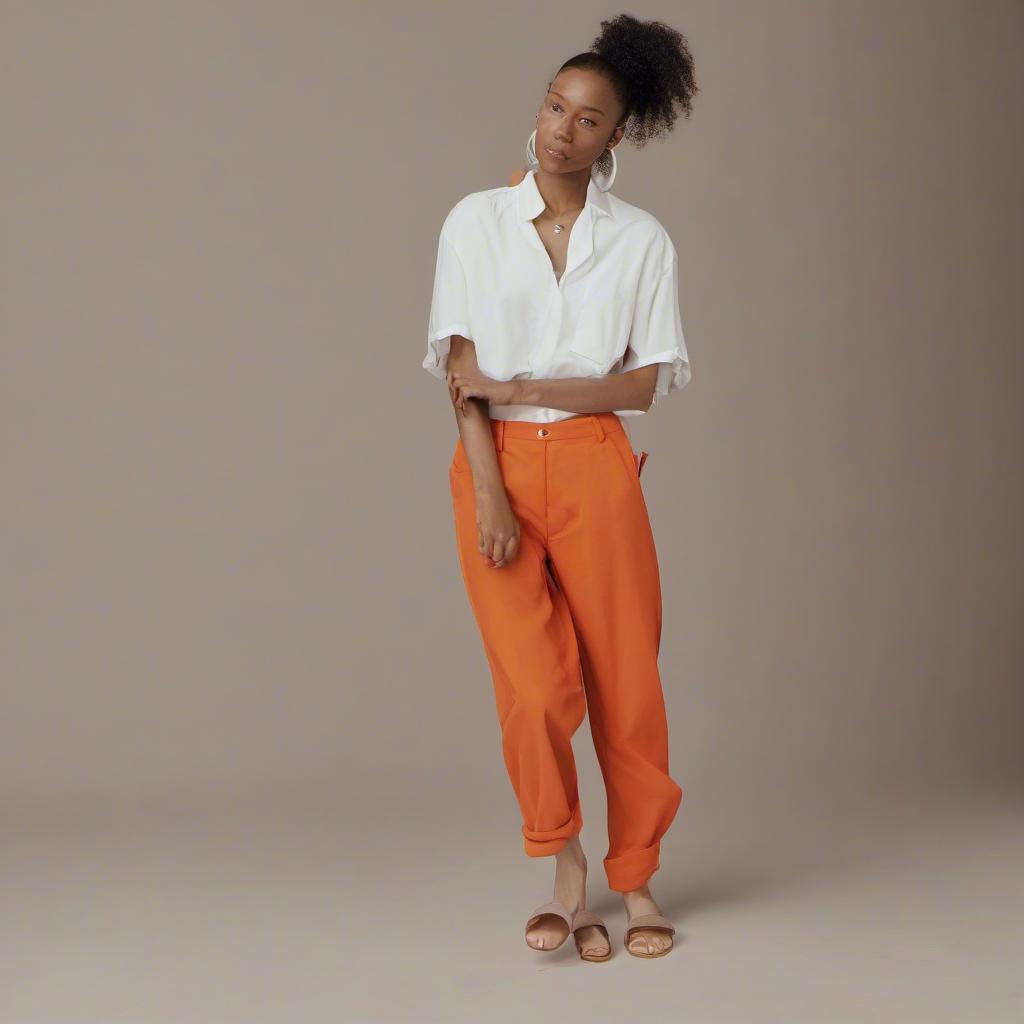}
    \includegraphics[width=0.18\linewidth]{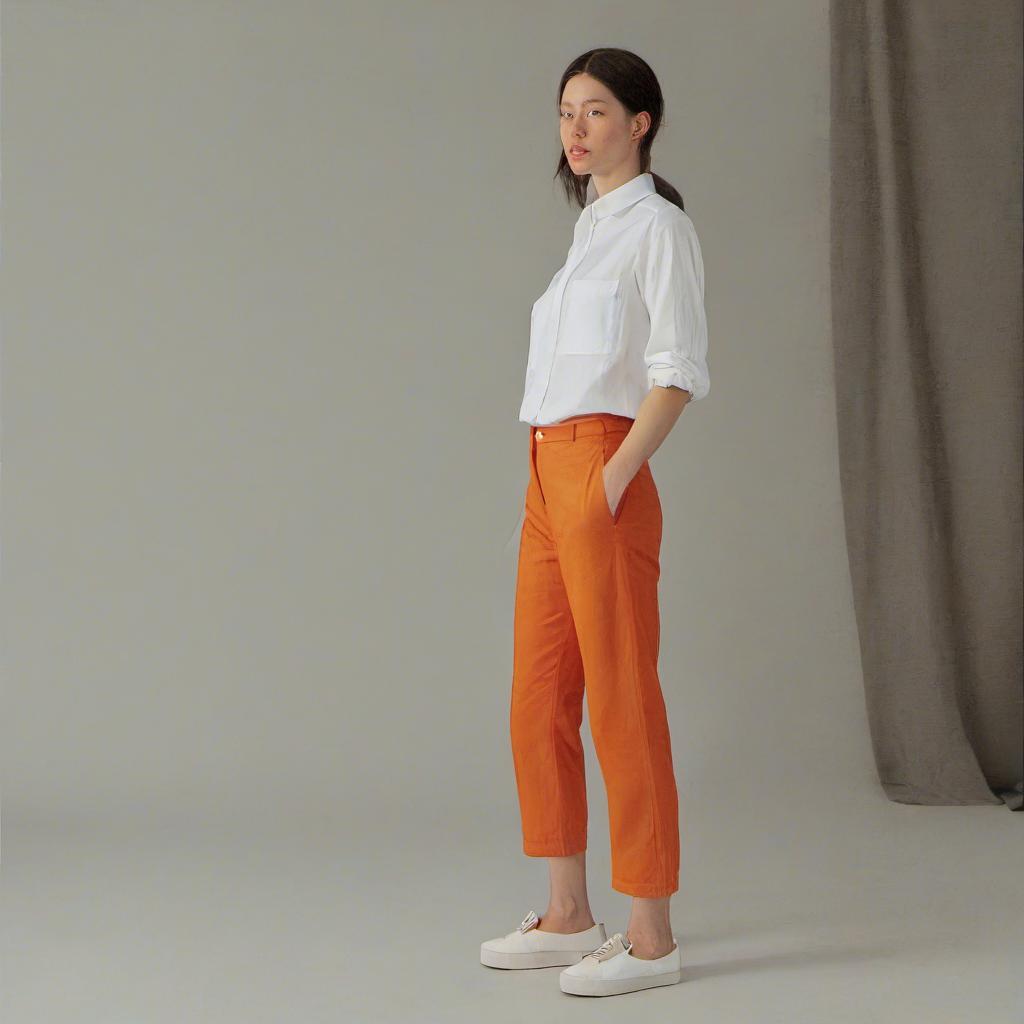}
    \includegraphics[width=0.18\linewidth]{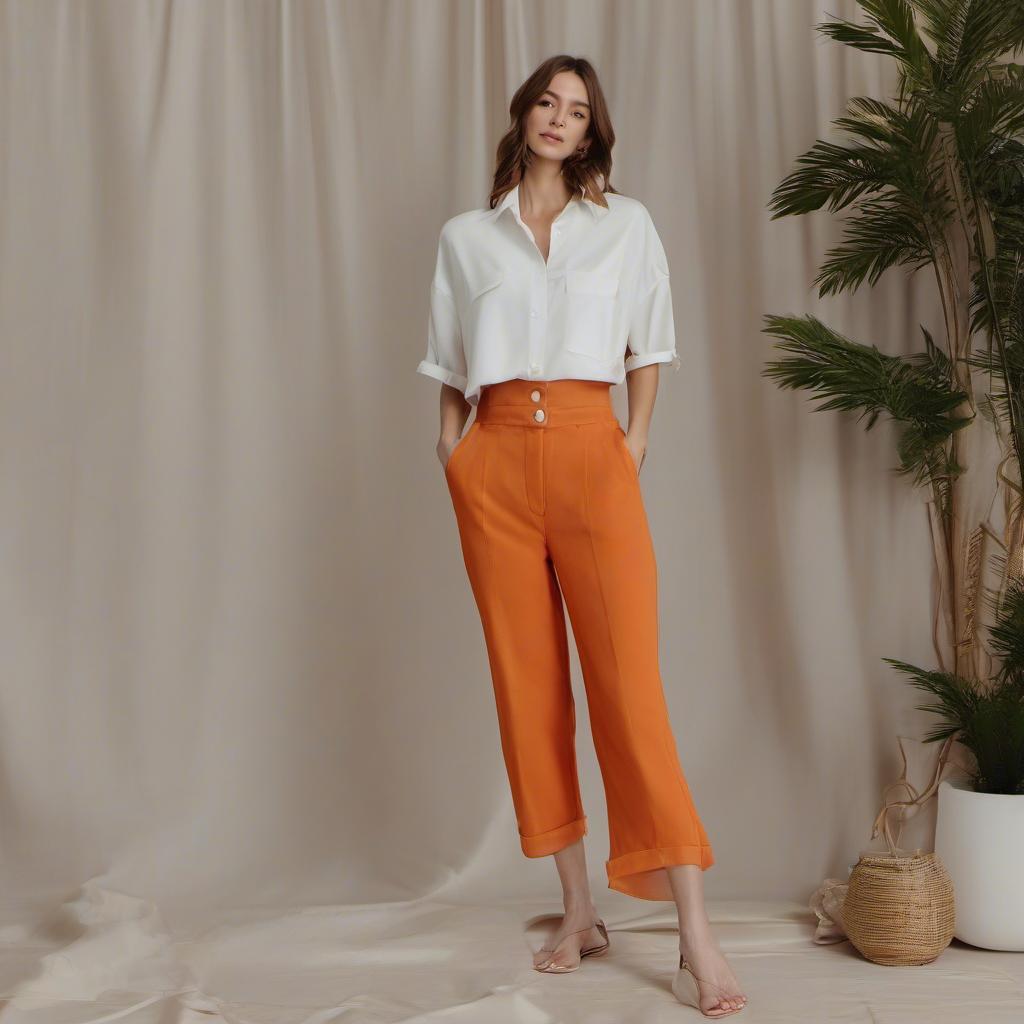}
    \includegraphics[width=0.18\linewidth]{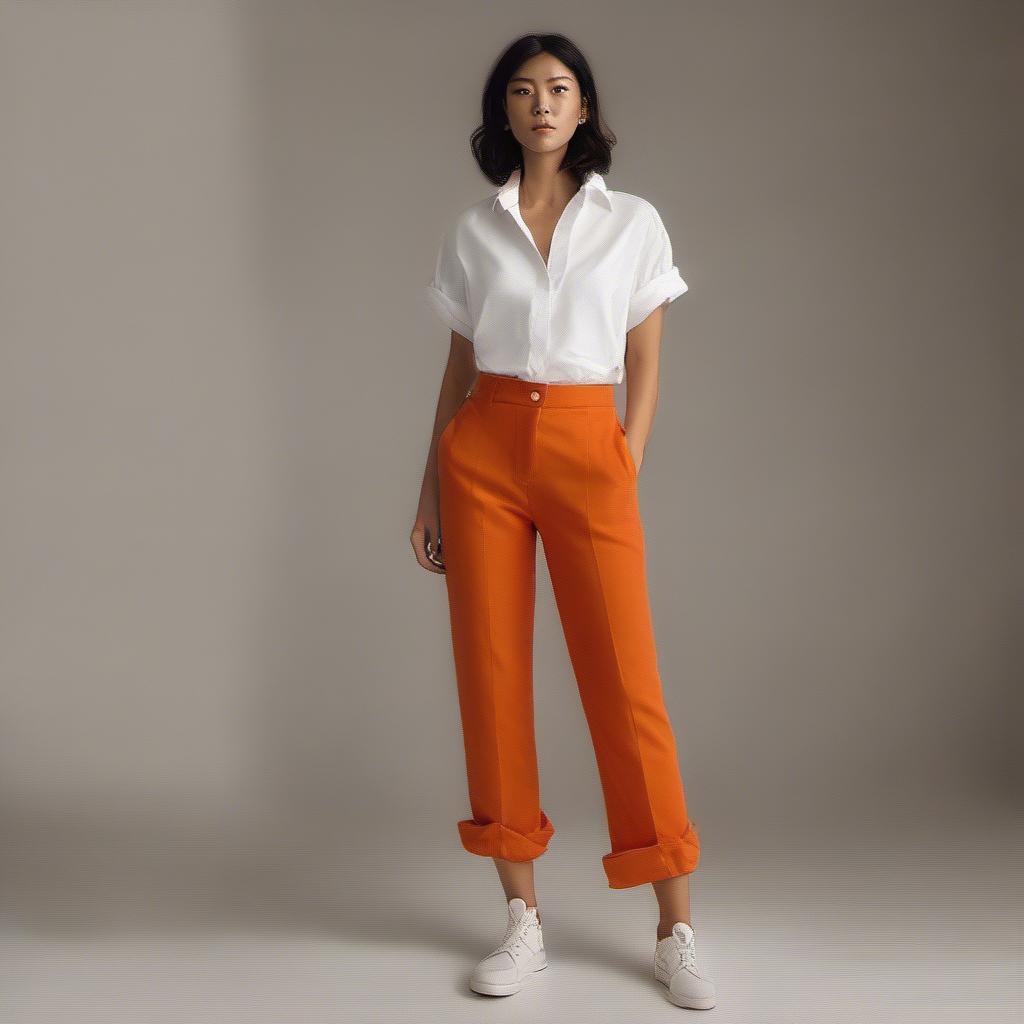}
    \includegraphics[width=0.18\linewidth]{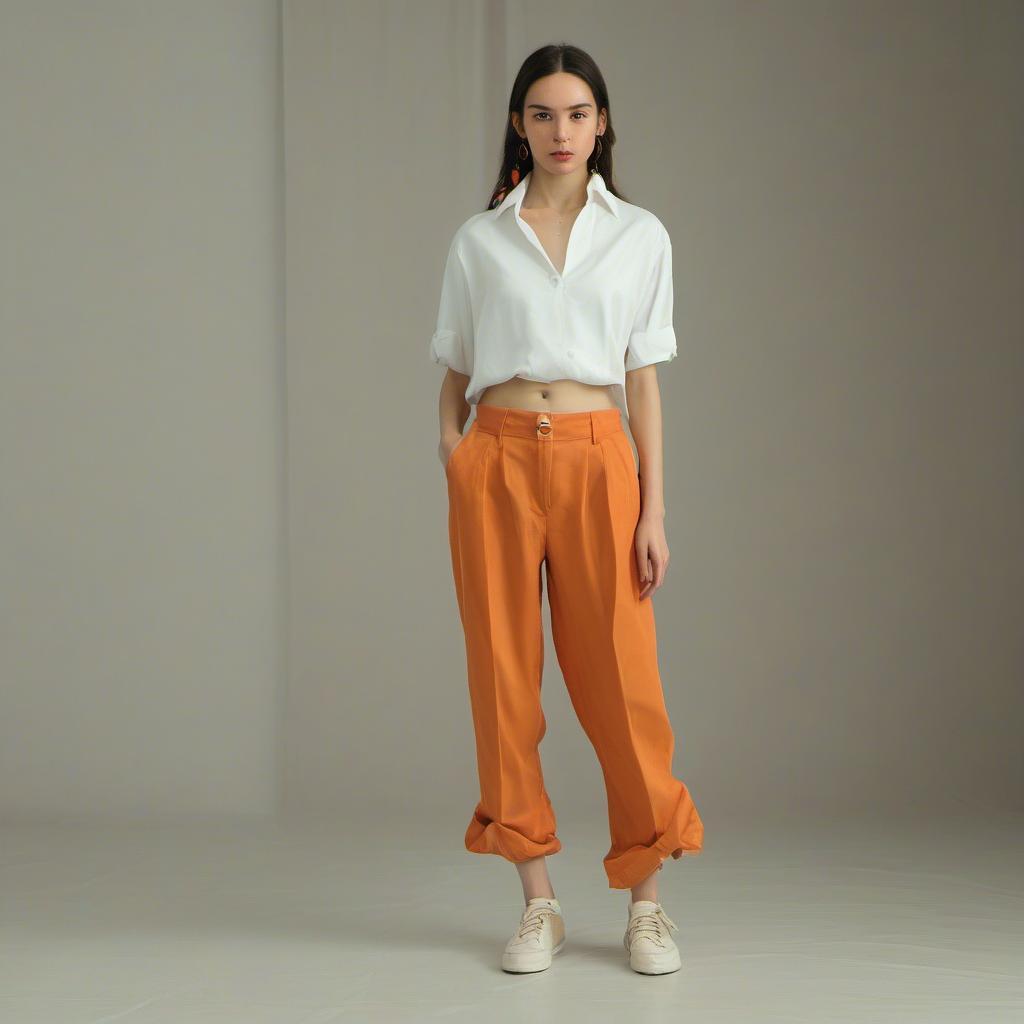}
    \parbox{1\linewidth}{\centering A woman in orange pants and a white shirt.}
    \includegraphics[width=0.18\linewidth]{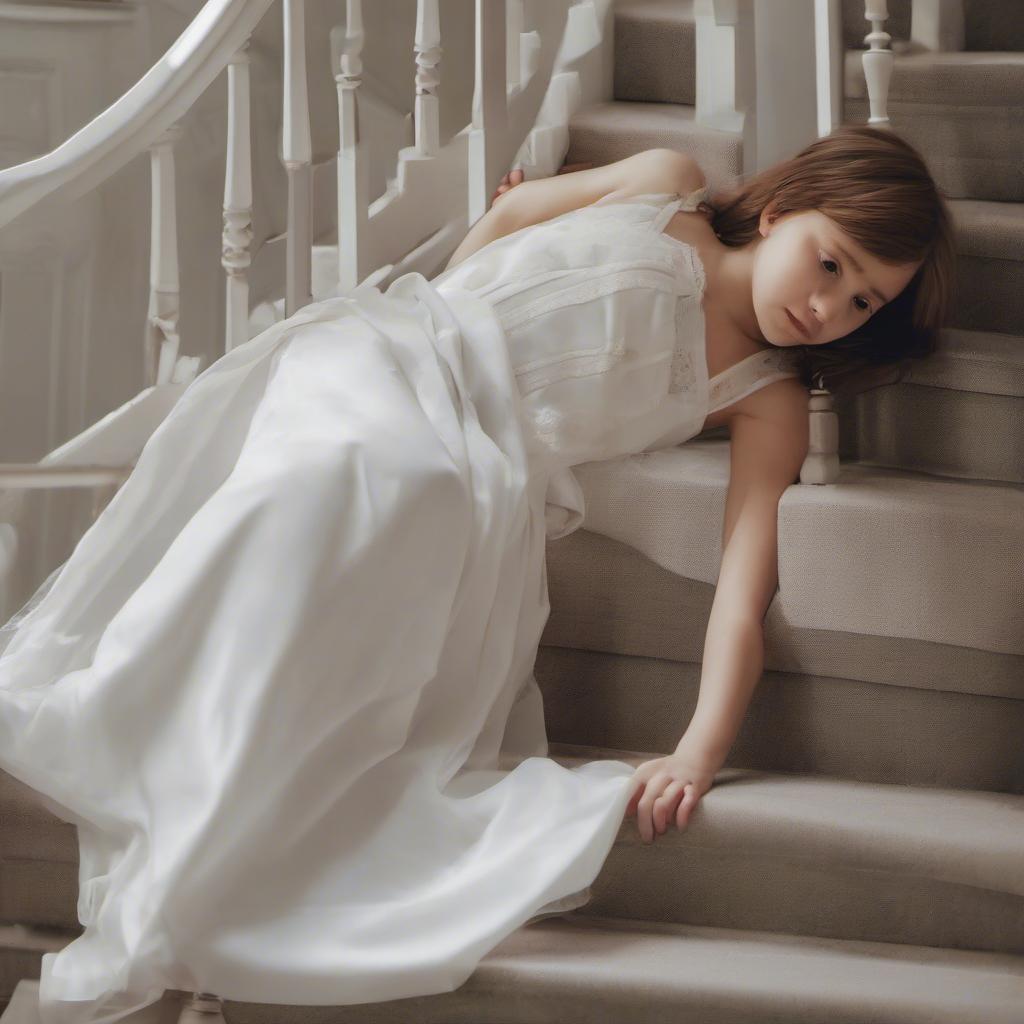}
    \includegraphics[width=0.18\linewidth]{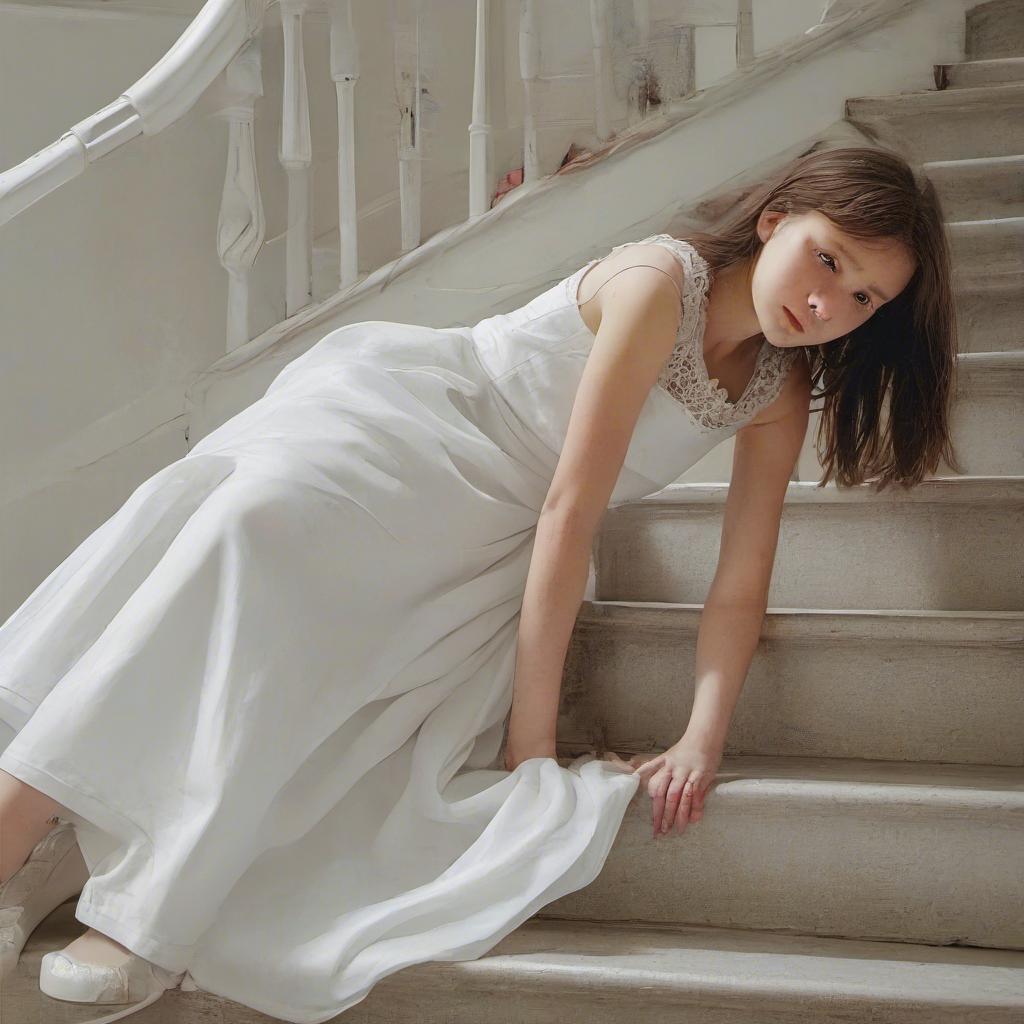}
    \includegraphics[width=0.18\linewidth]{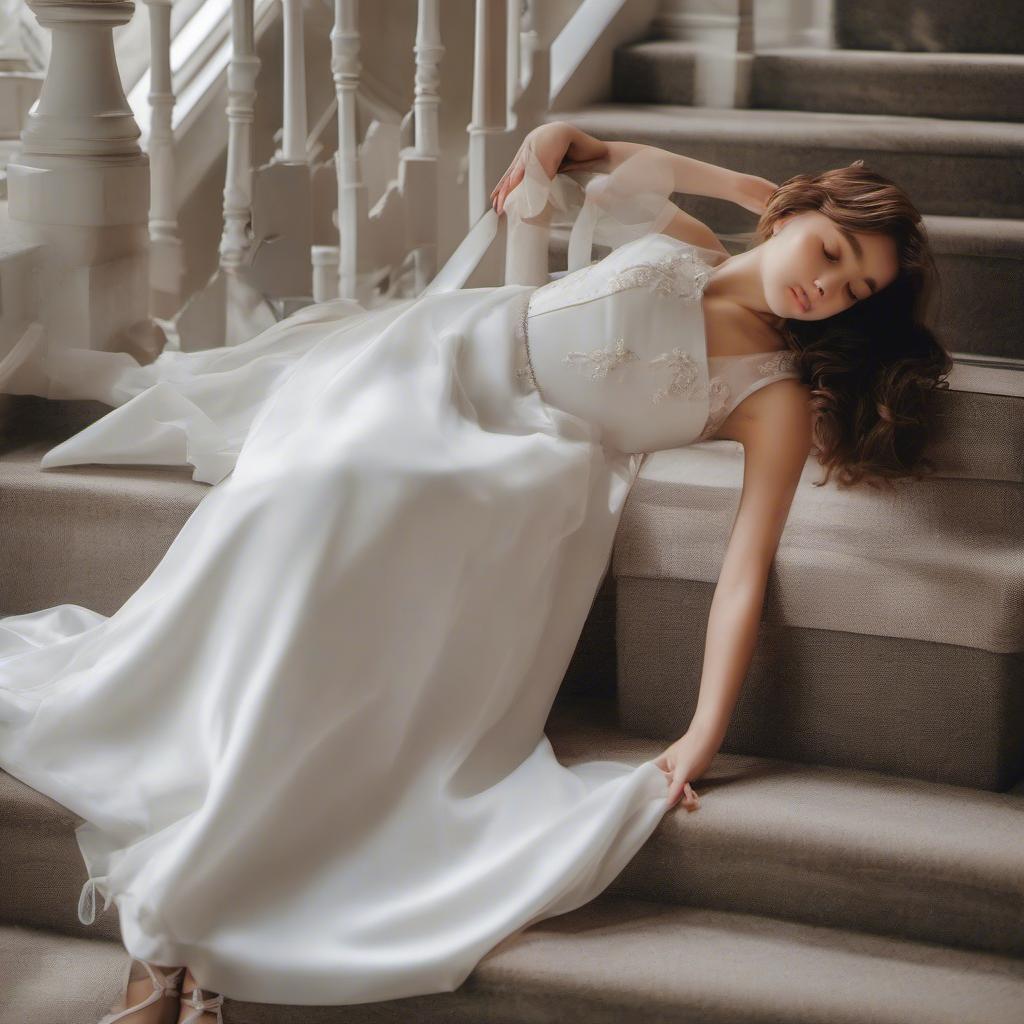}
    \includegraphics[width=0.18\linewidth]{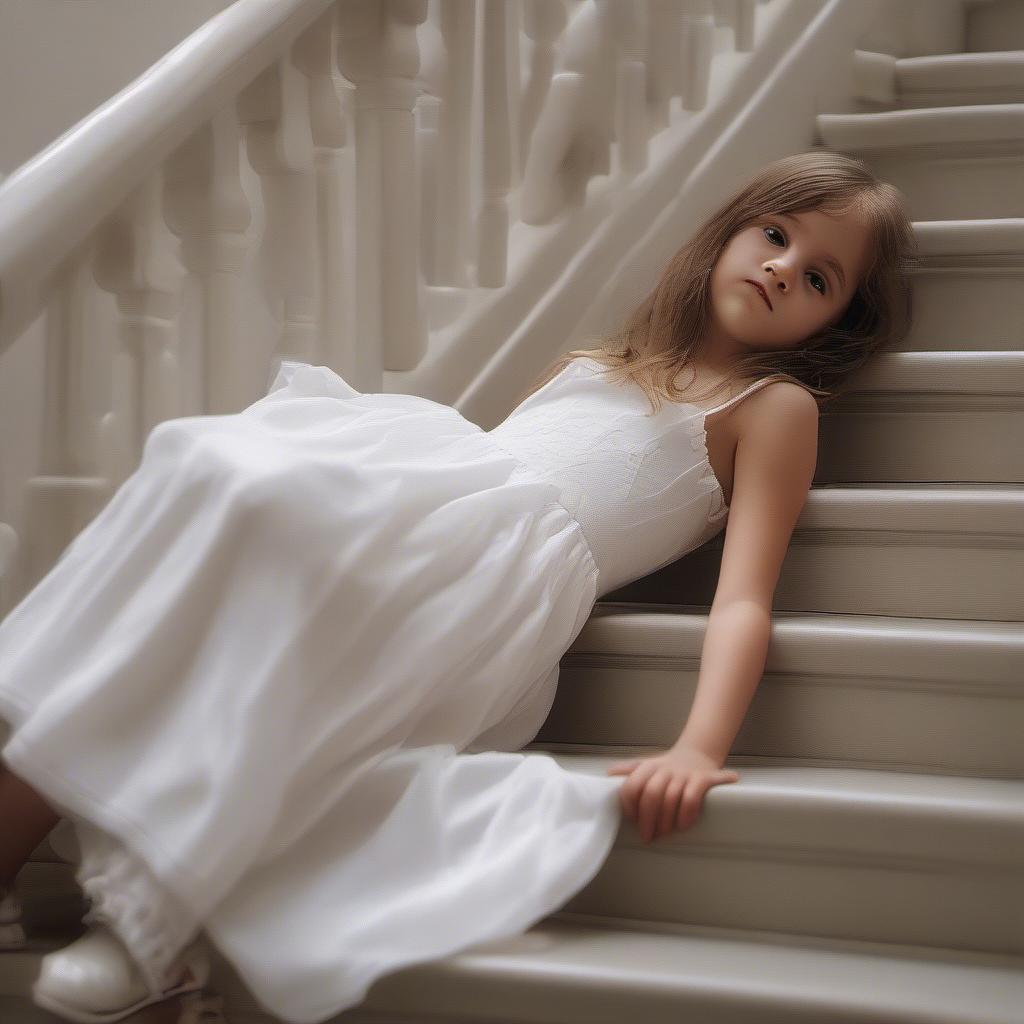}
    \includegraphics[width=0.18\linewidth]{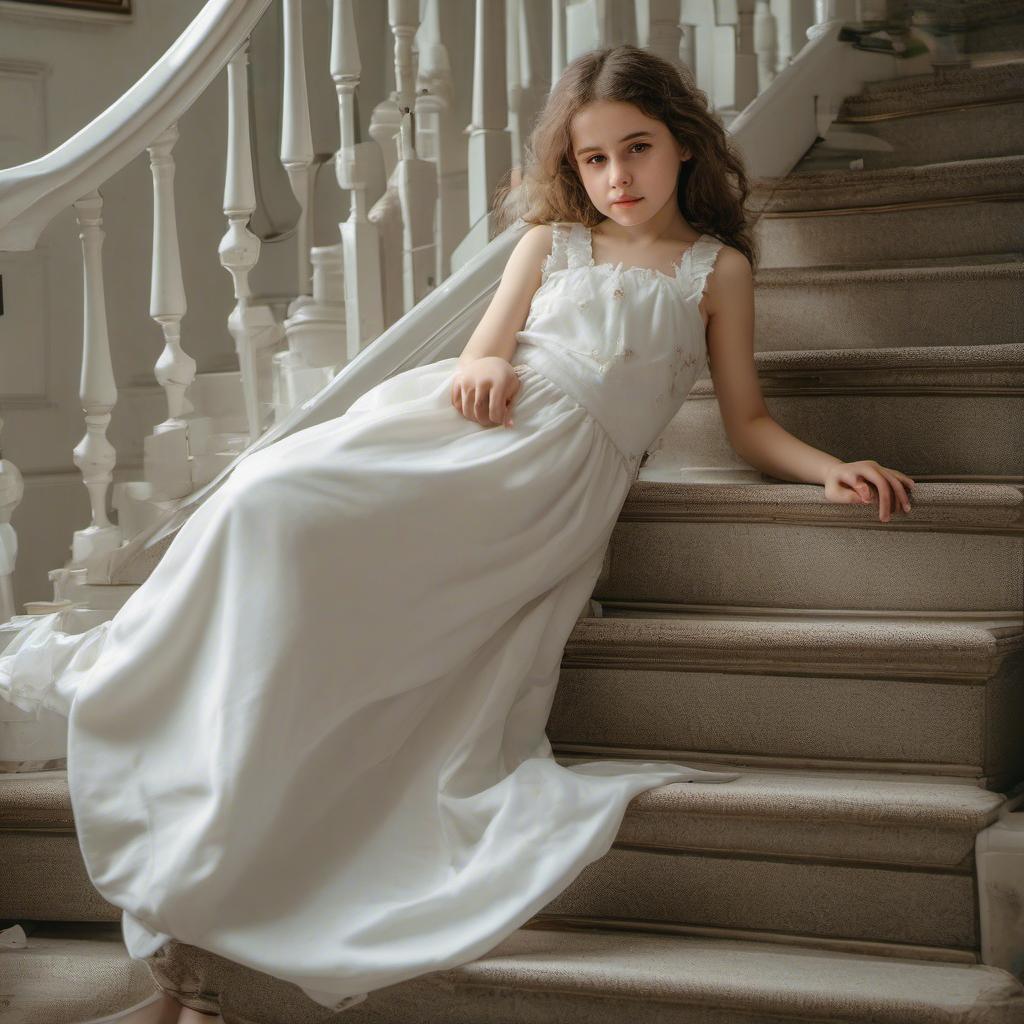}
    \parbox{1\linewidth}{\centering A young girl is laying on the stairs in a white dress.}
    \includegraphics[width=0.18\linewidth]{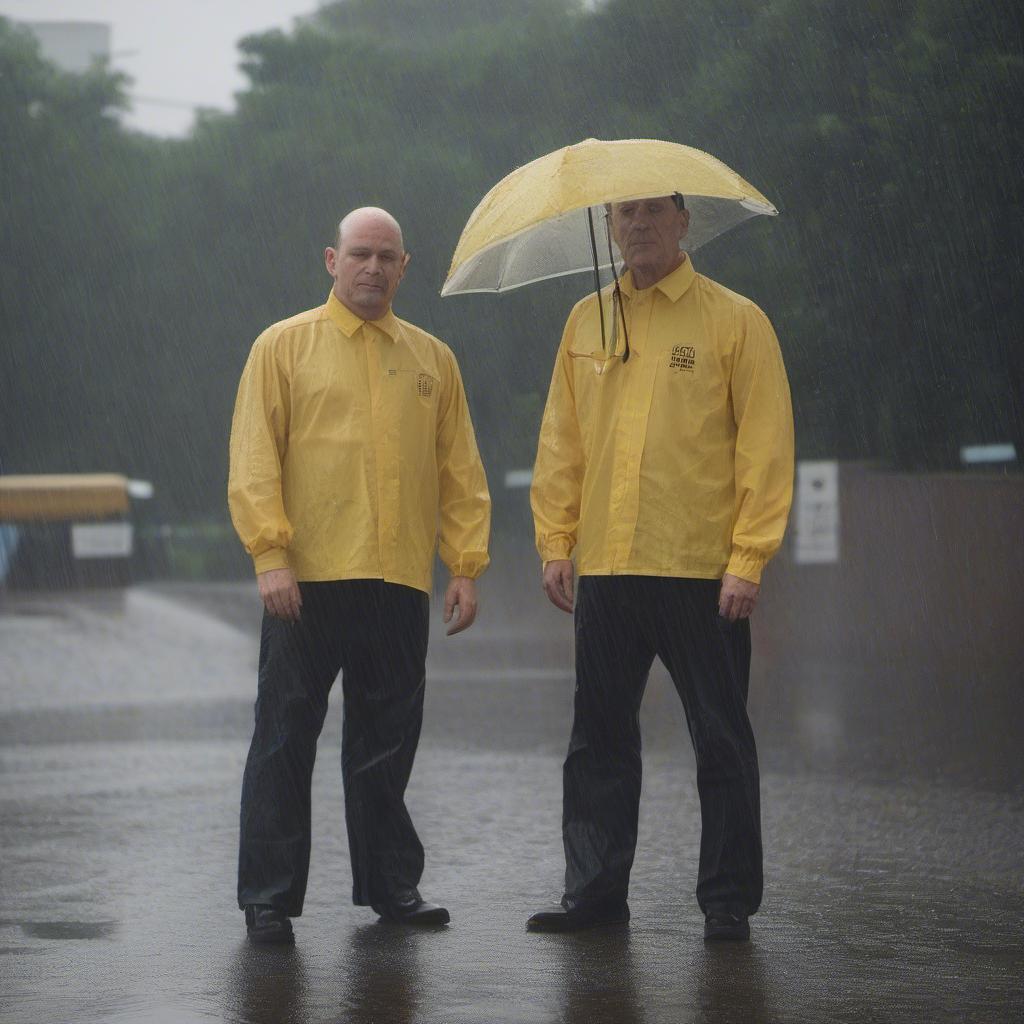}
    \includegraphics[width=0.18\linewidth]{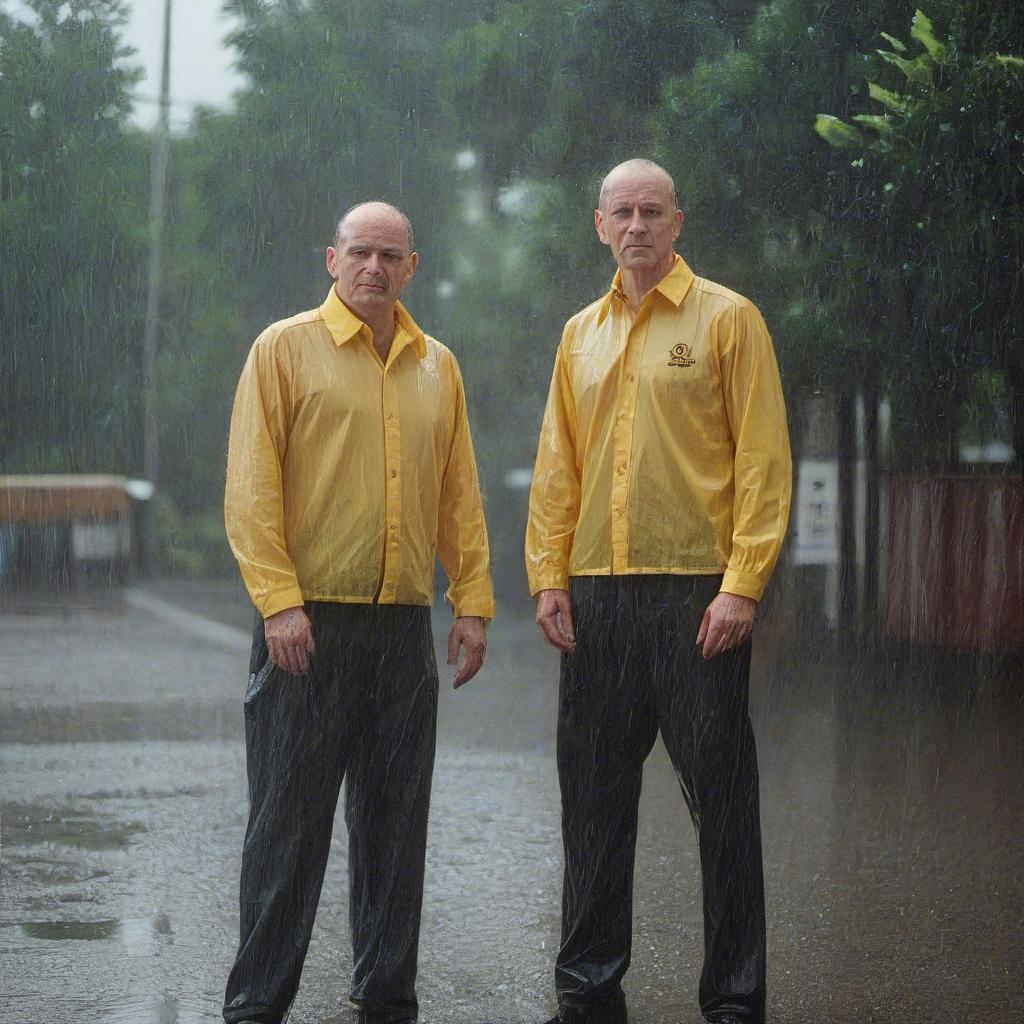}
    \includegraphics[width=0.18\linewidth]{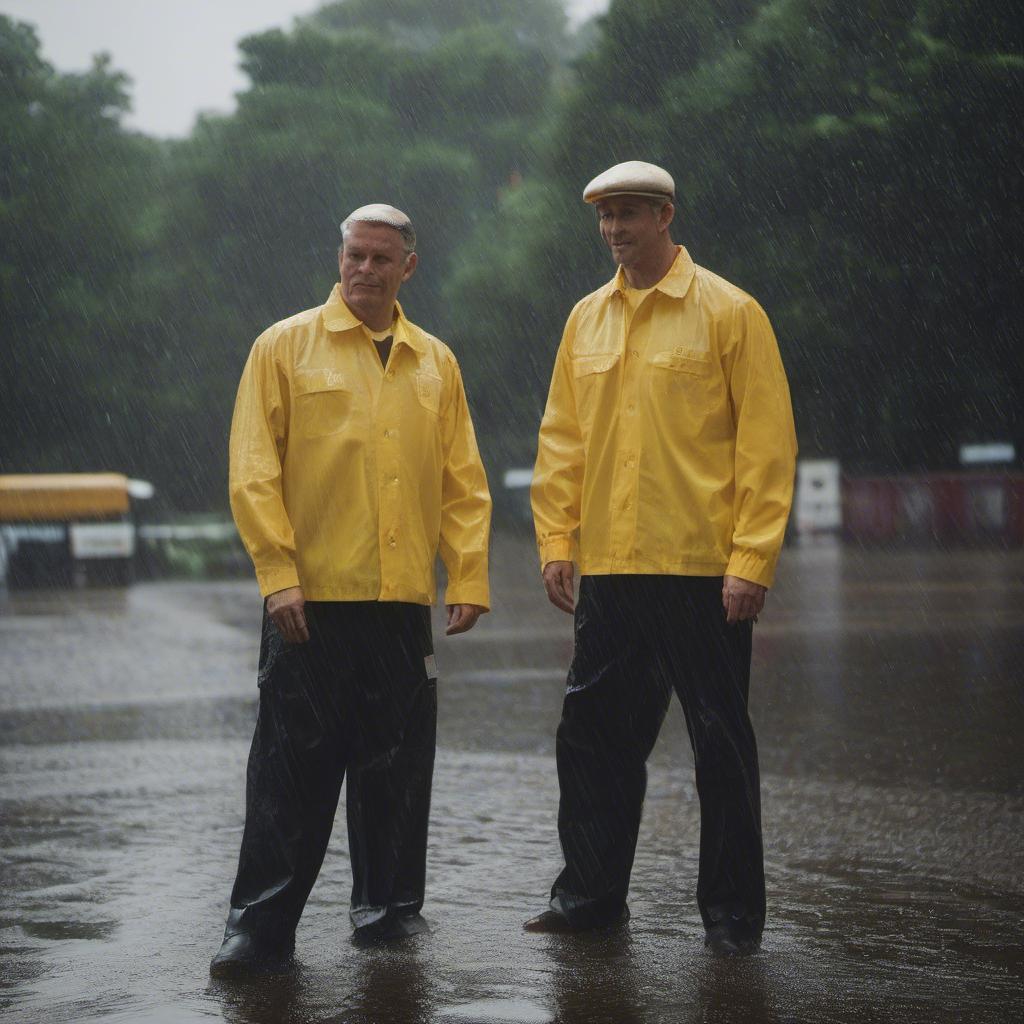}
    \includegraphics[width=0.18\linewidth]{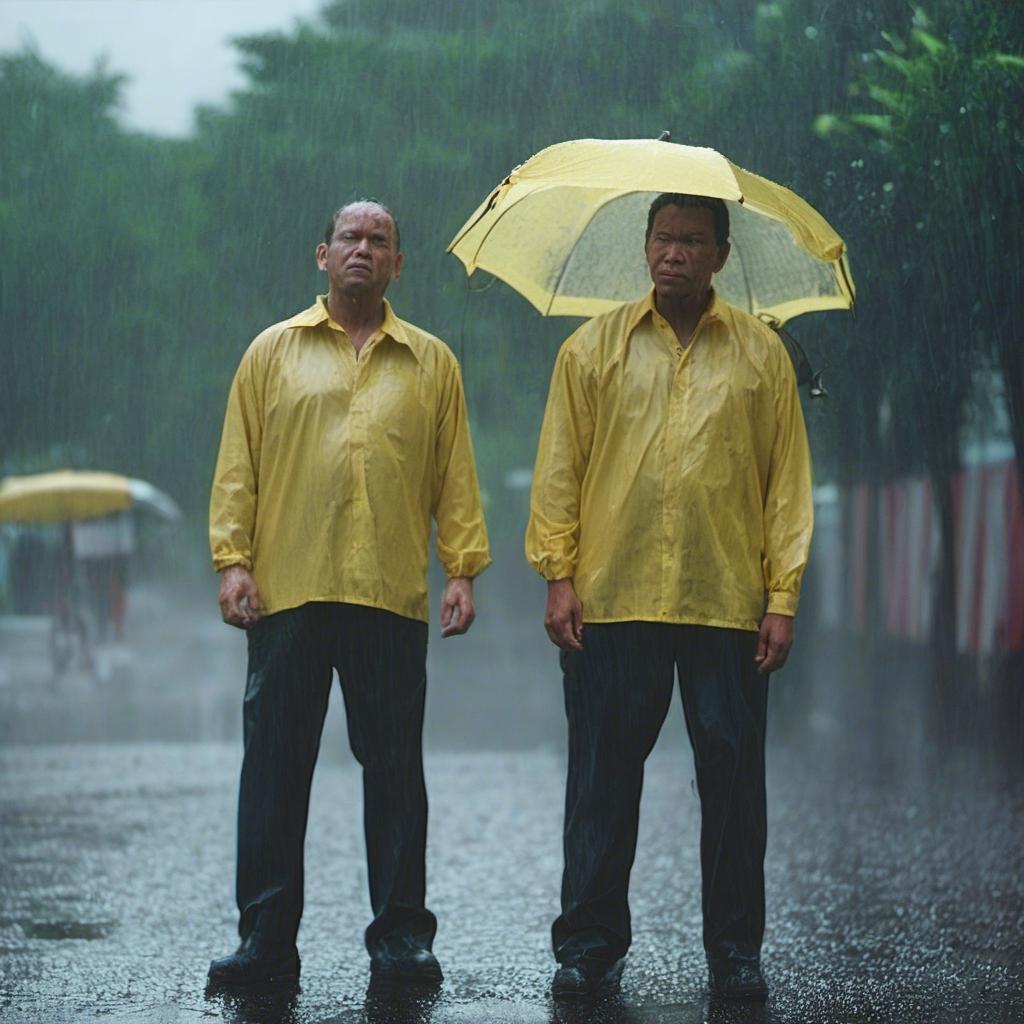}
    \includegraphics[width=0.18\linewidth]{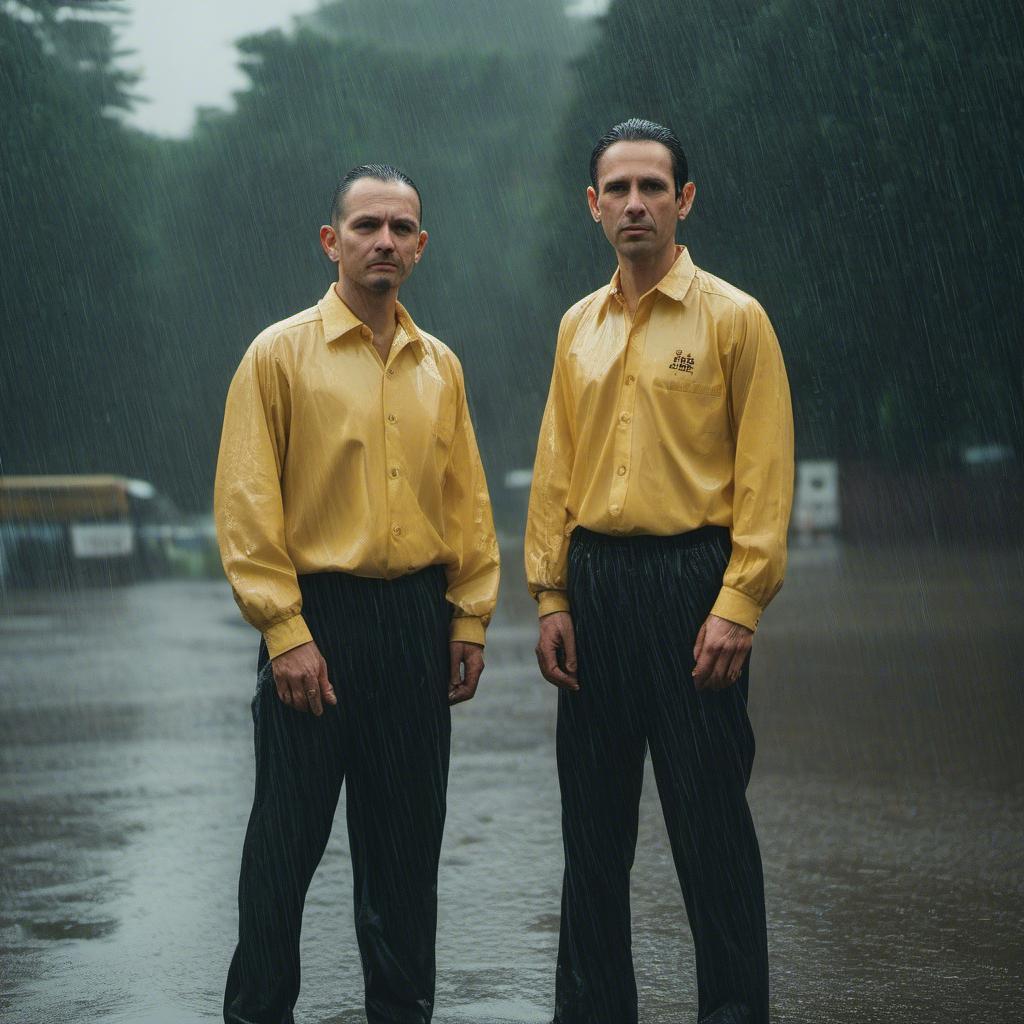}
    \parbox{1\linewidth}{\centering Two men in yellow shirts standing in the rain}
    \caption{Visual comparisons between our methods and baselines. 
    More examples with SDXL-Base in the Appendix.
    }
    \label{fig:visual}
\end{figure*}

As shown in Table~\ref{tab:baselines}, 
in terms of face quality, FS-DPO surpasses the baselines by considerable margins. 
The general image generation ability of FaceDPO can also be considerably improved compared to the base model.
Nevertheless, due to the limited variety of training prompts and the small size of the preference dataset, the improvements in general ability are less than those caused by DPO.
We also note that it is useful to use negative prompts and the FaceLoRA for better face quality (evidenced by higher \textit{FS}), but the general ability is decremented. 
We present some examples based on human-centric prompts from HumanArt in Figure~\ref{fig:visual}.
As shown, compared to the base model and that with negative prompts, our model generates more attractive images, containing fewer collapsed faces. 
Though SDXL-DPO generates globally appealing images, our method generates more normal faces than them, especially in the eye region.
This shows a gap between global aesthetics and detail generation.

\begin{figure}[t]
    \centering
    \includegraphics[width=0.3\linewidth]{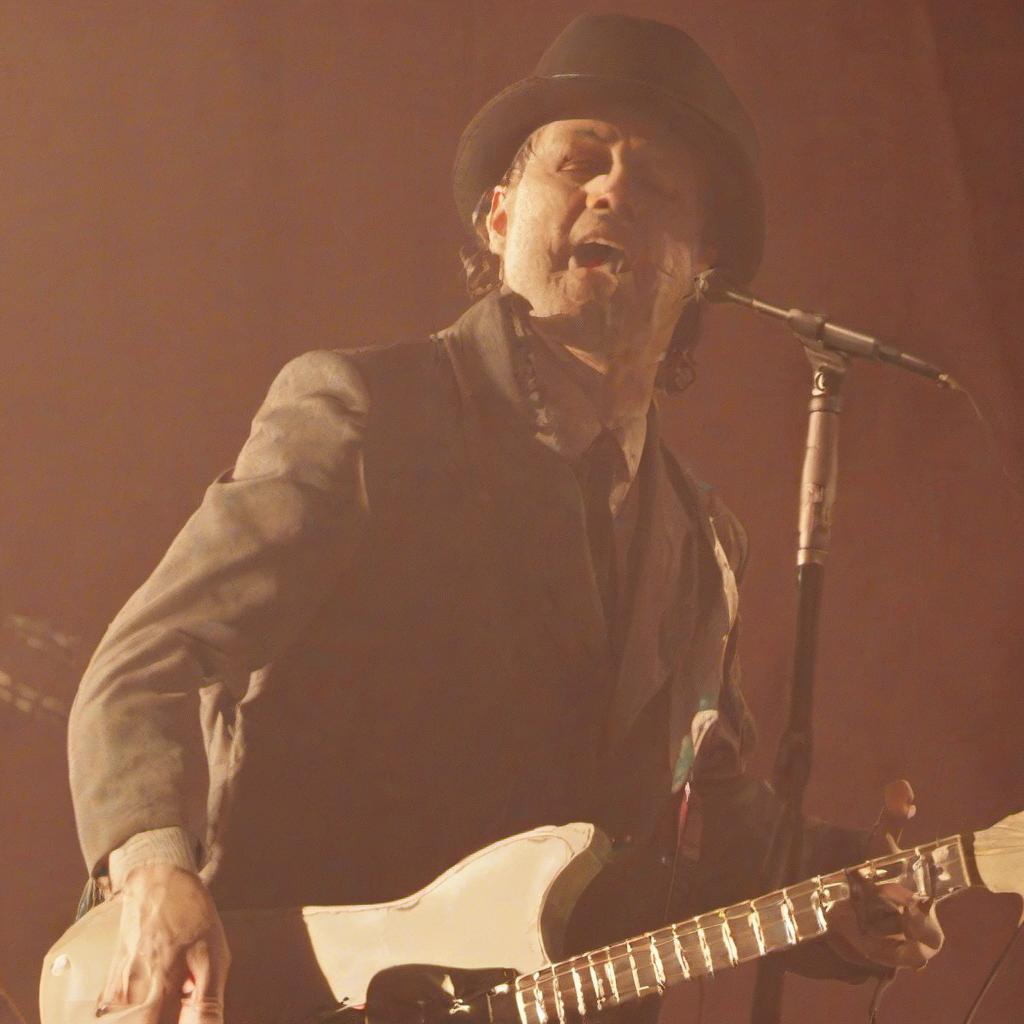}
    \includegraphics[width=0.3\linewidth]{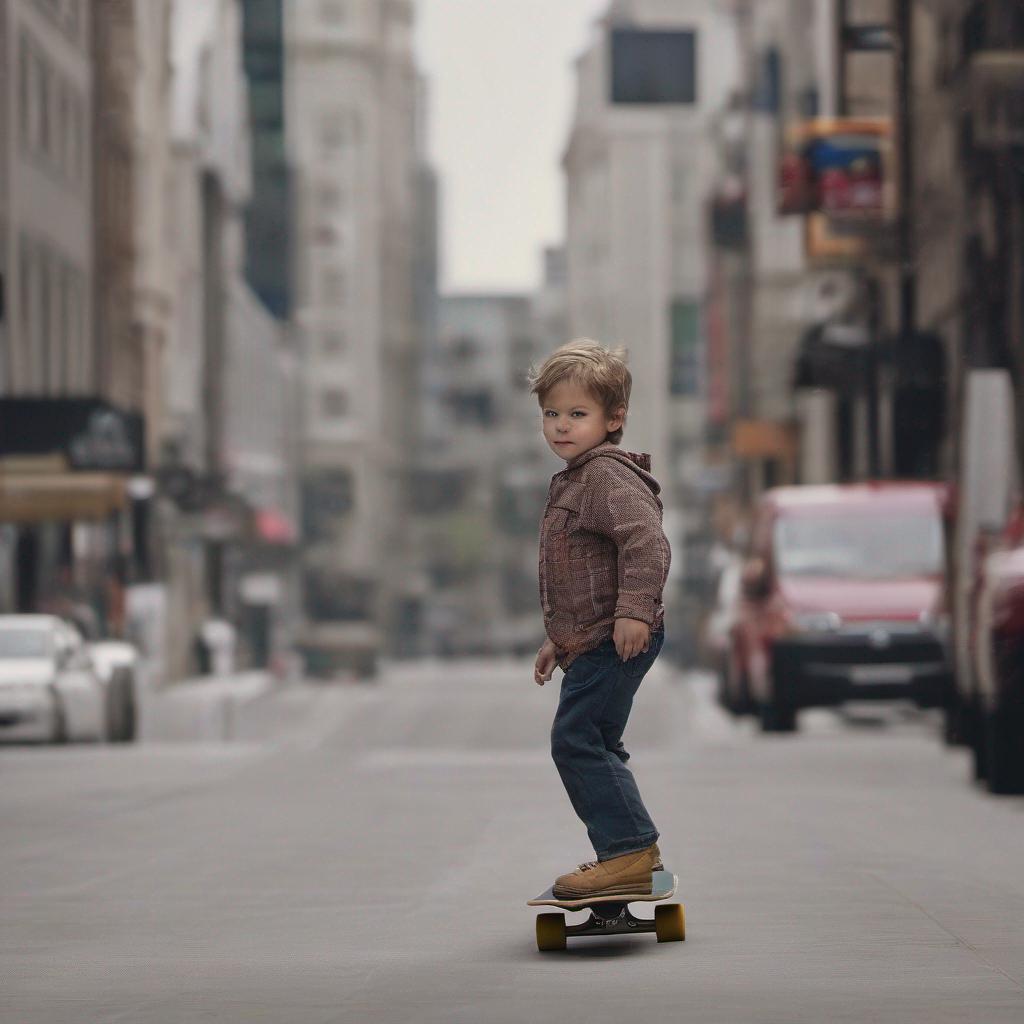}
    \includegraphics[width=0.3\linewidth]{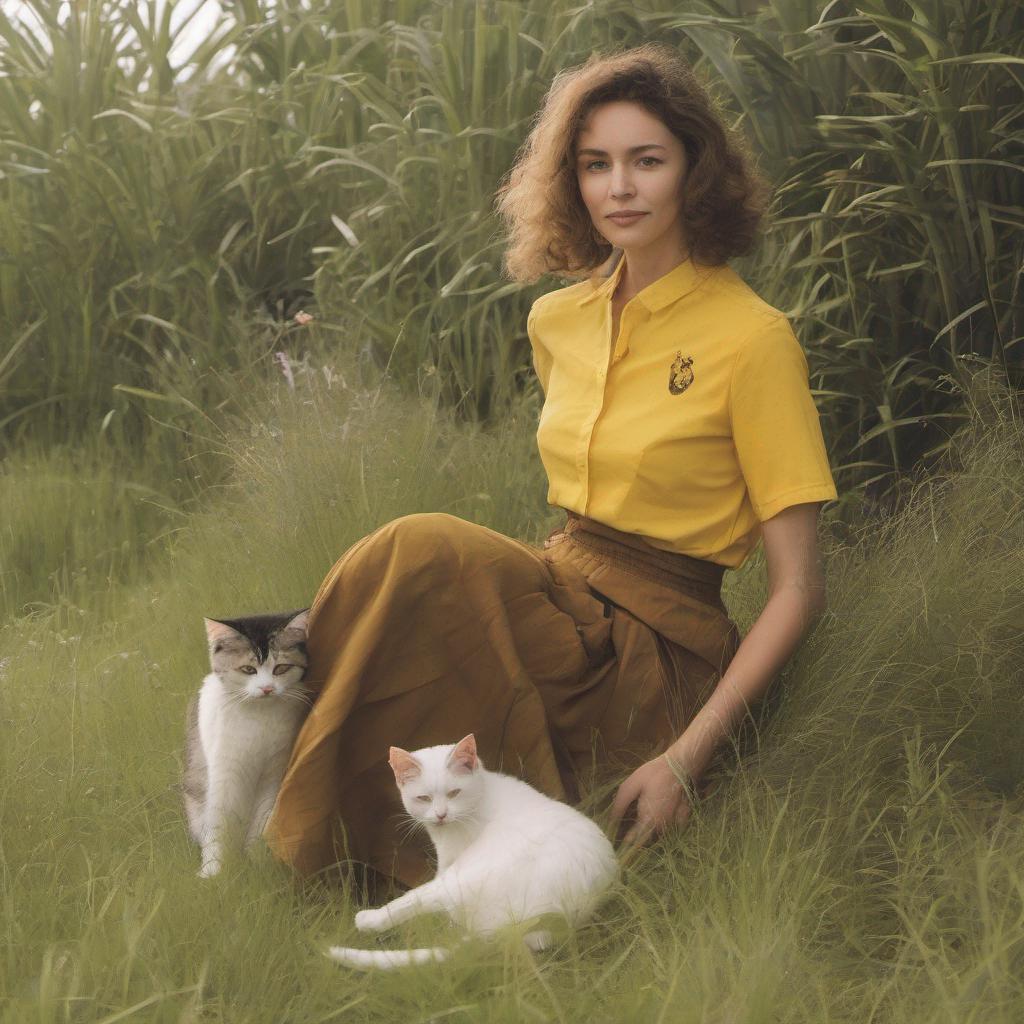}
     \parbox{0.3\linewidth}{\centering -0.49}
    \parbox{0.3\linewidth}{\centering 2.85}
    \parbox{0.3\linewidth}{\centering 5.96}
    \caption{The correlation between face quality and \textit{FS}.
    }
    \label{fig:facearea}
\end{figure}
\subsubsection{Human Evaluation.}
We also conduct a human preference study on face quality between our fine-tuned model and SDXL-Base/SDXL-DPO on human-centric images from HumanArt as mentioned above.
We ask five annotators, who can choose one of the images based on their preference, or choose ``tie'', meaning they are unable to decide due to the similar quality of the two images, and the choices are aggregated.
Figure~\ref{fig:human eval} shows that our fine-tuned model achieves a significant improvement in preference over SDXL-Base and SDXL-DPO, proving our model can generate better human faces and indicating that \textit{FS} on face quality is consistent with human preferences.

\subsubsection{Analysis.}
We observe that our fine-tuned model tends to generate larger, unoccluded frontal faces, meaning \textit{FS} inclines towards rating higher scores to such faces.
It is rational since larger faces contain more details than smaller faces.
We clarify it is the face quality rather than the face area that controls \textit{FS}.
We present examples in Figure~\ref{fig:facearea} to demonstrate the positive correlation between face quality and \textit{FS}.

\section{Conclusion}
In this work, we focus on the bad face issue raised by diffusion models.
We construct a dataset of (win, loss) face pairs implicitly without annotations to develop a new metric named \textit{FaceScore} specifically for the evaluation of rationality and aesthetics of faces in the synthetic images and use it to filter data to improve the face quality of SDXL.
\bibliography{aaai25}
\clearpage

\begin{figure}[t]
    \begin{center}
        \huge \textbf{Appendix}
    \end{center}
\end{figure}

\section{Sampling Guidance based on \textit{FS}.}
A scorer can offer guidance to the denoising process to improve the image quality.
As mentioned in the body, the FaceScore Model $s_\phi$ can assign scores to the faces in the image that measure the quality of faces.
This guidance can be incorporated into the sampling process in a classifier guidance manner.
Specifically, in each denoising step, we predict the clean latent variable $z_0$ with $z_t$ by the following equation:
\begin{equation}
    z_0=\frac{z_t-\sqrt{1-\alpha_t^2}\epsilon_\theta(z_t,t,c)}{\alpha_t},
\end{equation}
where $t$ is the current denoising timestep and $c$ is the context conditions.
With this, we can get the corresponding FaceScore for this image $s_\phi(\mathcal{D}(z_0))$, and can modify the noise prediction with the gradient of it:
\begin{equation}
    \epsilon_\theta^{new}(z_t,t,c) = \epsilon_\theta(z_t,t,c) - \mu \cdot \nabla s_\phi(\mathcal{D}(z_0)),
\end{equation}
where $\mu$ controls the intensity of the guidance.

\begin{figure}[t]
    \centering
    \includegraphics[width=1\linewidth]{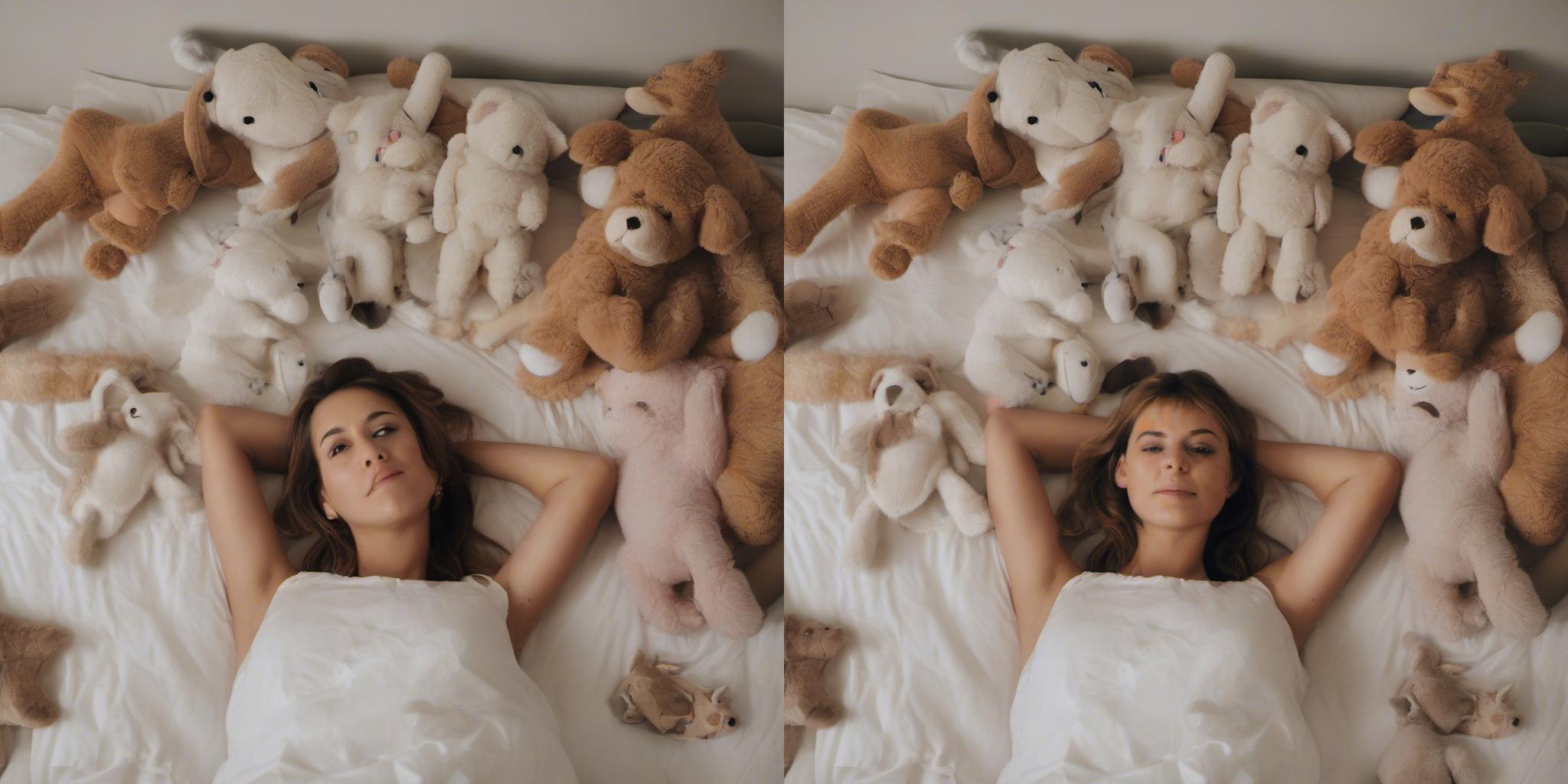}
    \parbox{1\linewidth}{\centering A woman laying on a bed with stuffed animals.}
    \includegraphics[width=1\linewidth]{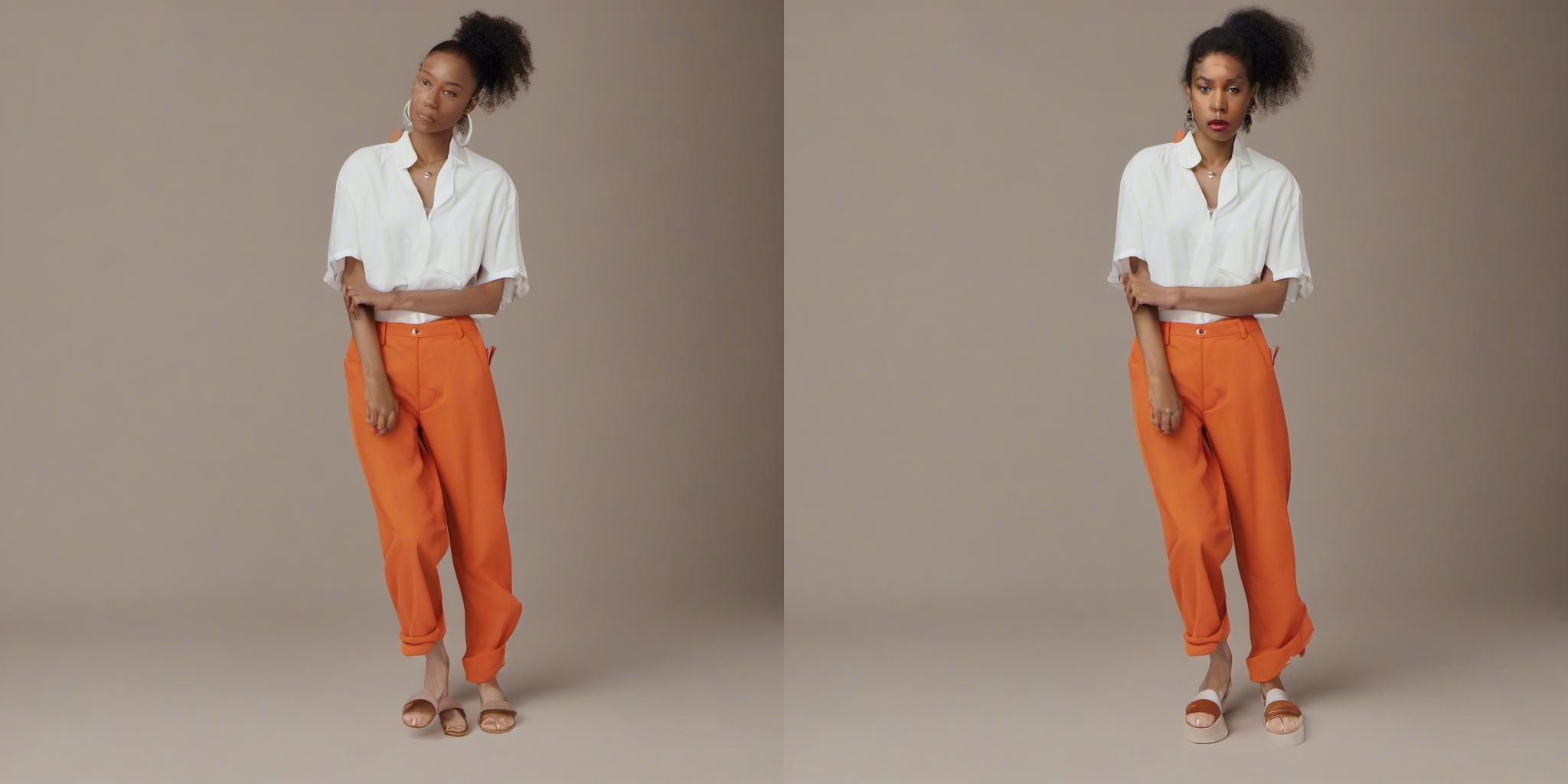}
    \parbox{1\linewidth}{\centering A woman in orange pants and a white shirt.}
    \caption{Some cases using \textit{FS} as guidance information during sampling process.
    The left one is generated by SDXL-Base, and the right one uses sampling guidance.}
    \label{fig:guidance case}
\end{figure}

We also observe that the guidance sometimes becomes quite large, which can lead to striping artifacts in the final image, so we perform clamping on the guidance.
We present some cases in Figure~\ref{fig:guidance case}.

\section{Ranking Criteria for Evaluation}
\begin{figure*}[t]
    \parbox{0.15\linewidth}{\centering Score 1}
    \parbox{0.15\linewidth}{\centering Score 2}
    \parbox{0.15\linewidth}{\centering Score 3}
    \hfill
    \parbox{0.15\linewidth}{\centering Score 1}
    \parbox{0.15\linewidth}{\centering Score 2}
    \parbox{0.15\linewidth}{\centering Score 3}
    \includegraphics[width=0.15\linewidth]{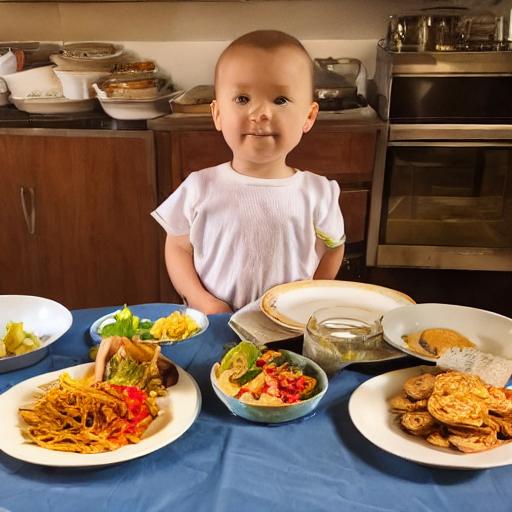}
    \includegraphics[width=0.15\linewidth]{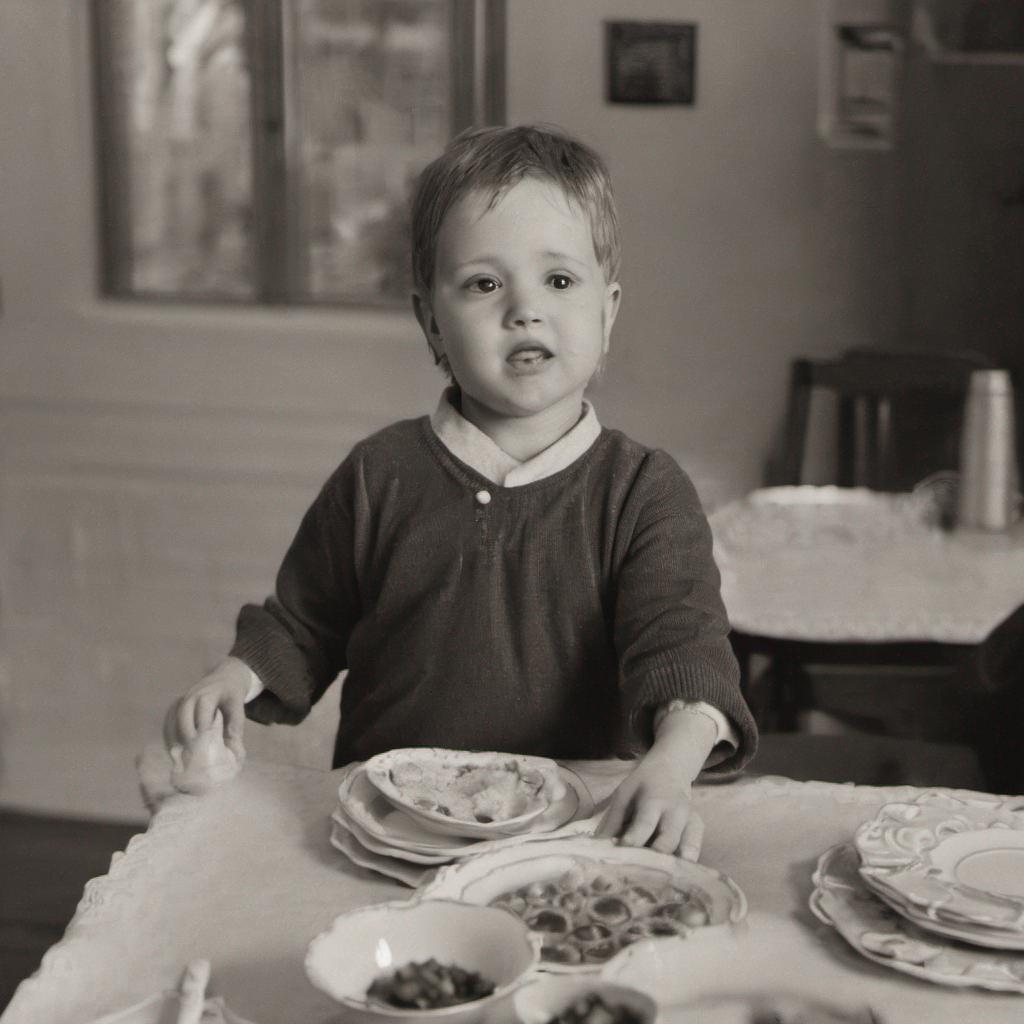}
    \includegraphics[width=0.15\linewidth]{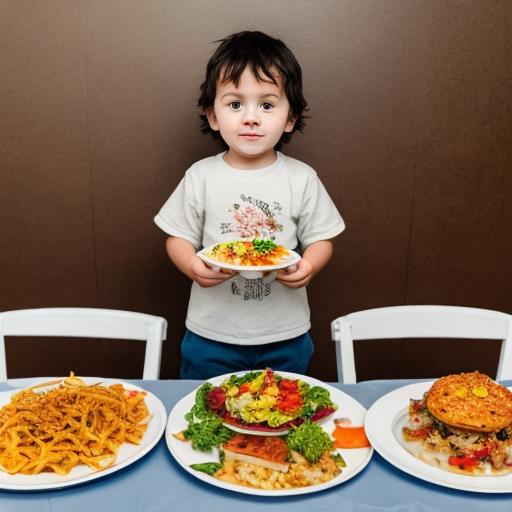}
    \hfill
    \includegraphics[width=0.15\linewidth]{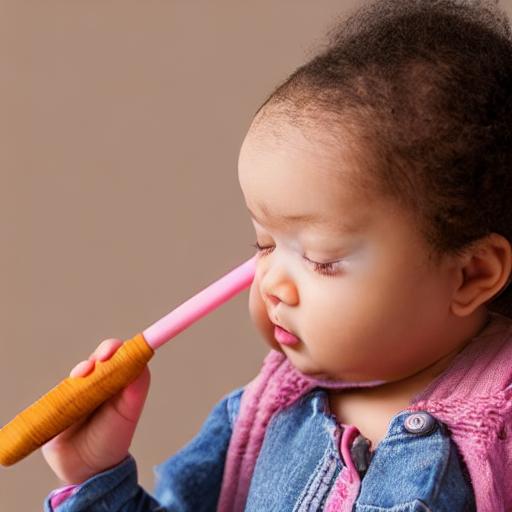}
    \includegraphics[width=0.15\linewidth]{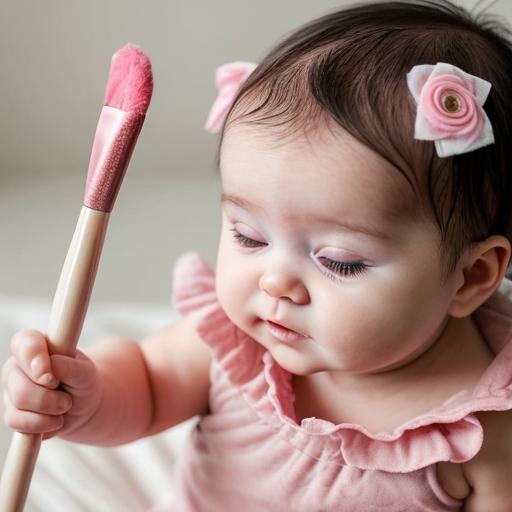}
    \includegraphics[width=0.15\linewidth]{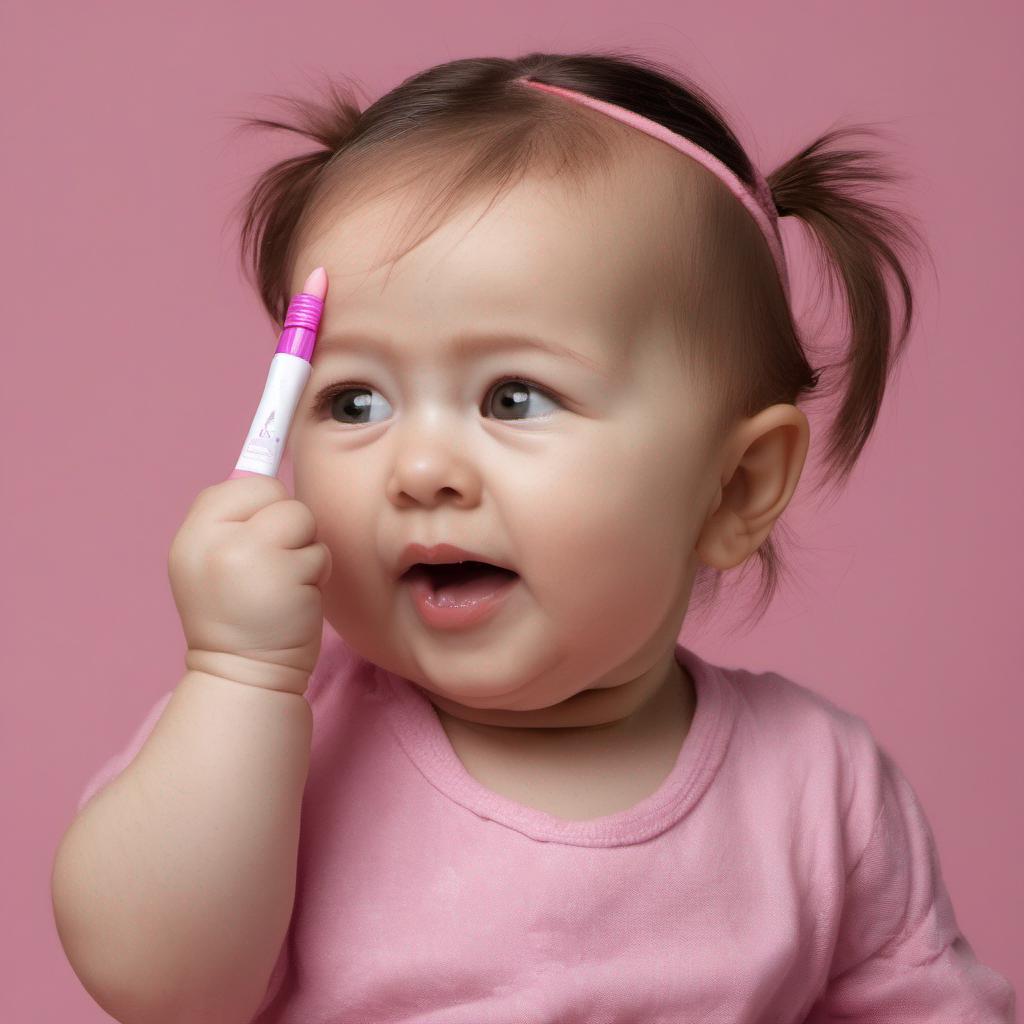}
    \parbox{0.45\linewidth}{\centering A young child standing \\in front of a table with plates.}
    \hfill
    \parbox{0.45\linewidth}{\centering A baby girl is holding a pink brush \\as she scratches her head.}

    \includegraphics[width=0.15\linewidth]{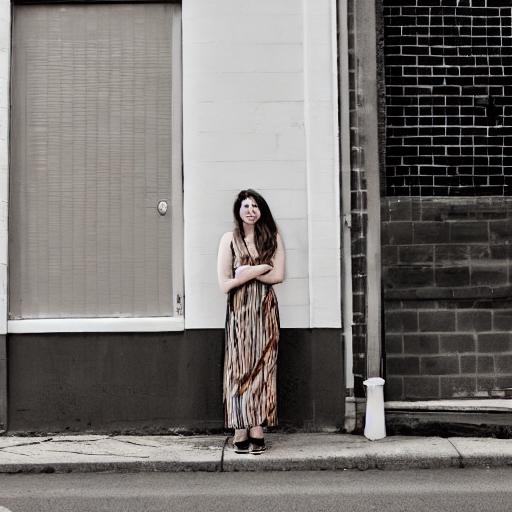}
    \includegraphics[width=0.15\linewidth]{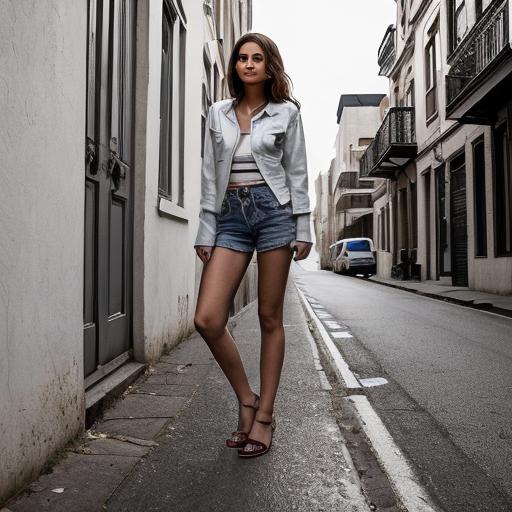}
    \includegraphics[width=0.15\linewidth]{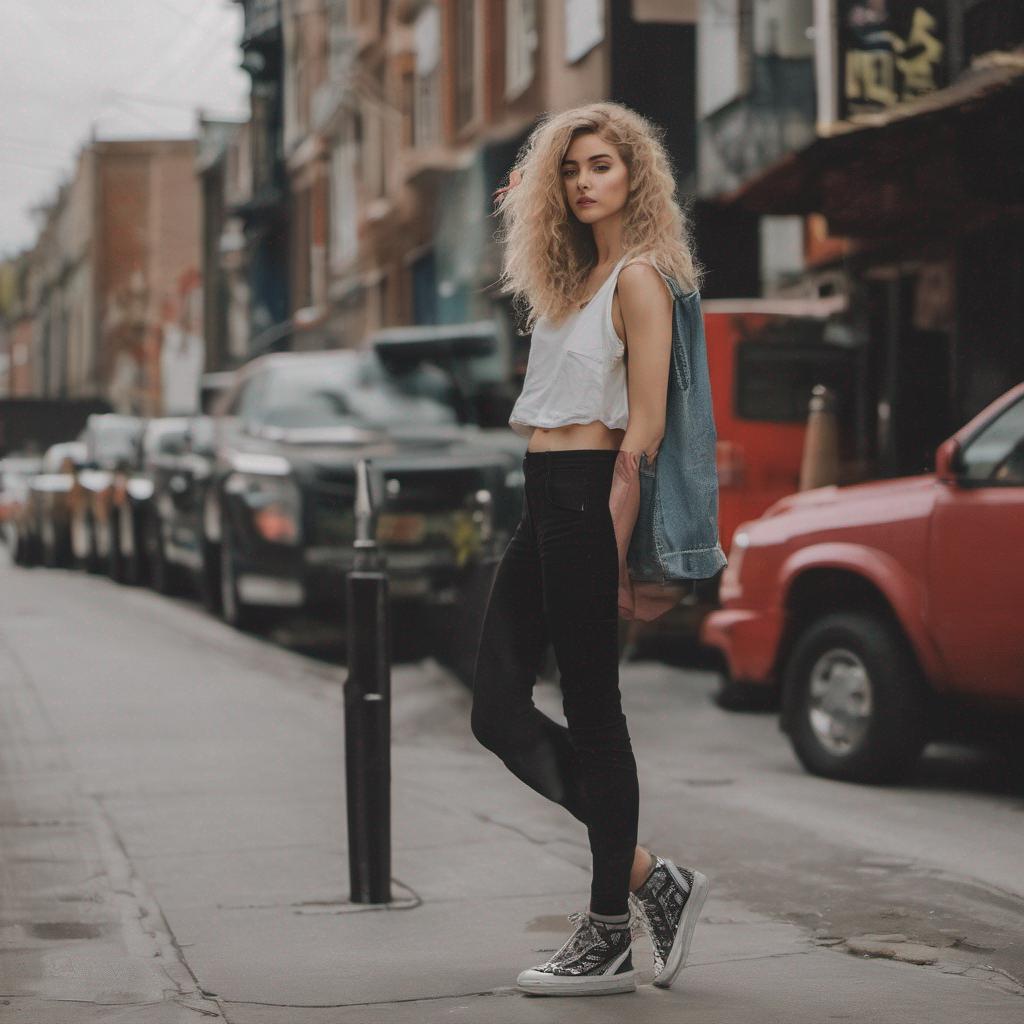}
    \hfill
    \includegraphics[width=0.15\linewidth]{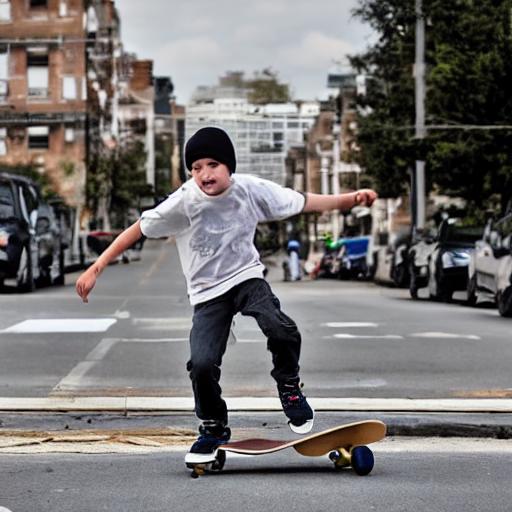}
    \includegraphics[width=0.15\linewidth]{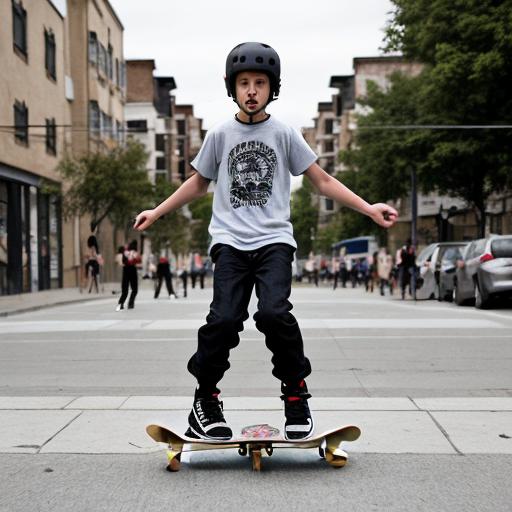}
    \includegraphics[width=0.15\linewidth]{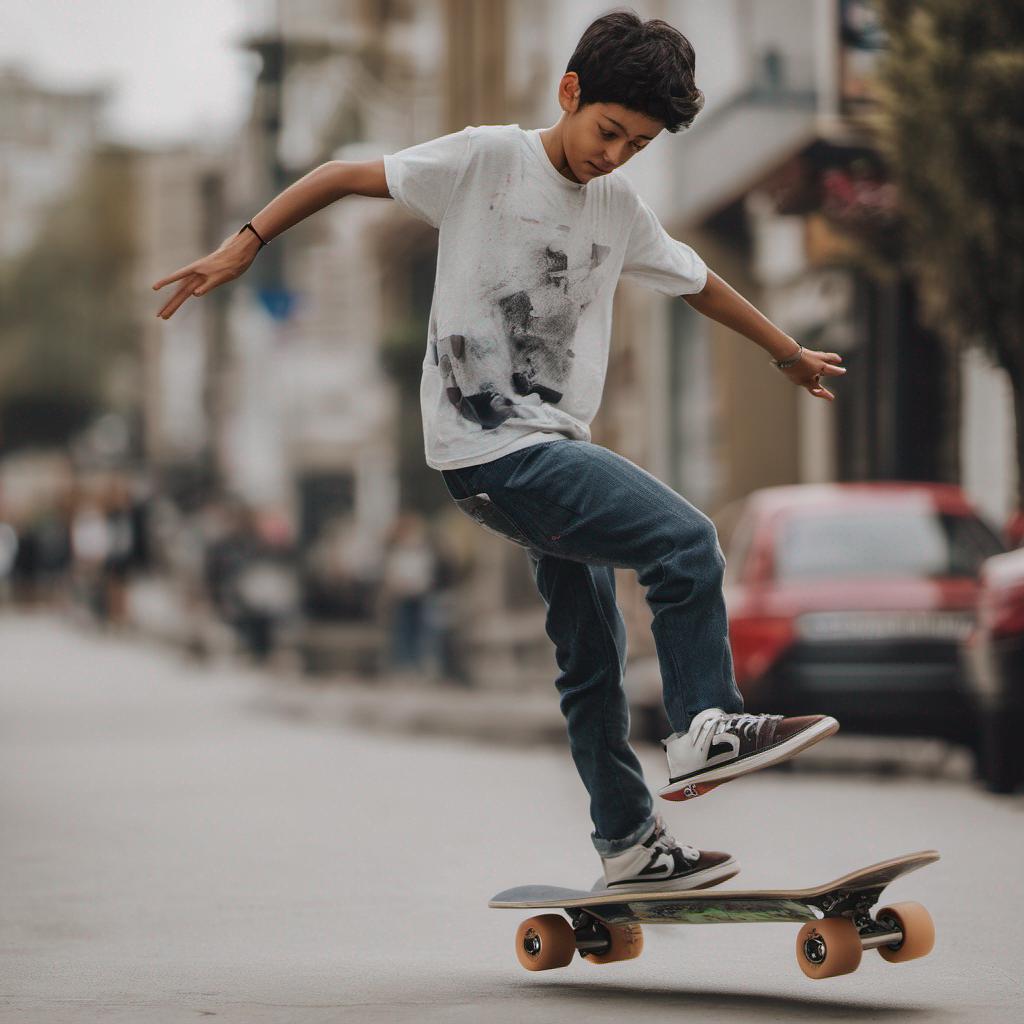}
    \parbox{0.45\linewidth}{\centering A beautiful woman standing on the side of \\a rad next to a street.}
    \hfill
    \parbox{0.45\linewidth}{\centering A boy doing a skateboard trick on a street.}

    \includegraphics[width=0.15\linewidth]{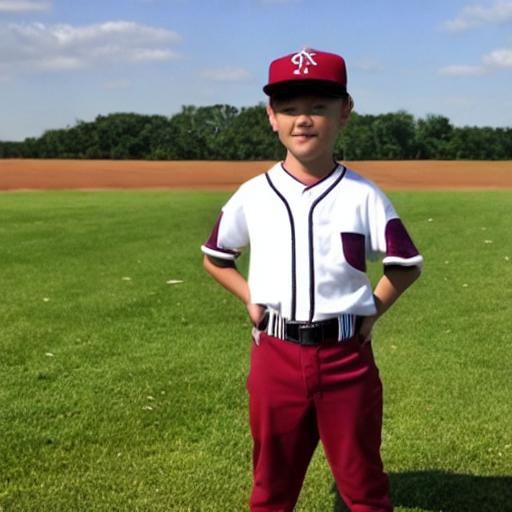}
    \includegraphics[width=0.15\linewidth]{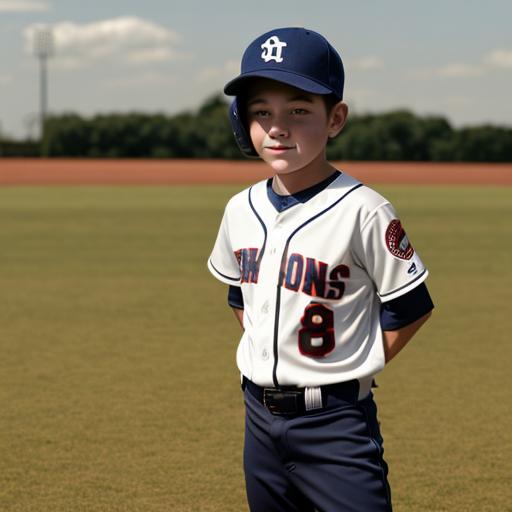}
    \includegraphics[width=0.15\linewidth]{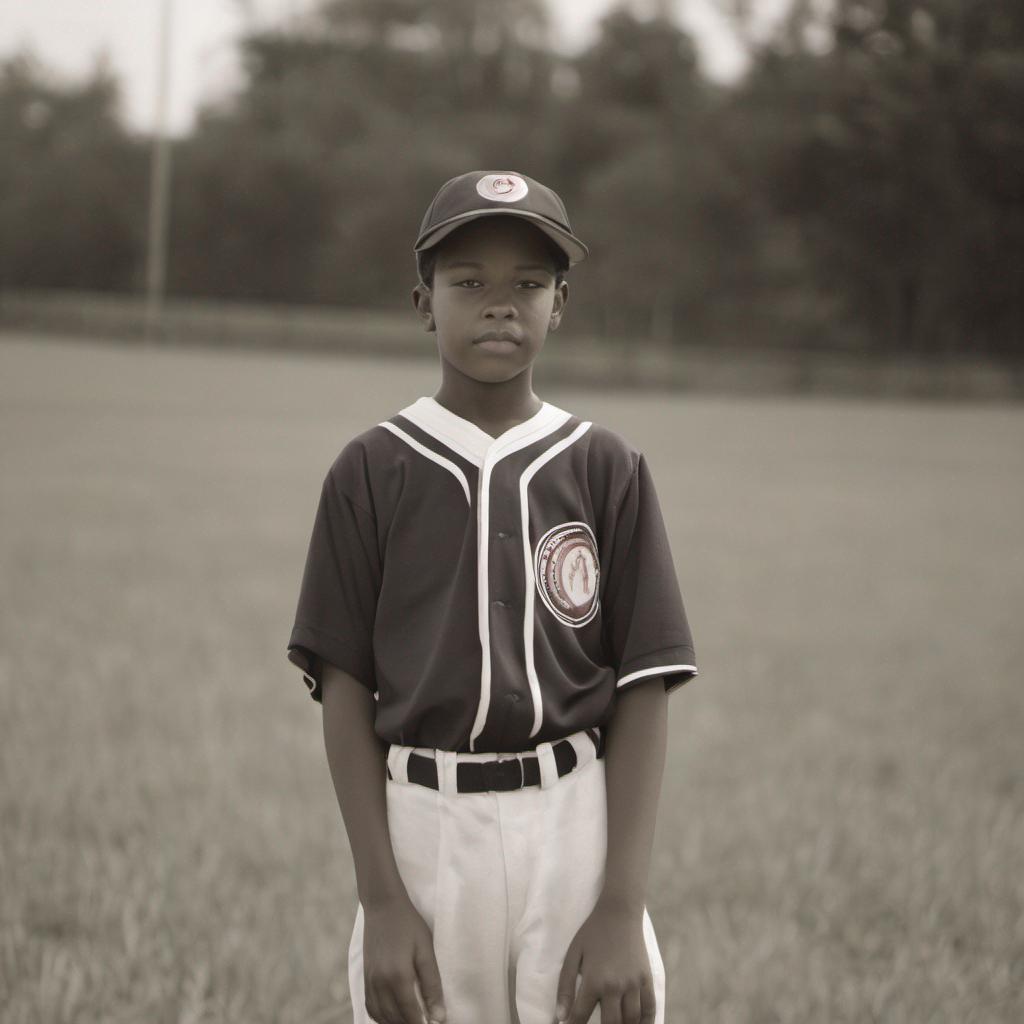}
    \hfill
    \includegraphics[width=0.15\linewidth]{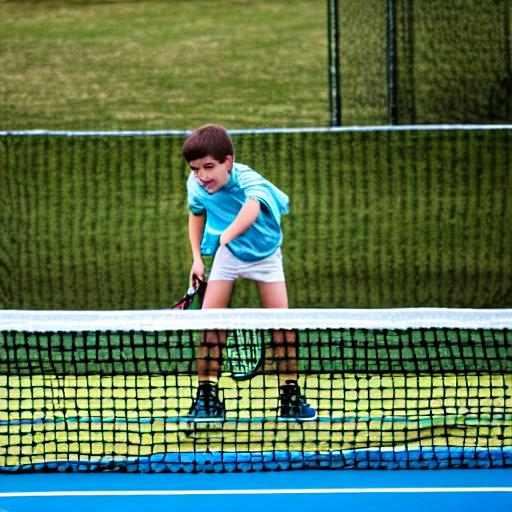}
    \includegraphics[width=0.15\linewidth]{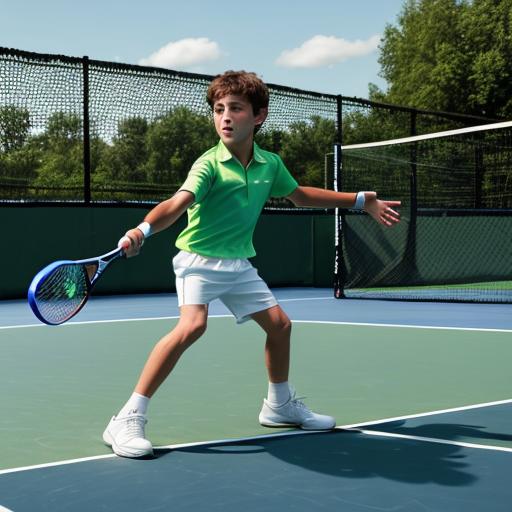}
    \includegraphics[width=0.15\linewidth]{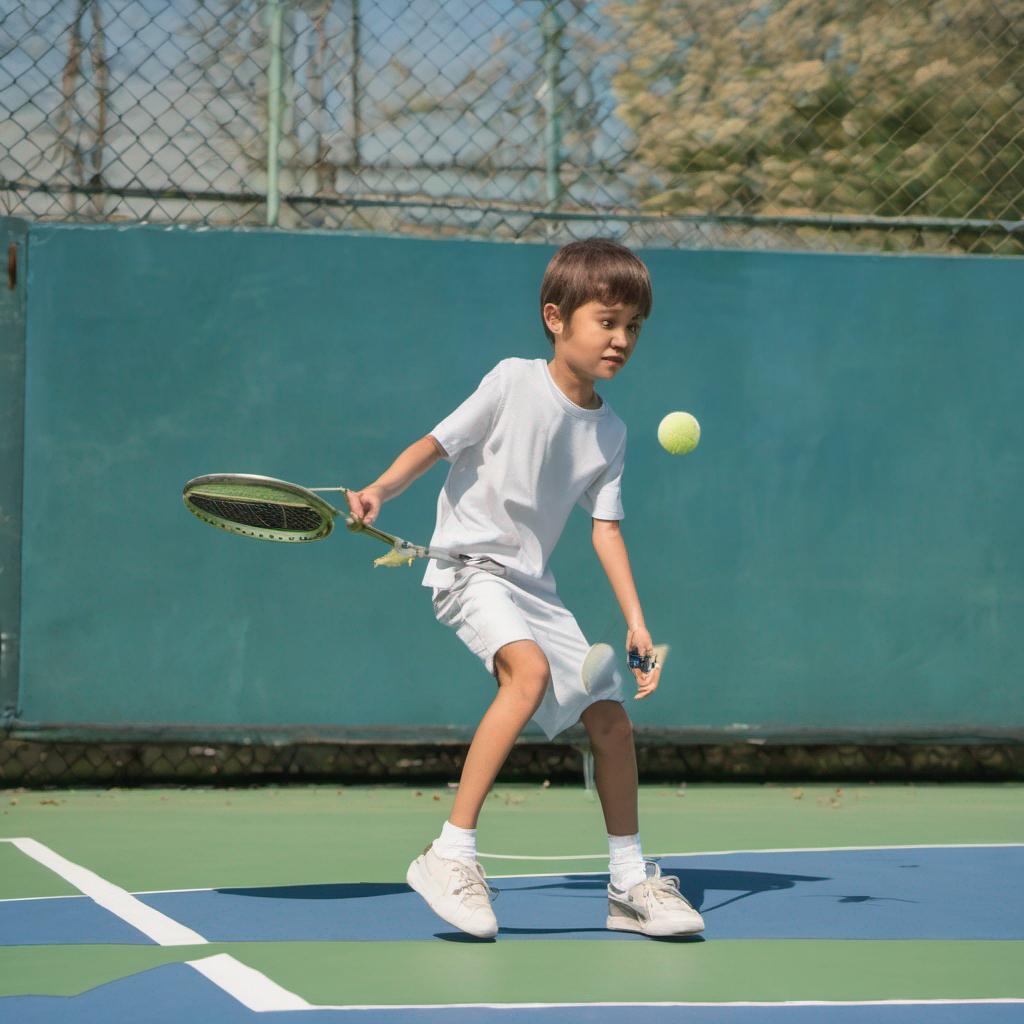}
    \parbox{0.45\linewidth}{\centering A boy dressed in a baseball uniform \\standing in a field.}
    \hfill
    \parbox{0.45\linewidth}{\centering A boy playing tennis on a \\blue and green tennis court.}

    \includegraphics[width=0.15\linewidth]{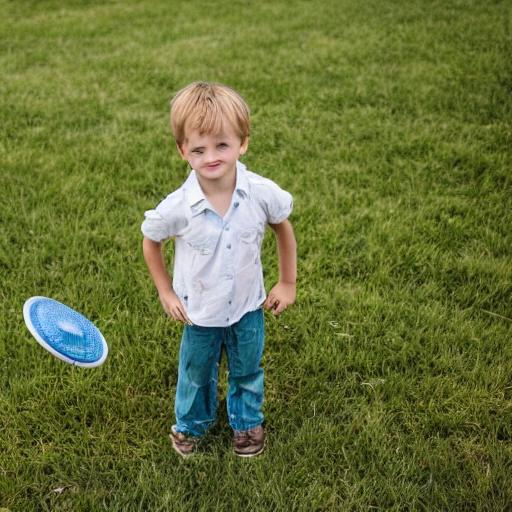}
    \includegraphics[width=0.15\linewidth]{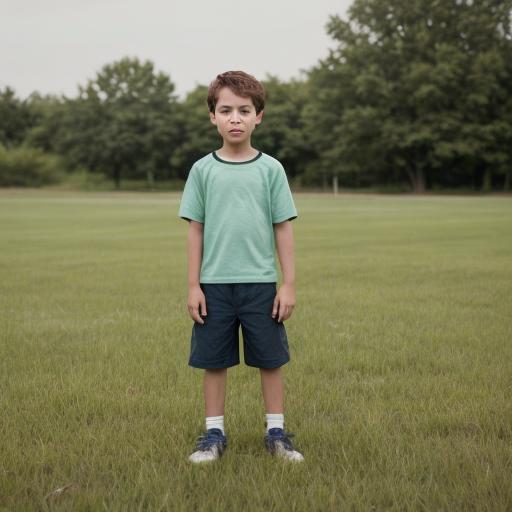}
    \includegraphics[width=0.15\linewidth]{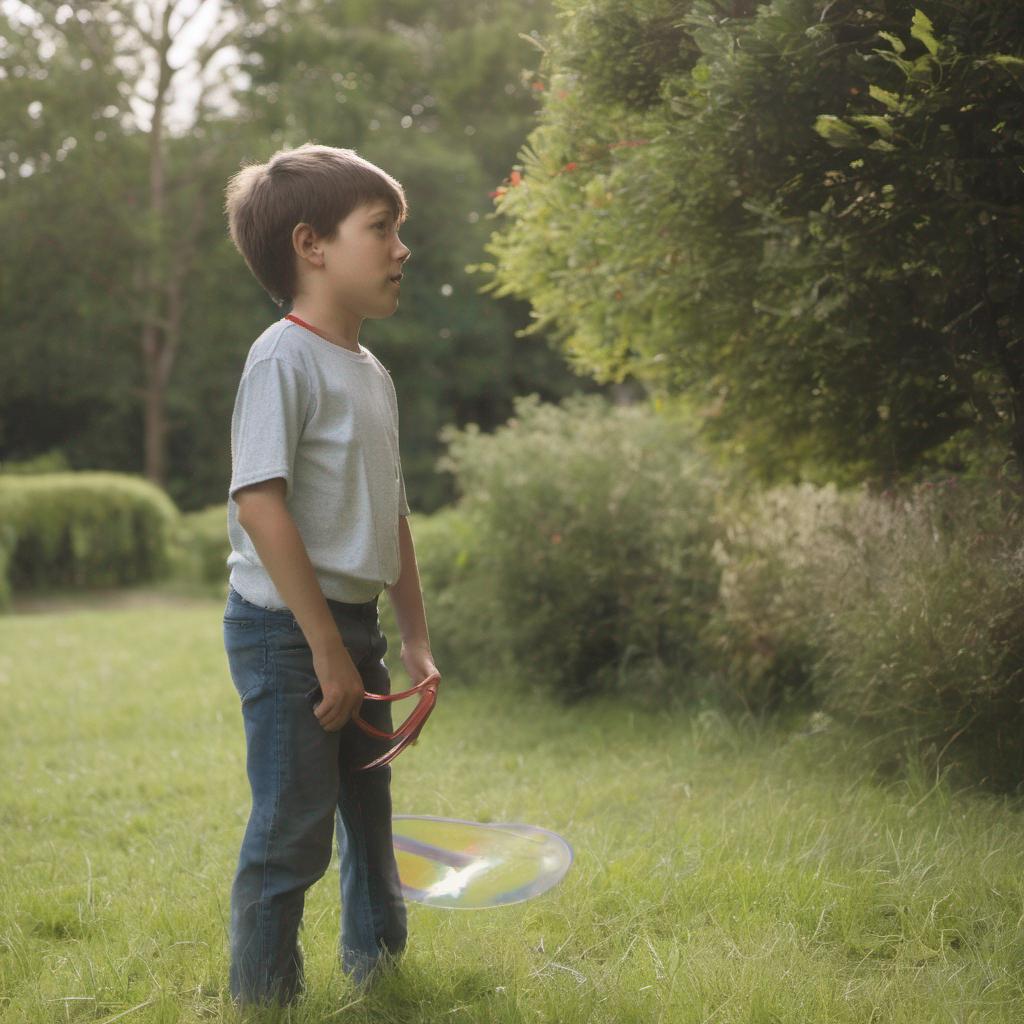}
    \hfill
    \includegraphics[width=0.15\linewidth]{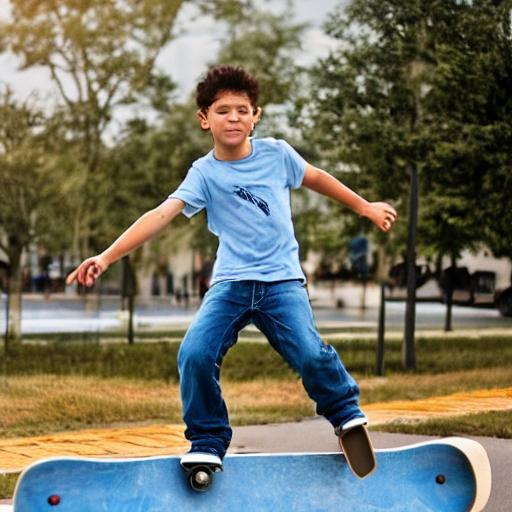}
    \includegraphics[width=0.15\linewidth]{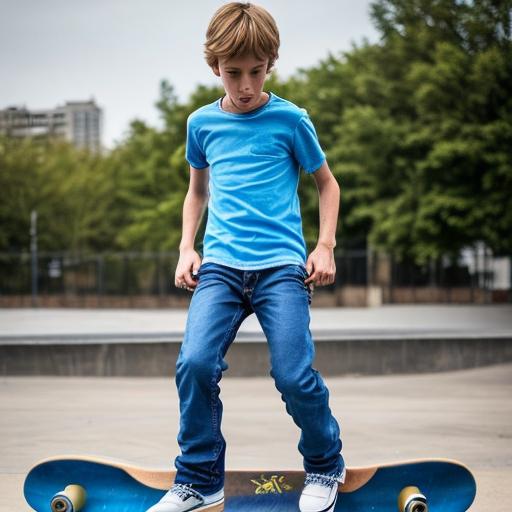}
    \includegraphics[width=0.15\linewidth]{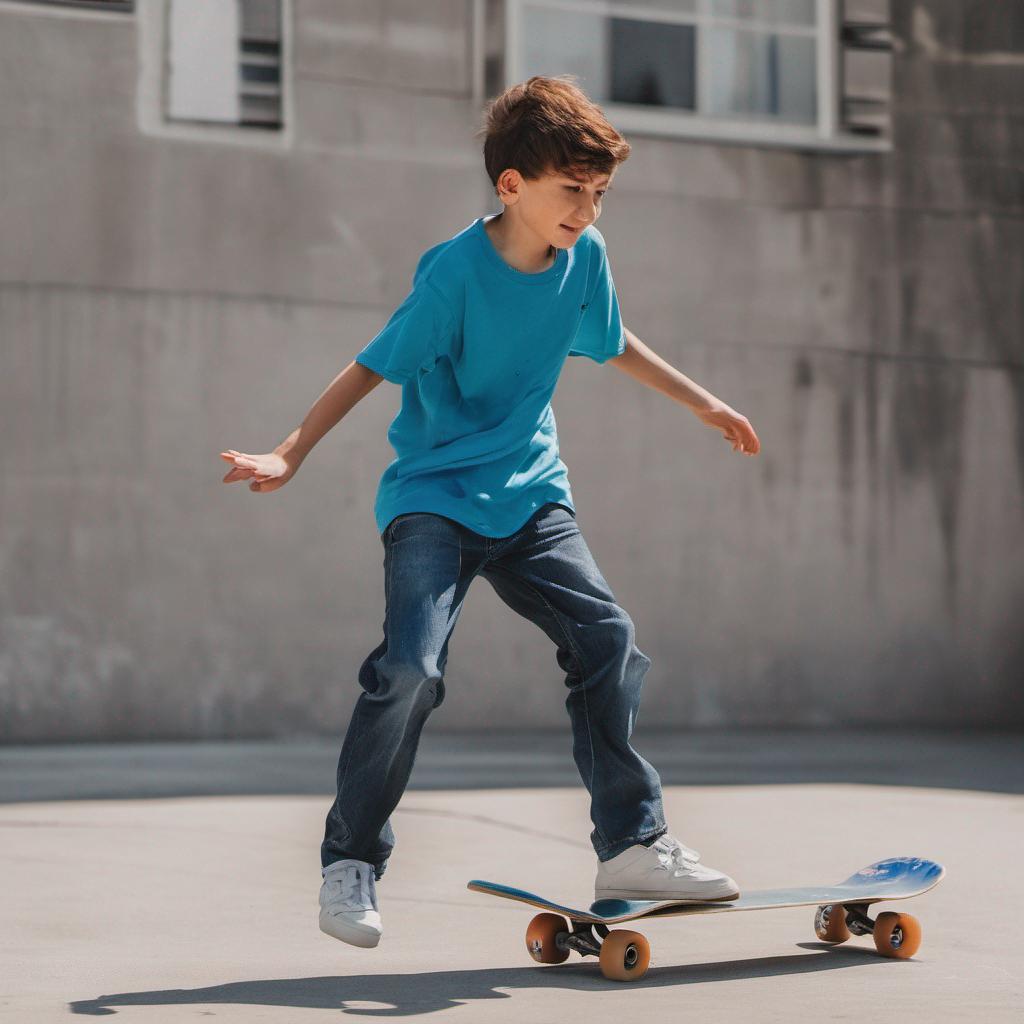}
    \parbox{0.45\linewidth}{\centering A boy standing in the grass with a frisbee.}
    \hfill
    \parbox{0.45\linewidth}{\centering A boy with a blue shirt and jean pants \\doing a trick with his skateboard.}

    \includegraphics[width=0.15\linewidth]{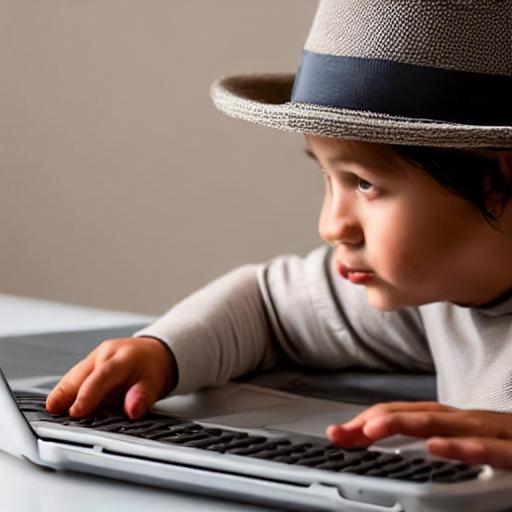}
    \includegraphics[width=0.15\linewidth]{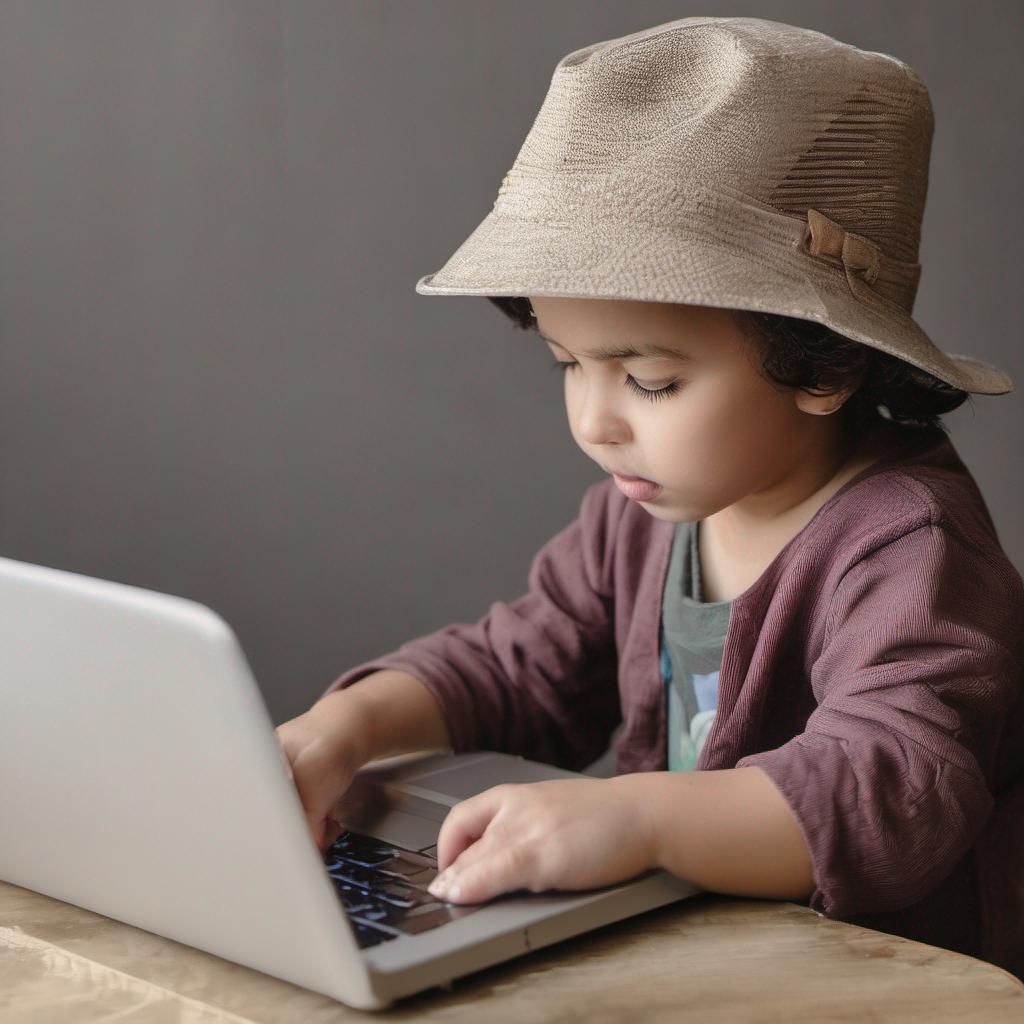}
    \includegraphics[width=0.15\linewidth]{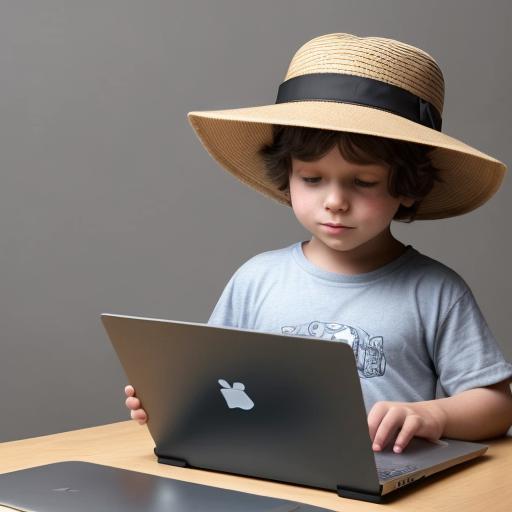}
    \hfill
    \includegraphics[width=0.15\linewidth]{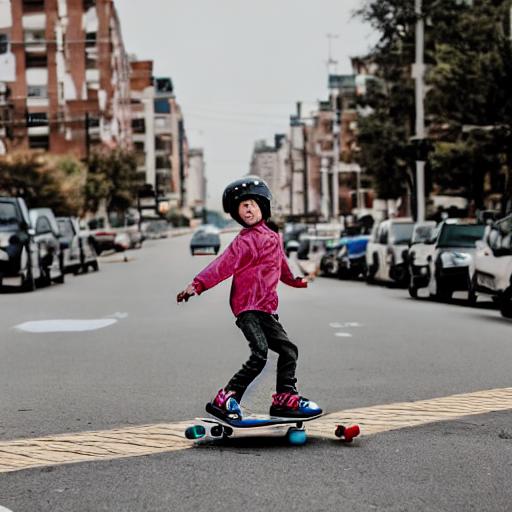}
    \includegraphics[width=0.15\linewidth]{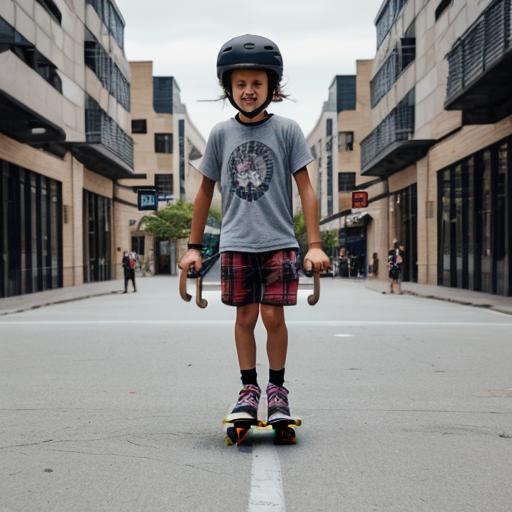}
    \includegraphics[width=0.15\linewidth]{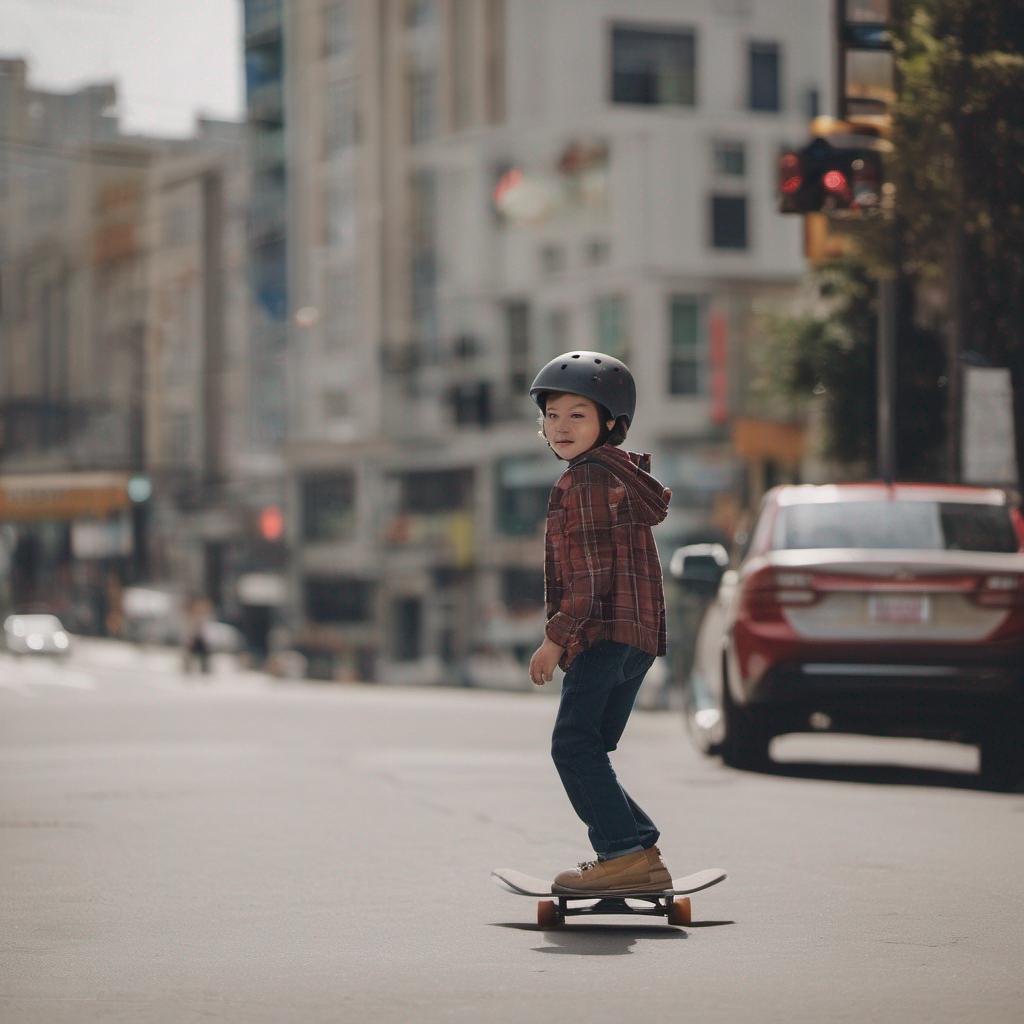}
    \parbox{0.45\linewidth}{\centering A child in a hat playing on a laptop computer.}
    \hfill
    \parbox{0.45\linewidth}{\centering A child riding a skateboard on a city street.}

    \includegraphics[width=0.15\linewidth]{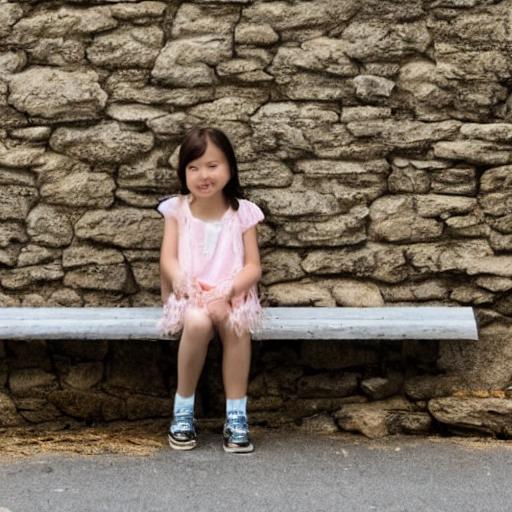}
    \includegraphics[width=0.15\linewidth]{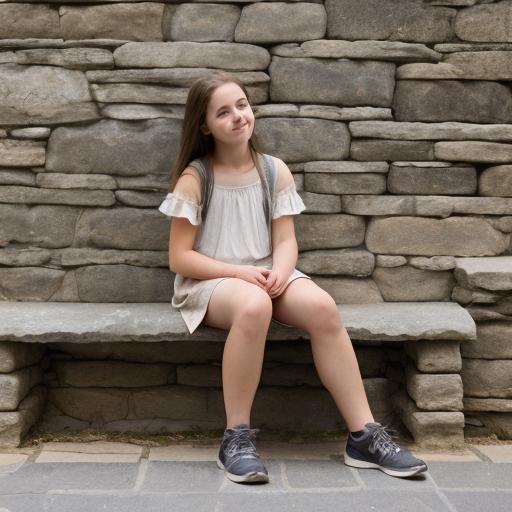}
    \includegraphics[width=0.15\linewidth]{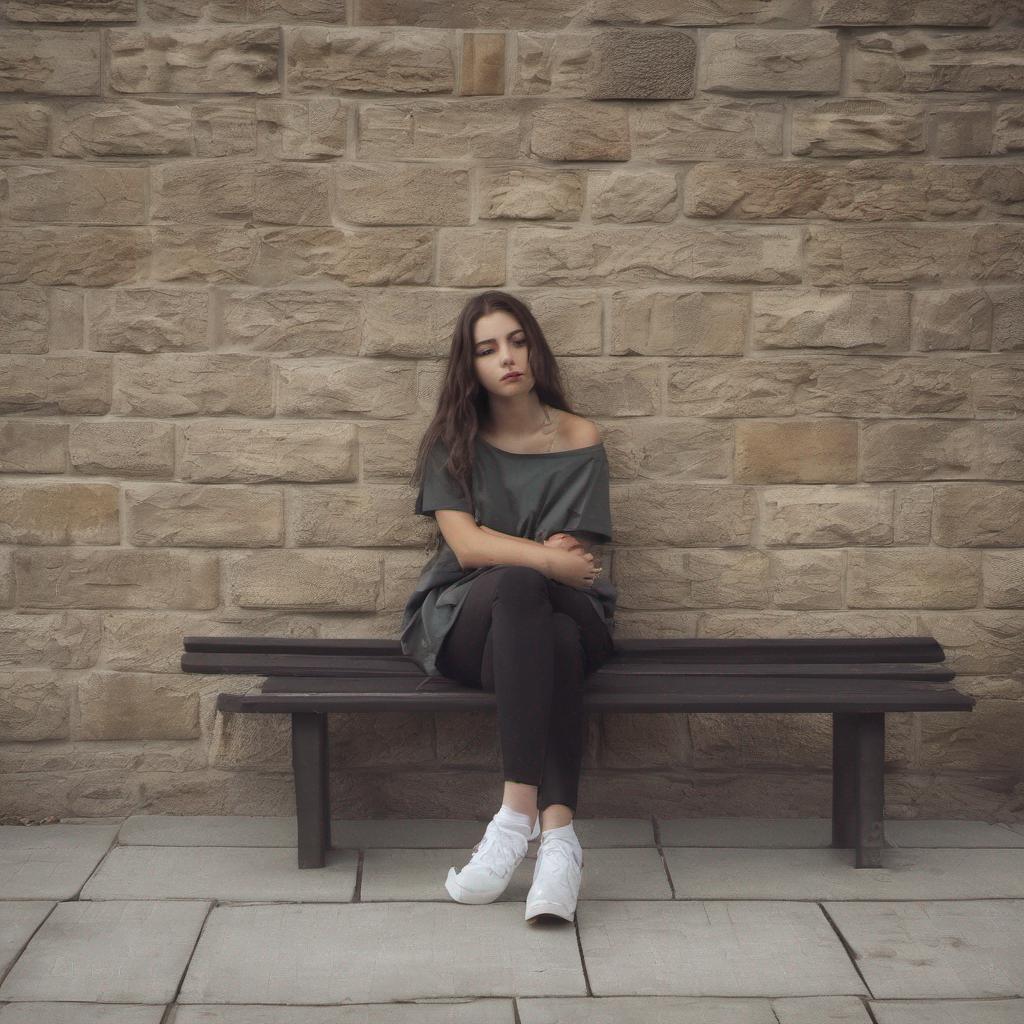}
    \hfill
    \includegraphics[width=0.15\linewidth]{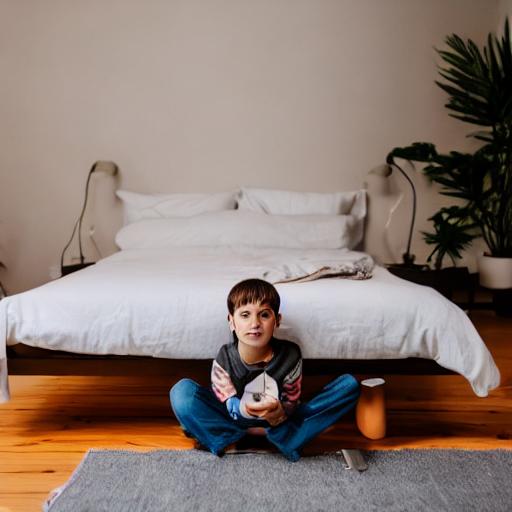}
    \includegraphics[width=0.15\linewidth]{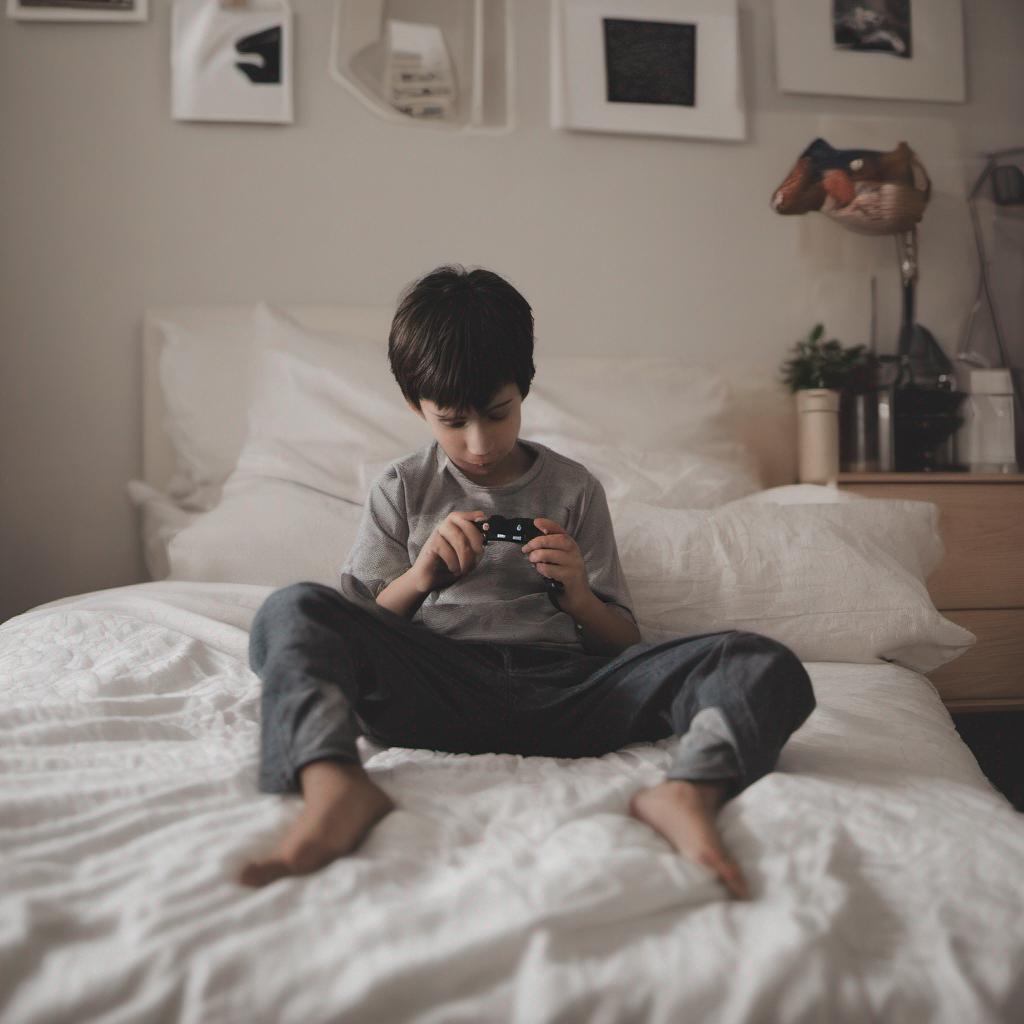}
    \includegraphics[width=0.15\linewidth]{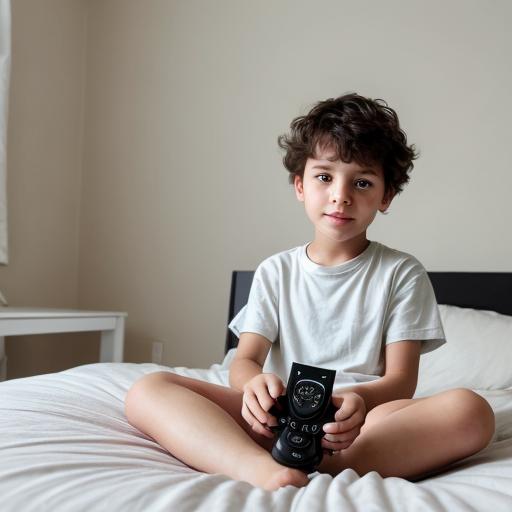}
    \parbox{0.45\linewidth}{\centering A girl sitting on a bench \\in front of a stone wall.}
    \hfill
    \parbox{0.45\linewidth}{\centering A kid sitting on a bed with a remote.}

    \caption{More examples of the human-annotated triplet. The image with higher face quality is assigned a higher score. 
    }
    \label{fig:evaluation dataset}
\end{figure*}

We establish the following annotation rules for human annotators:
\begin{itemize}
    \item[$\bullet$] Discard triplets if there are no valid faces in any image;
    \item[$\bullet$] Focus solely on the faces and do not need to consider the alignment between the prompt and the image, the aesthetic aspect of the image itself, or any irrelevant factors;
    \item[$\bullet$] Prioritize the rationality of the face before considering its aesthetic aspect.
    \item[$\bullet$] Select the most frontal and representative face for comparison purposes in multi-person scenes.
\end{itemize}
We also provide more image triplets in Figure~\ref{fig:evaluation dataset} to see the correlation between face quality and human preference.

\begin{figure*}
    \includegraphics[width=0.45\linewidth]{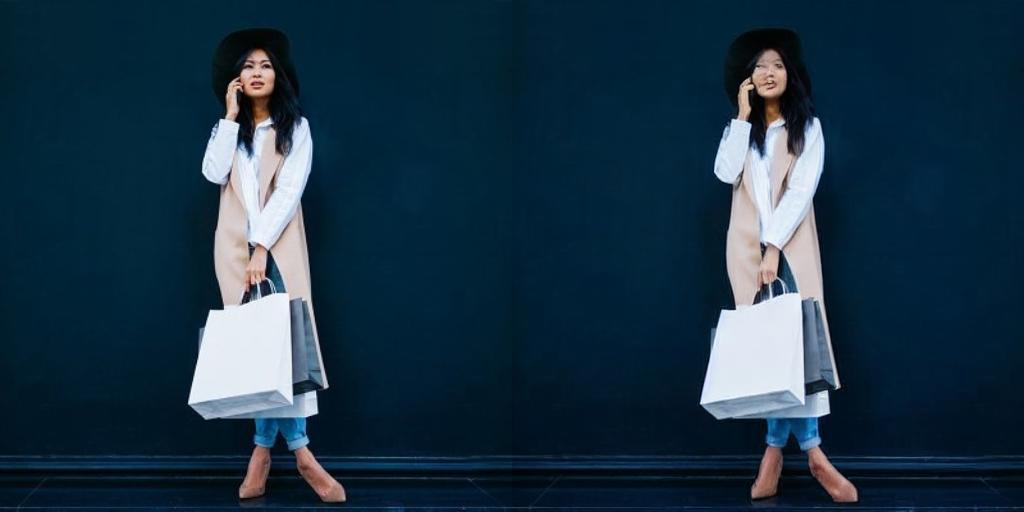}
    \hfill
    \includegraphics[width=0.45\linewidth]{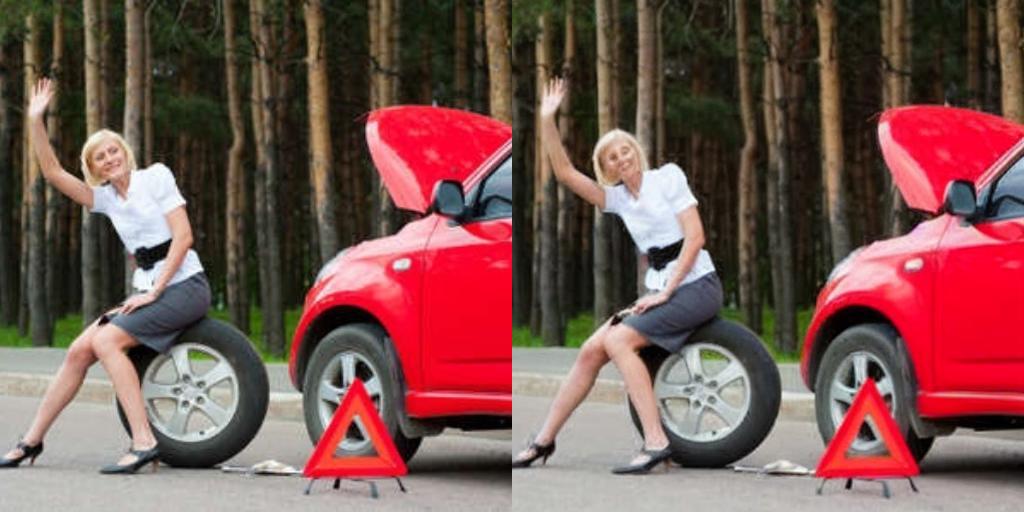}\\
    \parbox{0.45\linewidth}{\centering Woman holding shopping bags and talking on phone.}
    \hfill
    \parbox{0.45\linewidth}{\centering Woman with broken car and tire on road.}

    \includegraphics[width=0.45\linewidth]{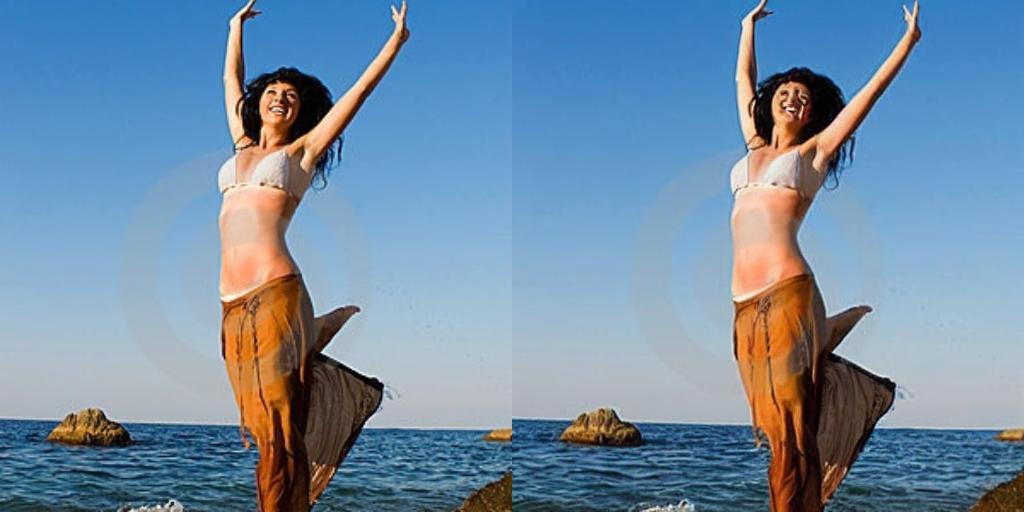}
    \hfill
    \includegraphics[width=0.45\linewidth]{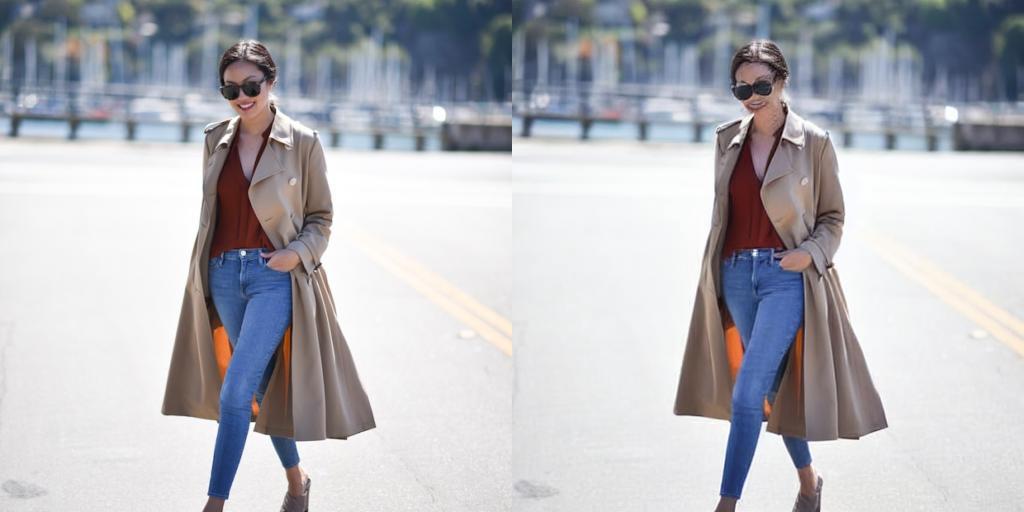}\\
    \parbox{0.45\linewidth}{\centering Woman dancing on the beach.}
    \hfill
    \parbox{0.45\linewidth}{\centering Woman in jeans and trench coat walking on the treet.}

    \includegraphics[width=0.45\linewidth]{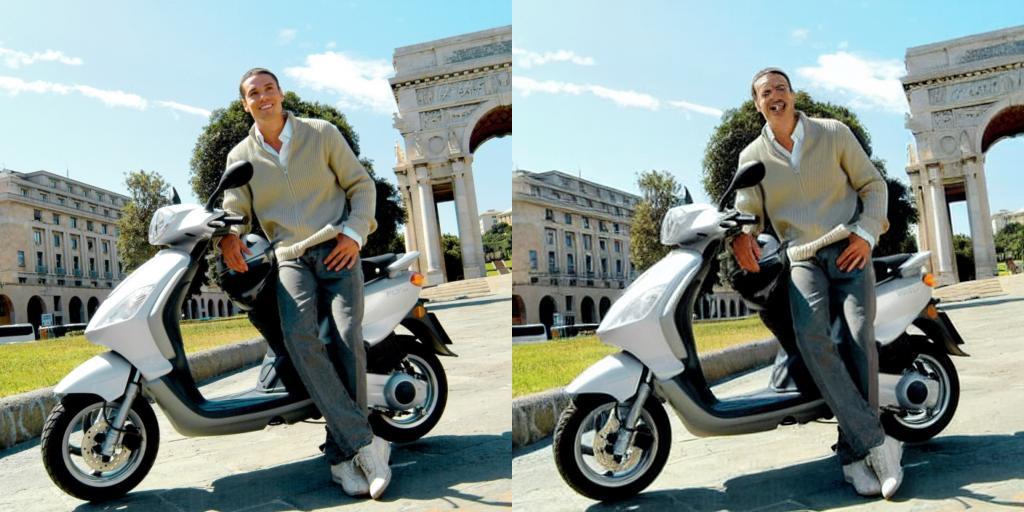}
    \hfill
    \includegraphics[width=0.45\linewidth]{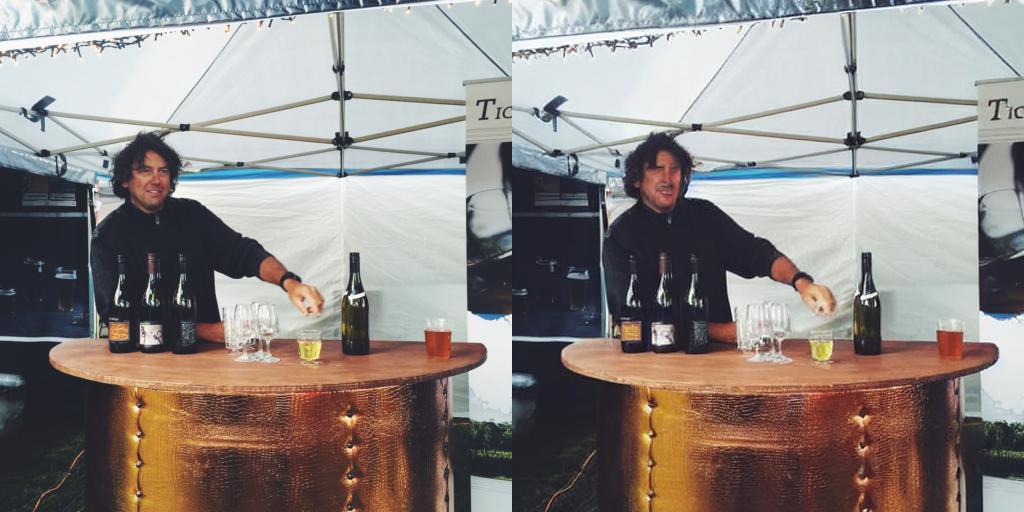}\\
    \parbox{0.45\linewidth}{\centering Man on a scooter.}
    \hfill
    \parbox{0.45\linewidth}{\centering Man pouring wine.}

    \includegraphics[width=0.45\linewidth]{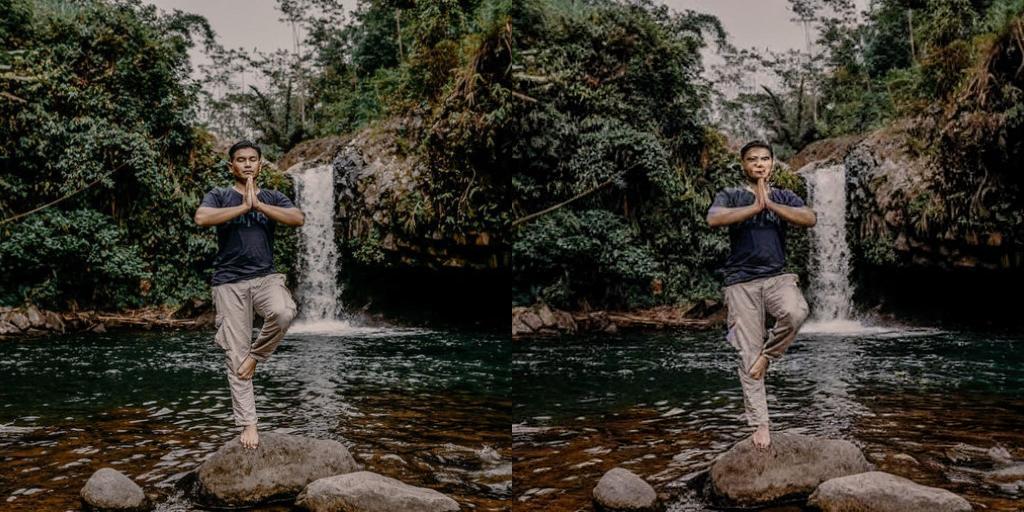}
    \hfill
    \includegraphics[width=0.45\linewidth]{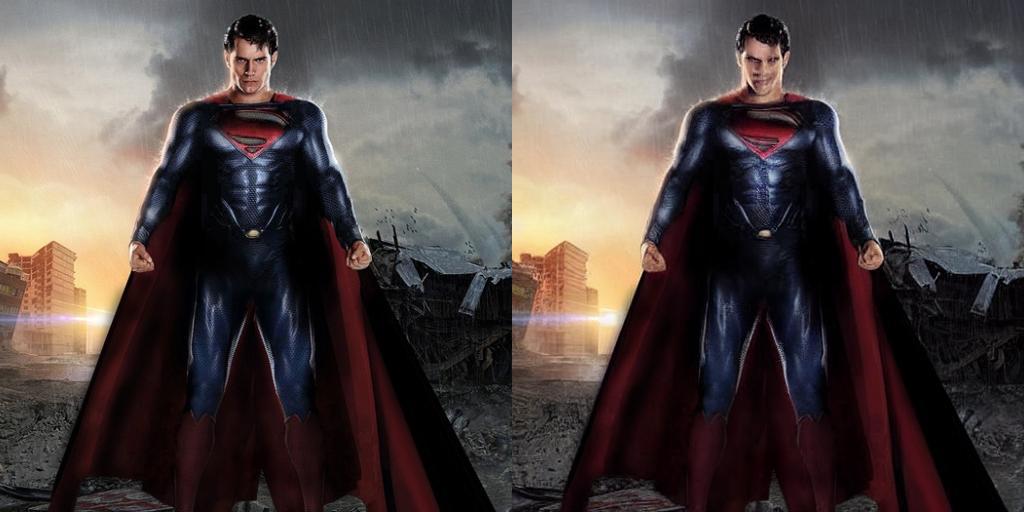}\\
    \parbox{0.45\linewidth}{\centering Man doing yoga in front of waterfall.}
    \hfill
    \parbox{0.45\linewidth}{\centering Man of steel hd wallpaper.}

    \includegraphics[width=0.45\linewidth]{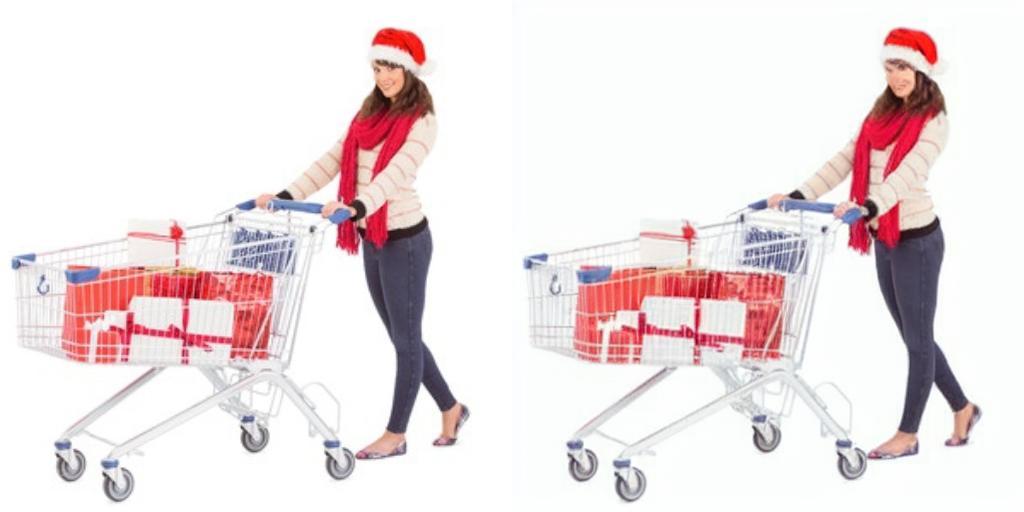}
    \hfill
    \includegraphics[width=0.45\linewidth]{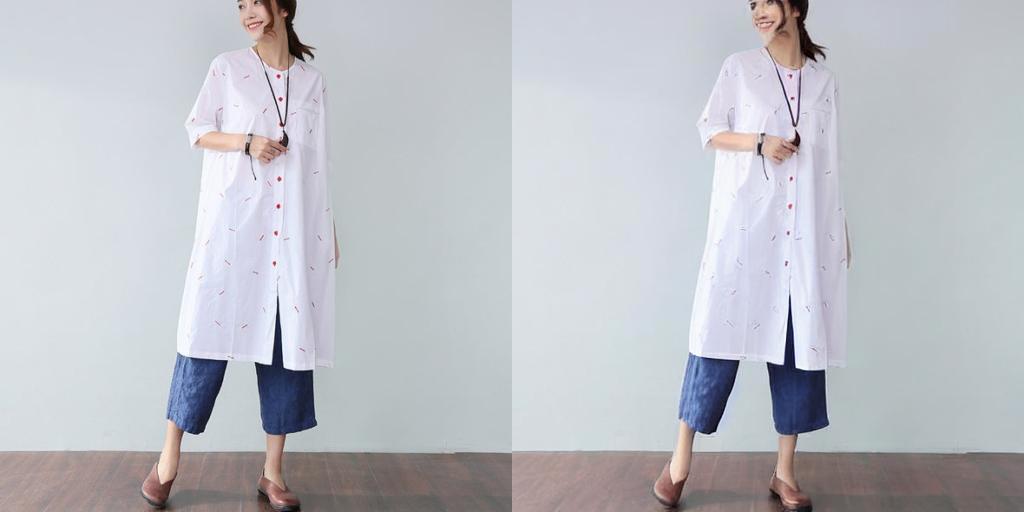}\\
    \parbox{0.45\linewidth}{\centering Woman pushing shopping cart with Christmas gifts.}
    \hfill
    \parbox{0.45\linewidth}{\centering Woman wearing white shirt dress \\with red and white embroidery.}
    \caption{More examples of face pairs. We leverage the inpainting pipeline and control the noise factor for a degraded version, thereby forming a (win, loss) face pair.}
    \label{fig: ds}
    
\end{figure*}

\section{Details in Statistics}
In Table 2 in the main paper, we calculate the ranking alignment of existing popular metrics with human preference on generated face images.
Specifically, the annotated triplets labeled 1 to 3 as scores; 
the higher the score, the better the face.
We also average over all samples to obtain Table 1 in the main paper.
We calculate the accuracy by leveraging such triplets.
In each triplet, we can get three comparison pairs by pairwise comparison.
All the methods give rates to the images and perform pairwise comparisons.
In this case, we can match the results from different methods with the human labels to get the accuracy.

\section{Visualization}
\begin{figure*}
    \includegraphics[width=0.45\linewidth]{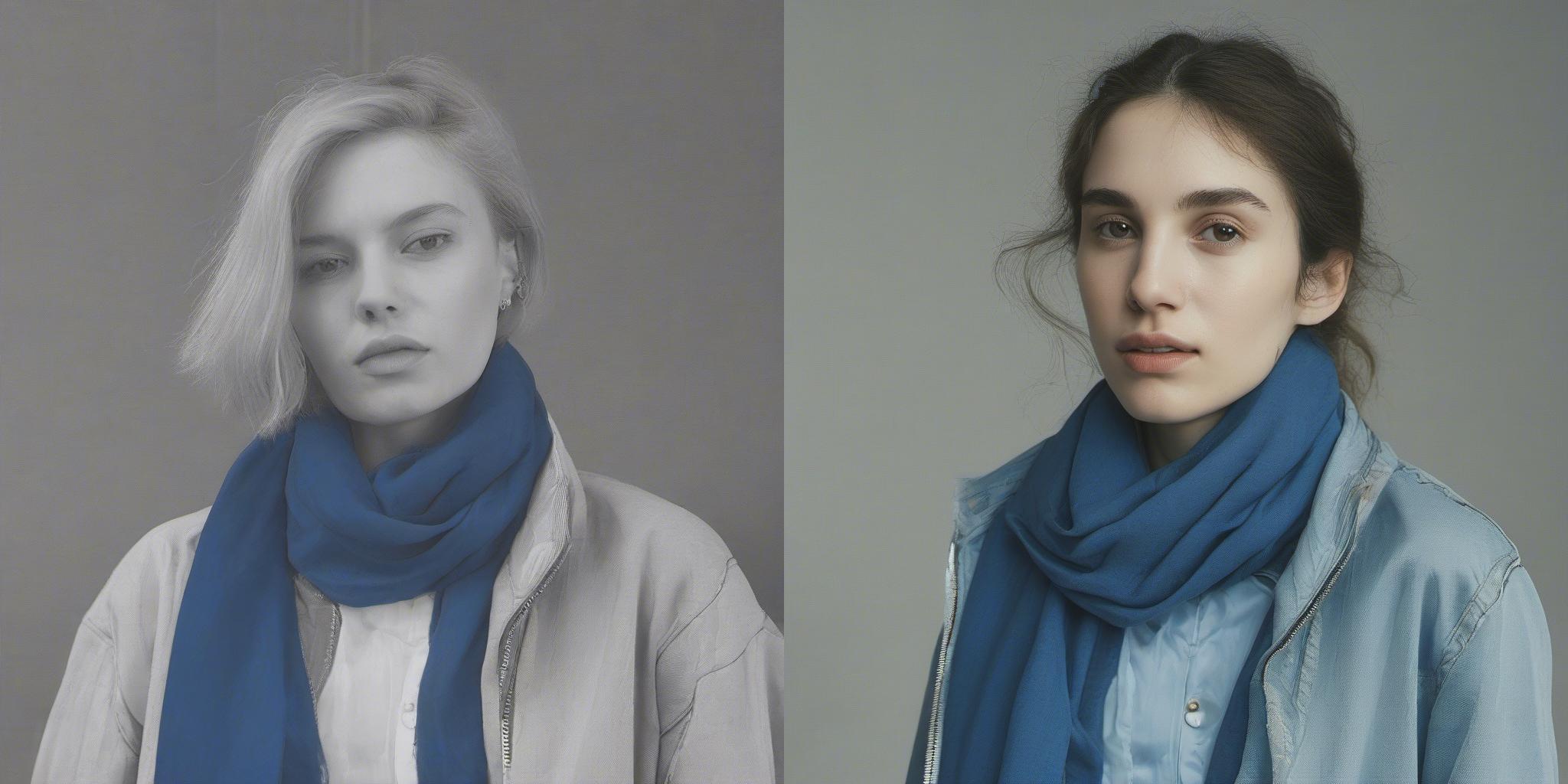}
    \hfill
    \includegraphics[width=0.45\linewidth]{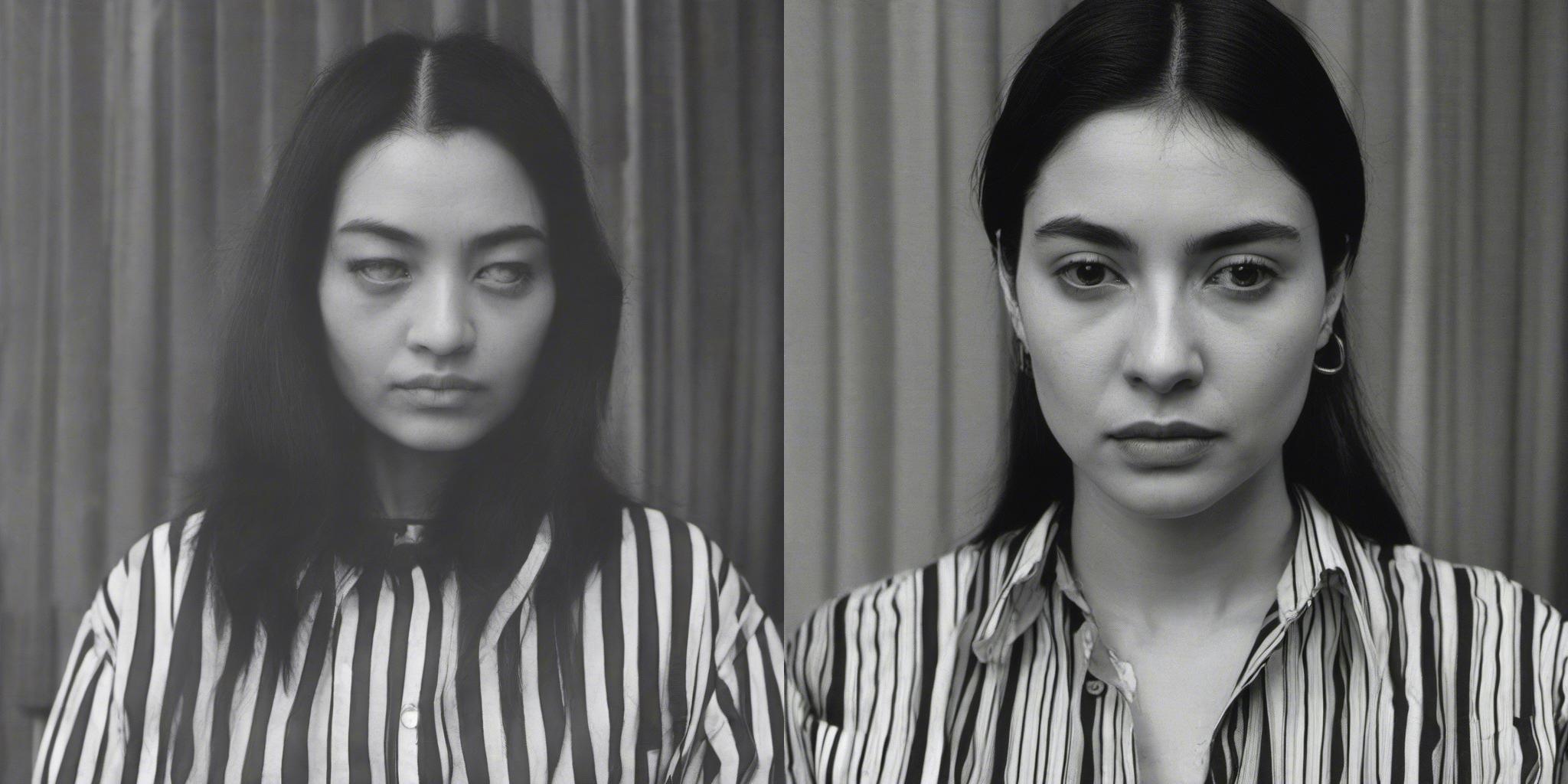}\\
    \parbox{0.45\linewidth}{\centering A woman wearing a blue jacket and scarf.}
    \hfill
    \parbox{0.45\linewidth}{\centering A woman with black hair and a striped shirt.}

    \includegraphics[width=0.45\linewidth]{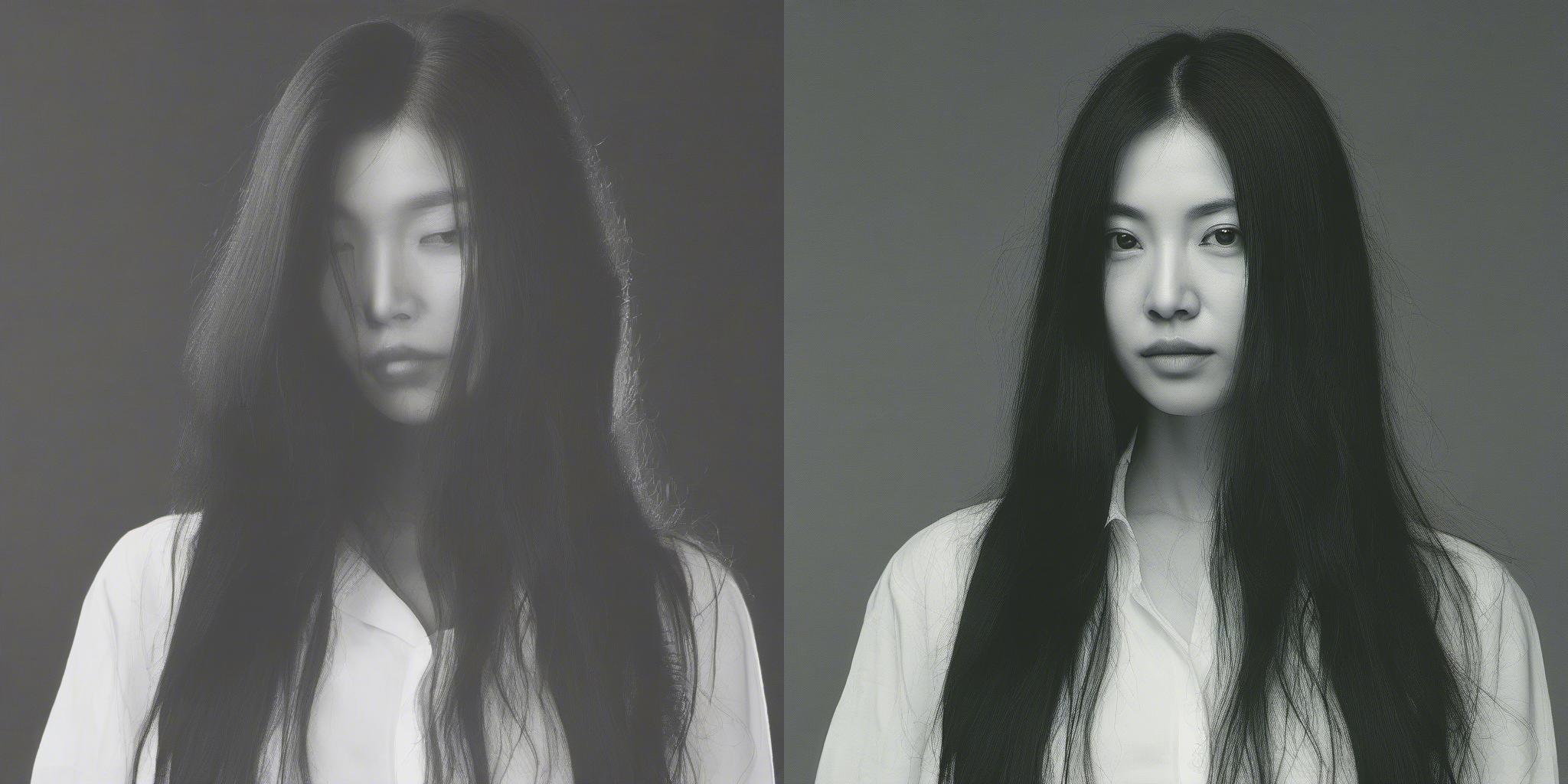}
    \hfill
    \includegraphics[width=0.45\linewidth]{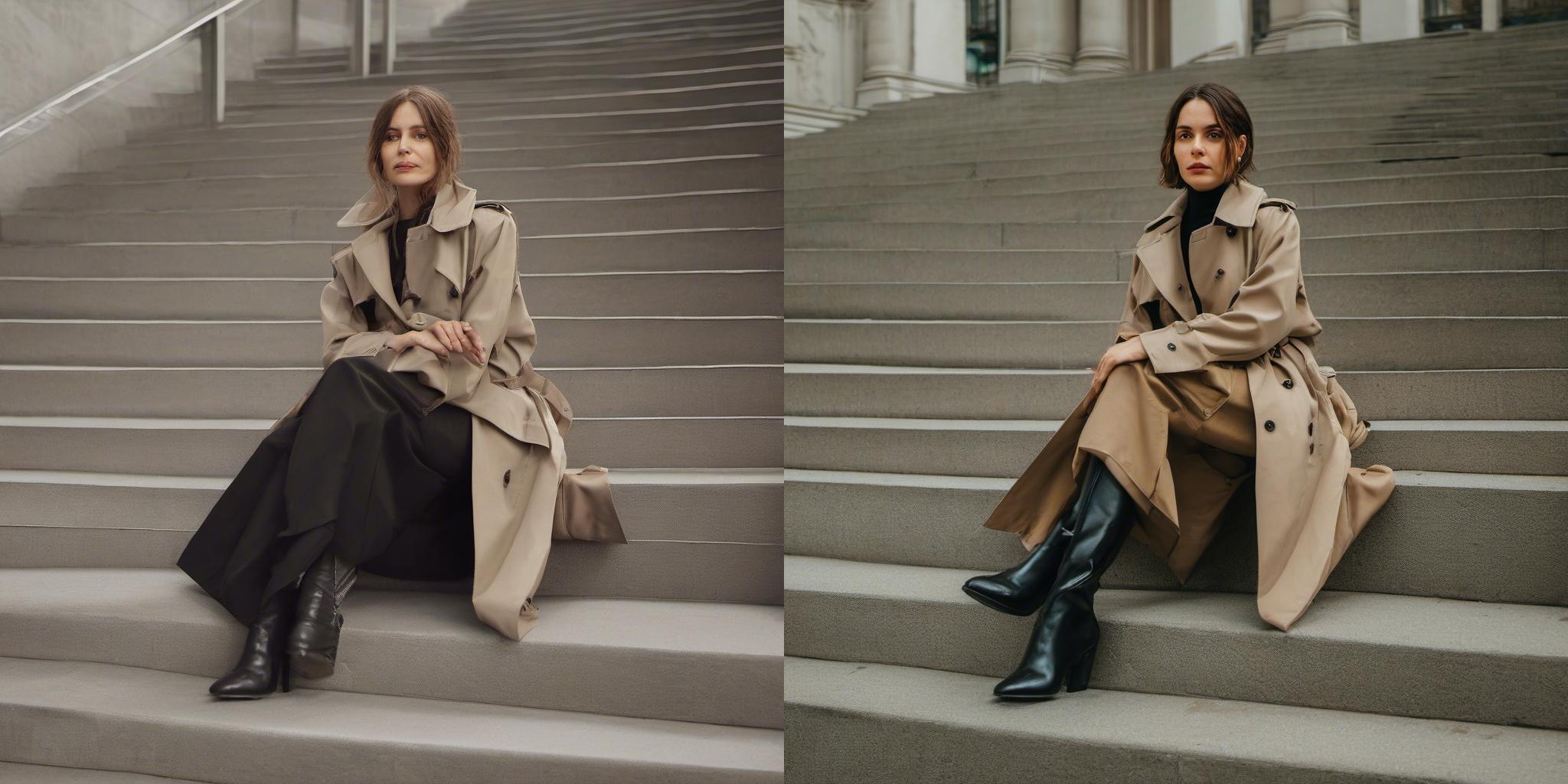}\\
    \parbox{0.45\linewidth}{\centering A woman with long black hair and a white shirt.}
    \hfill
    \parbox{0.45\linewidth}{\centering A woman sitting on some stairs in a trench coat.}

    \includegraphics[width=0.45\linewidth]{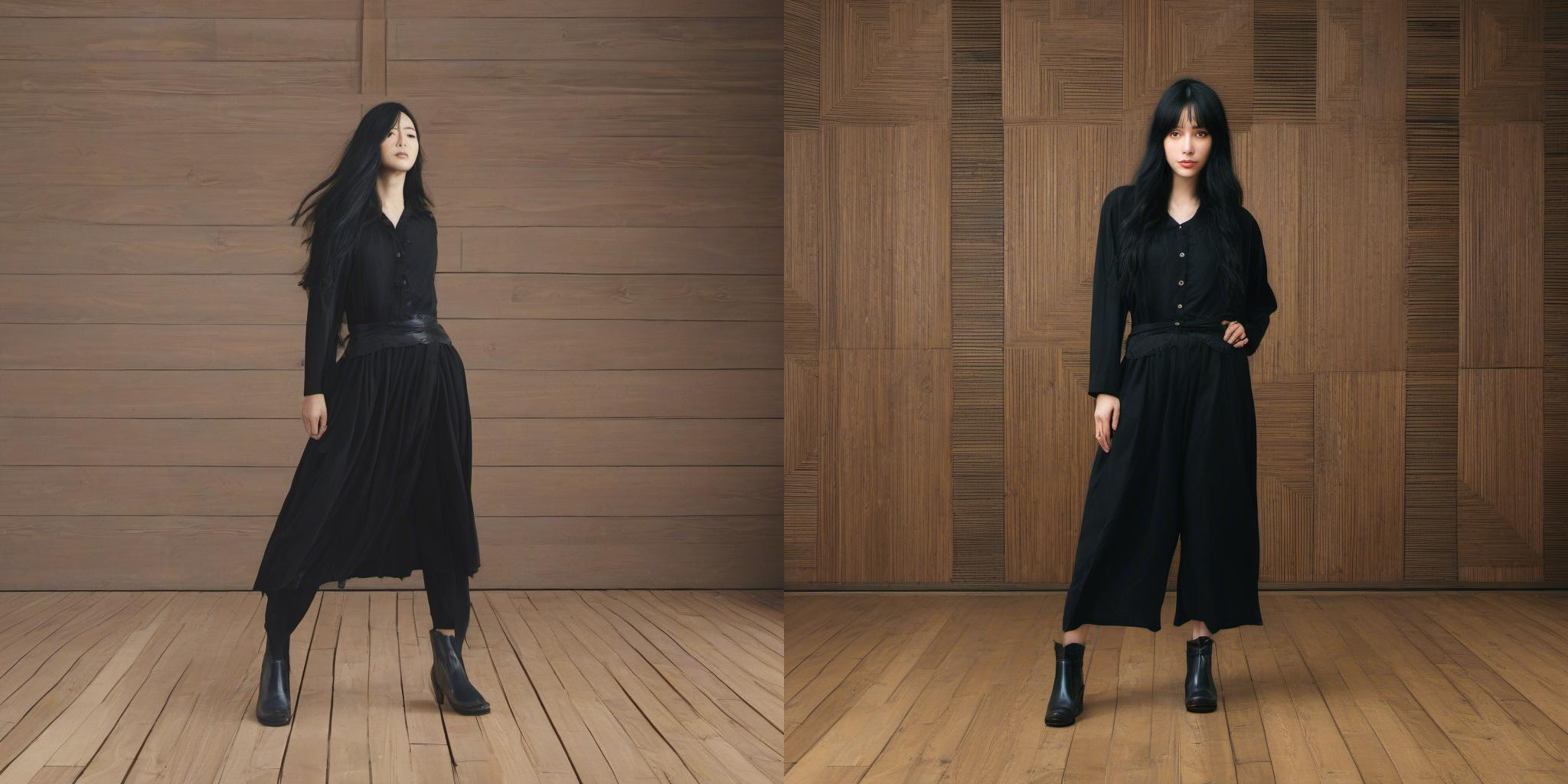}
    \hfill
    \includegraphics[width=0.45\linewidth]{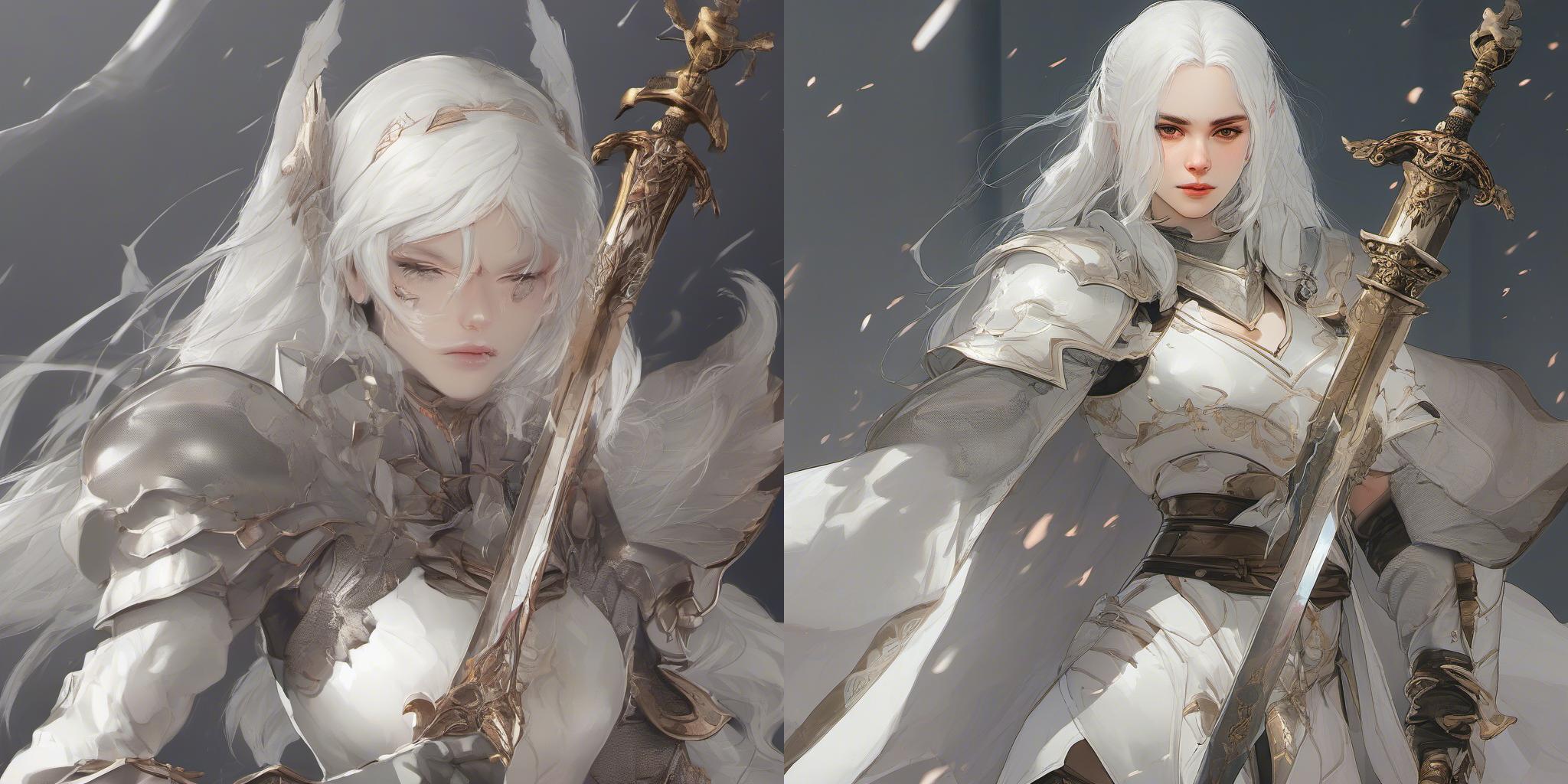}\\
    \parbox{0.45\linewidth}{\centering A woman with long black hair standing on a wooden floor.}
    \hfill
    \parbox{0.45\linewidth}{\centering A woman with white hair and white armor \\is holding a sword.}

    \includegraphics[width=0.45\linewidth]{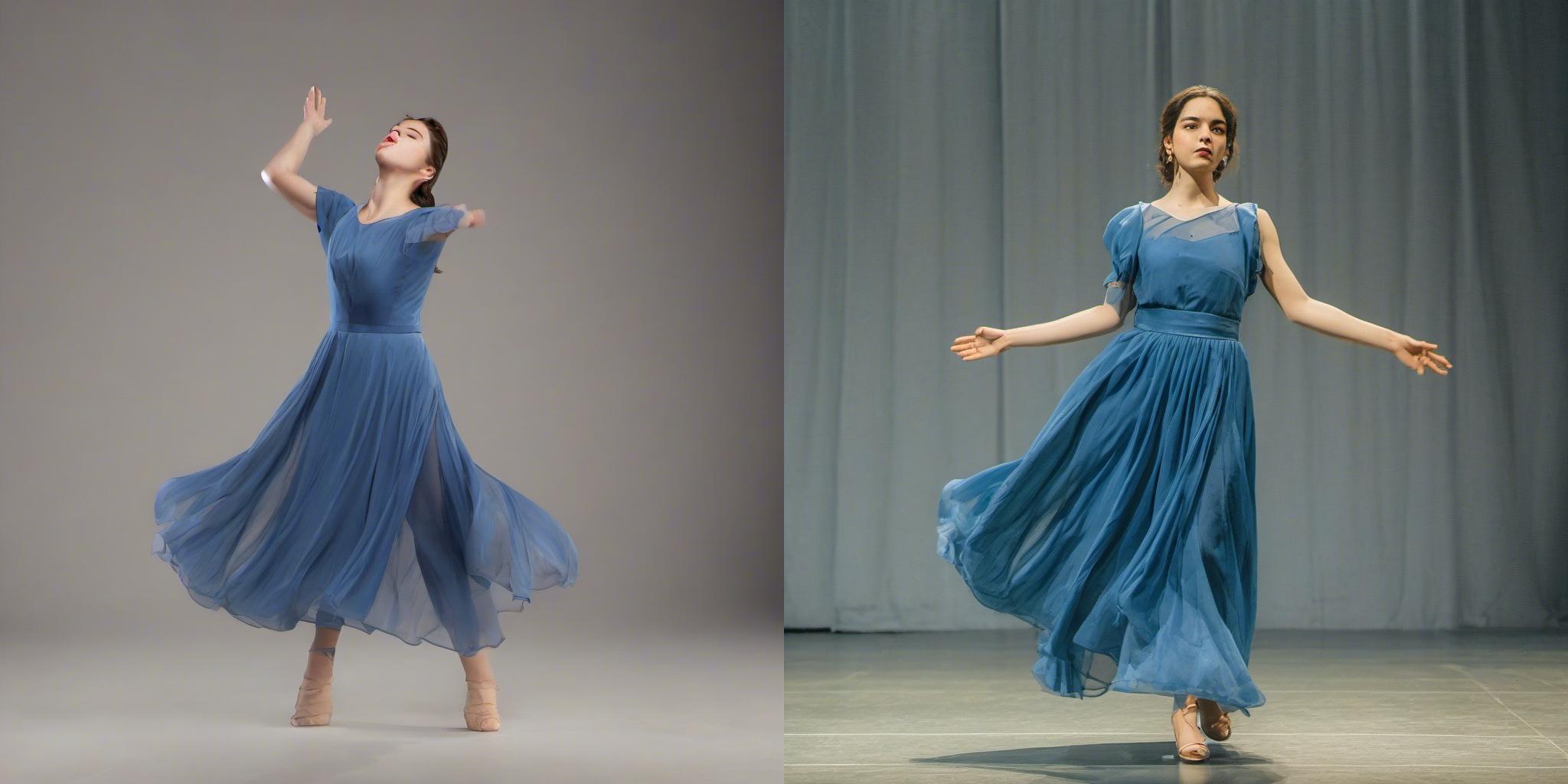}
    \hfill
    \includegraphics[width=0.45\linewidth]{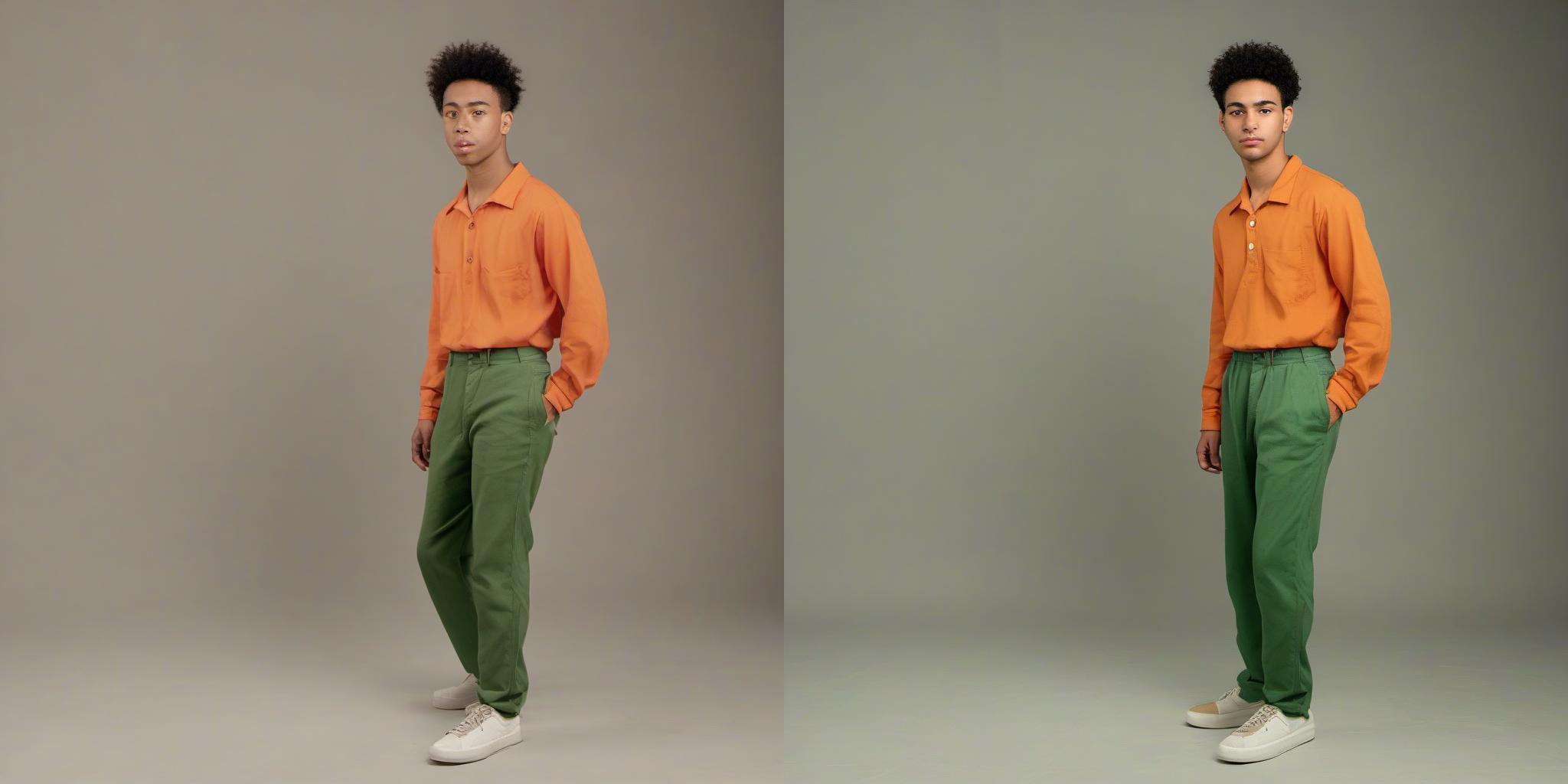}\\
    \parbox{0.45\linewidth}{\centering A young woman in a blue dress performing on stage.}
    \hfill
    \parbox{0.45\linewidth}{\centering A young man in an orange shirt and green pants.}

    \includegraphics[width=0.45\linewidth]{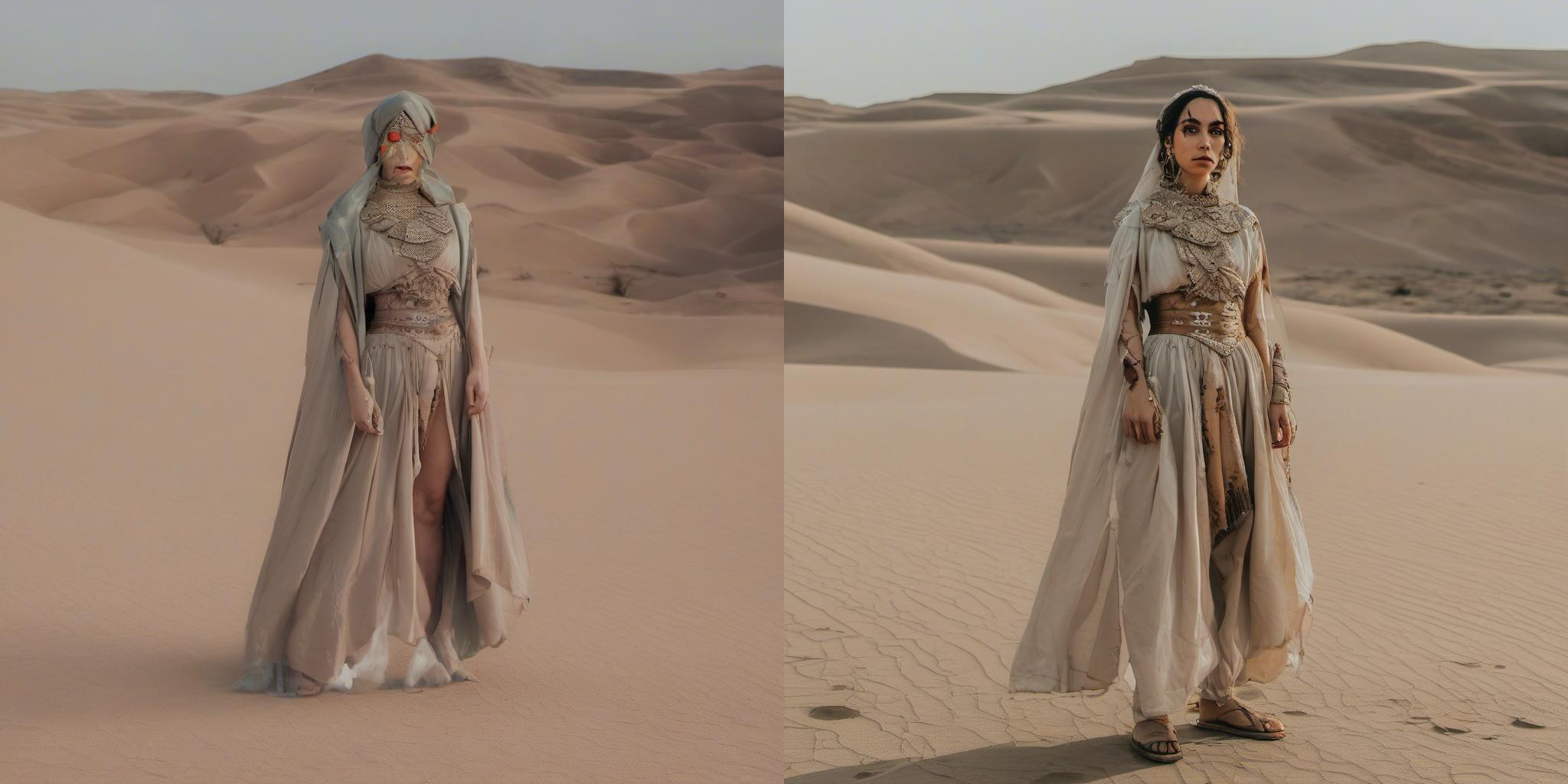}
    \hfill
    \includegraphics[width=0.45\linewidth]{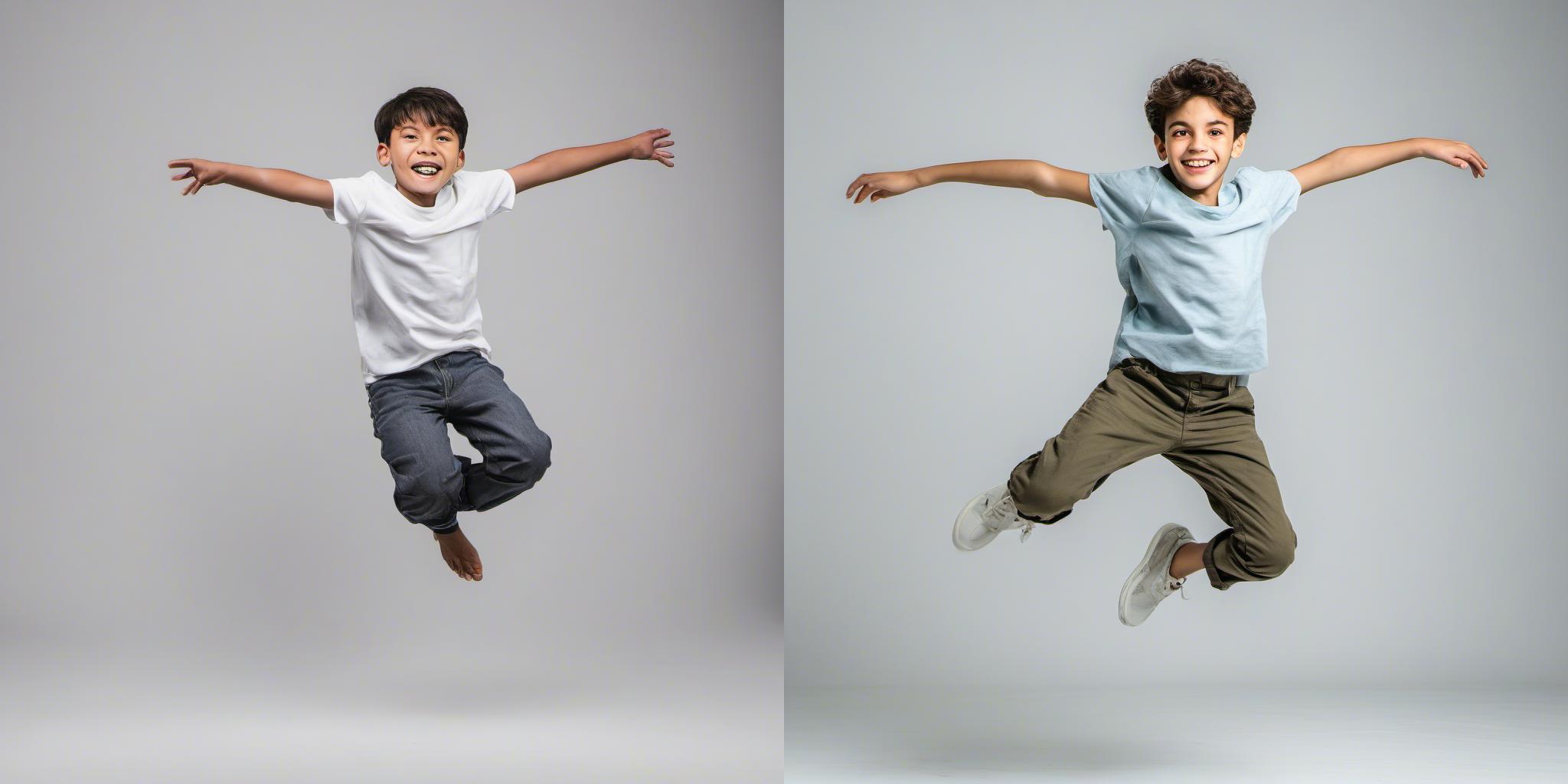}\\
    \parbox{0.45\linewidth}{\centering A woman in a costume standing in the desert.}
    \hfill
    \parbox{0.45\linewidth}{\centering A young boy jumping in the air on a white background.}

    \caption{More comparison visualization between SDXL-Base (left) and our fine-tuned model (right).
    We can see that our model not only generates more normal and attractive faces but also maintains or even increases the overall quality of the images.
    Zoom in for more face details.}
    \label{fig:more compare}
\end{figure*}

We present more annotated triplets in Figure~\ref{fig:evaluation dataset} and more (win, loss) face pair in Figure~\ref{fig: ds}.
More comparison results between our fine-tuned model and SDXL-Base are in Figure~\ref{fig:more compare}, showing a great improvement in face quality.

\end{document}